\newcommand{\para}[1]{\noindent \textbf{#1.}}

\newcommand{\morenomod}[1]{\textcolor{black}{#1}}

\documentclass[10pt,journal,compsoc]{IEEEtran}

\usepackage{cite}

\usepackage{subcaption}
\usepackage{amsmath,amsfonts}
\usepackage{algorithmic}
\usepackage{algorithm}
\usepackage{array}
\usepackage{textcomp}
\usepackage{stfloats}
\usepackage{url}
\usepackage{verbatim}
\usepackage{graphicx}
\usepackage{cite}
\usepackage{booktabs}
\usepackage{hyperref}
\usepackage{cleveref}
\usepackage{dsfont}
\usepackage{colortbl}
\usepackage[dvipsnames]{xcolor}
\usepackage{multicol}
\usepackage{multirow}
\usepackage[thinc]{esdiff}
\usepackage{listings}

\usepackage{pgfplots}
\pgfplotsset{compat=1.17}
\usepackage{wrapfig}

\DeclareMathOperator*{\argmax}{arg\,max}

\newcommand{\singlecaption}{t}
\newcommand{\datasetcaptions}{\mathcal{T}}
\newcommand{\knowledgebase}{\mathcal{B}}

\newcommand{\bias}{\mathcal{B}_j}

\newcommand{\indicatorfunction}{\mathds{1}}

\newcommand{\biasname}{b}

\newcommand{\generator}{\mathrm{G}}

\newcommand{\myquote}[1]{\emph{``#1"}}

\newcommand{\eg}{\textit{e.g.}}
\newcommand{\ie}{\textit{i.e. }}

\ifCLASSINFOpdf
\else
\fi

\begin{document}
\title{GradBias: Unveiling Word Influence on Bias in Text-to-Image Generative Models}

\author{Moreno~D'Incà,
        Elia~Peruzzo,
        Massimiliano~Mancini,
        Xingqian~Xu,
        Humphrey~Shi,
        Nicu~Sebe%
\thanks{ M. D'Incà, E. Peruzzo, M. Mancini, and N. Sebe are with the University of Trento (Trento, Italy). X. Xu and H. Shi are with SHI Labs @ Georgia Tech \& UIUC \& Picsart AI Research (PAIR). Corresponding authors: H. Shi (shi@gatech.edu) and N. Sebe (niculae.sebe@unitn.it)}
}

\IEEEtitleabstractindextext{%
\begin{abstract}
Recent progress in Text-to-Image (T2I) generative models has enabled high-quality image generation. As performance and accessibility increase, these models are gaining significant attraction and popularity: ensuring their fairness and safety is a priority to prevent the dissemination and perpetuation of biases. However, existing studies in bias detection focus on closed sets of predefined biases (\eg, gender, ethnicity). In this paper, we propose a general framework to identify, quantify, and explain biases in an open set setting, \ie without requiring a predefined set. This pipeline leverages a Large Language Model (LLM) to propose biases starting from a set of captions. Next, these captions are used by the target generative model for generating a set of images. Finally, Vision Question Answering (VQA) is leveraged for bias evaluation. We show two variations of this framework: OpenBias and GradBias. OpenBias detects and quantifies biases, while GradBias determines the contribution of individual prompt words on biases. OpenBias effectively detects both well-known and novel biases related to people, objects, and animals and highly aligns with existing closed-set bias detection methods and human judgment. GradBias shows that neutral words can significantly influence biases and it outperforms several baselines, including state-of-the-art foundation models. Code available here: \href{https://github.com/Moreno98/GradBias}{https://github.com/Moreno98/GradBias}.
\end{abstract}

}

\maketitle

\IEEEdisplaynontitleabstractindextext

\IEEEpeerreviewmaketitle

\section{Introduction}\label{sec:intro}
Text-to-Image (T2I) generation has caught the attention of the general public thanks to the recent advancements in quality and fidelity, %
reaching photo-realistic results~\cite{saharia2022photorealistic, nichol2021glide, hierarchical_text_to_image_CLIP, LDM_2022_CVPR, podell2023sdxl}. These advancements %
unlocked multiple use cases, including image editing~\cite{epstein2023diffusion, brooks2023instructpix2pix, hertz2022prompt, goel2023pairdiffusion}, personalization~\cite{Ruiz_2023_CVPR, gal2022textual} and additional conditioning~\cite{avrahami2023spatext, ControlNet_2023_ICCV, huang2023composer} allowing even unexperienced users to obtain high-quality images. With this growing popularity, %
it is becoming increasingly important to investigate the safety and fairness of such models to avoid unwanted and unexpected behaviours~\cite{ITIGEN_2023_ICCV, fairDiffusion2023, DALL_EVAL_2023_ICCV}. %
Studying ethical issues in T2I generative models is an open area of research as defining the concept of \textit{bias} presents significant challenges due to its inherently subjective nature and diversity~\cite{fairDiffusion2023}. As biases can range from more visual ones (\eg, gender and ethnicity), to more complex ones (\eg, socioeconomic status and cultural representation), %
establishing a universal definition remains challenging~\cite{verma2018fairness}.

\begin{figure*}[htbp]
    \centering
    \includegraphics[width=0.8\linewidth]{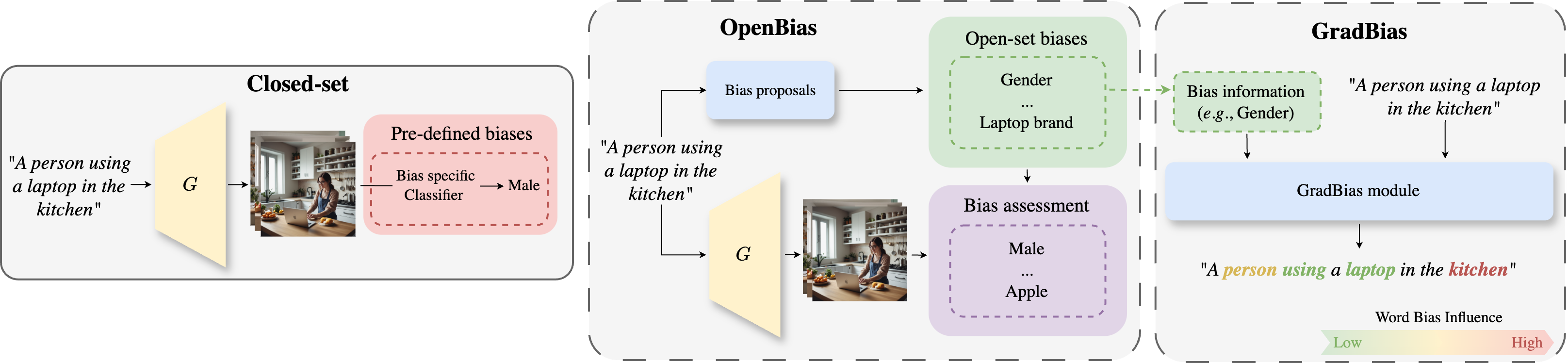}
    \vspace{-0.2cm}
    \caption{We propose a modular framework for open-set bias evaluation. In contrast to previous works~\cite{fairDiffusion2023, ITIGEN_2023_ICCV, kenfack2022repfairgan}, our pipeline does not require a predefined list of concepts but proposes a set of novel biases. We implement two variations: OpenBias discovers general biases in T2I models, while GradBias detects the influence of each prompt word on the bias.}
    \label{fig:teaser}
    \vspace{-0.3cm}
\end{figure*}

Given a concept $t$ and a set of classes $\mathcal{C}$ describing $t$, we define a T2I model $G$ as biased against the concept $t$ if the generated images do not follow a uniform distribution across $\mathcal{C}$~\cite{fairDiffusion2023, xu2018fairgan}. Therefore $G$ tends to favor a particular class $c \in \mathcal{C}$ (\eg, ``man") when provided with a class-neutral textual prompt (\eg, \myquote{A chef cooking.}).
Over the years, bias mitigation techniques have been proposed to control or remove specific biases without compromising performance~\cite{ITIGEN_2023_ICCV, fairDiffusion2023}, ranging from training-time~\cite{nam2020learning, savani2020intra, Wang_2020_CVPR, jung2022learning} to data augmentation methods~\cite{Agarwal_2022_WACV, D'Inca_2024_WACV}. 
Nevertheless, applying these techniques poses a significant challenge, as retraining T2I models is often prohibitively resource-intensive. Consequently, inference-time methods have gained attention, including fairer guidance methods~\cite{fairDiffusion2023, parihar2024balancing, shen2024finetuning} and prompt learning~\cite{ITIGEN_2023_ICCV}. While these methods yield great results, they all require a pre-defined set of biases, limiting their applicability to well-known concepts such as gender, age, ethnicity~\cite{DALL_EVAL_2023_ICCV, fairDiffusion2023, parihar2024balancing, shen2024finetuning} and facial attributes~\cite{ITIGEN_2023_ICCV}.

We argue that T2I models may exhibit unknown biases not captured by close-set detection pipelines. The example in~\cref{fig:teaser} illustrates an overview of the close-set and open-set frameworks. A caption \myquote{A person using a laptop in the kitchen} allows close-set pipelines to identify predefined biases such as gender, age, and race, but may miss other significant biases like laptop brand or kitchen style. This limitation prompts the question: \myquote{Can we detect biases in an open-set fashion?}. This novel task poses challenges, including the difficulty of identifying all potential biases and the prohibitive cost of data collection.

We introduce a general pipeline for discovering and evaluating biases in T2I models, without requiring predefined categories. This pipeline exploits a Large-Language-Model (LLM) to propose biases starting from textual captions. %
Afterward, it queries a Visual Question Answering (VQA) model with the generated images and the potential biases, previously identified. %
This design removes the need for attribute-specific classifiers, which are unsuitable for open-set settings, thereby addressing another key limitation of previous works~\cite{fairDiffusion2023, ITIGEN_2023_ICCV, su2023unbiased, parihar2024balancing, shen2024finetuning}.

We implement two variations of this pipeline: OpenBias and GradBias. OpenBias studies the general behavior of T2I models, querying the VQA to evaluate the presence of the proposed biases. GradBias analyzes biases at the sentence level, effectively modeling how different words in a sentence contribute to the overall bias. This reflects a practical scenario, mirroring the standard use case of such models, where users seek to control the generation based on a specific prompt. For example, a gender-neutral word like ``\textit{kitchen}" might influence the model to produce a gender-biased output (as shown in~\cref{fig:teaser}). Investigating these spurious correlations can offer deeper insights into T2I models, aiding researchers in creating fairer systems.

The contributions of this work are:
\begin{itemize}
    \item To the best of our knowledge, we are the first to address the problem of bias detection in an open-set fashion without requiring a predefined list of biases.
    \item \morenomod{We present a novel framework specifically designed for open-set bias evaluation. This pipeline leverages an LLM to build a knowledge base of biases and a VQA model for bias assessment. }
    \item \morenomod{We propose two variations of this pipeline: OpenBias and GradBias. OpenBias studies biases at high-level, automatically detecting well-known biases and uncovering novel ones. GradBias identifies the learned spurious correlations between individual words and the generated content, providing insights into the role of specific words in contributing to the bias.}
    \item We evaluate both OpenBias and GradBias on multiple versions of Stable Diffusion~\cite{fairDiffusion2023, parihar2024balancing, seshadri2023bias}. OpenBias demonstrates its agreement with closed-set bias detection methods and human judgment, while GradBias outperforms several newly introduced baselines including those based on large multi-modal models.
\end{itemize}

\morenomod{This paper extends our previous work~\cite{D'Inca_2024_CVPR} by drawing a link between bias discovery and explanation under the same, unified framework. In this regard, we introduce GradBias, a novel method specifically tailored to determine the influence of individual prompt words on biases in T2I models. %
Specifically, while assessing the bias with the VQA model, GradBias computes the cross-entropy loss between the VQA's output and the attribute representing the targeted bias. Through standard backpropagation of this loss, we can estimate the relative contribution of each token of the conditioning prompt with respect to the bias under study.
As this research direction is under-explored, we introduce several baselines, exploiting recent advancements in LLMs and large multi-modal models. GradBias outperforms such baselines and unveils the important correlation between neutral words and bias in T2I models. Beyond introducing GradBias, we also expand the related work section to include works modeling language influence on bias and revise the method section to make the link between OpenBias and GradBias explicit.}

\section{Related work}\label{sec:related_work}
\para{Pipeline with Foundation Models} Large-scale deep learning models, often referred to as foundation models~\cite{bommasani2021opportunities}, are trained on extensive datasets and have demonstrated remarkably high performance in zero-shot settings\cite{NEURIPS2020_gpt_3_few_shot, NEURIPS2022_Chain_of_Thougth, subramanian2022reclip}. These models are tipically trained with self-supervision objectives~\cite{bommasani2021opportunities} making them suitable across different modalities, including text~\cite{touvron2023llama, NEURIPS2020_gpt_3_few_shot}, vision~\cite{oquab2023dinov2, caron2021emerging, dosovitskiy2020image} and even multiple modalities simultaneously~\cite{zhu2023chatgpt_blip2, liu2023llava, radford2021learning}. These advancements enabled us to achieve complex tasks previously unthinkable. Recent works have demonstrated the capabilities of LLMs to generate Python code for querying vision and language models~\cite{gupta2023visual,suris2023vipergpt}, enabling the creation of automatic evaluation pipelines. For instance, TIFA~\cite{TIFA_2023_ICCV} evaluates the image-text faithfulness of T2I models by querying a VQA model with LLM-generated questions. Moreover, captioning can be improved by querying a VQA model with LLM-generated questions, leading to more accurate and detailed descriptions~\cite{zhu2023chatgpt_blip2,chen2023video}. A similar study~\cite{kim2023biastotext} identifies spurious correlations in generated images through captioning and model interpretation. However, the discovered biases are not quantified or categorized. 

\textit{Similarly to the above pipelines, our general framework exploits foundation models for automatic open-set bias detection.} It constructs a knowledge base of biases using an LLM and assesses them via VQA.

\para{Bias detection in generative models} Bias mitigation in generative models has been extensively studied, particularly with GAN-based approaches. \cite{tan2021improving}~alters the latent space semantic distribution at inference time to improve fairness. \cite{kenfack2022repfairgan}~ensures fairer representations for sensitive groups with gradient clipping. The research community has recently shifted its attention towards T2I models. In FairDiffusion~\cite{fairDiffusion2023} user-provided instructions are used to add a fair guidance term to enhance classifier-free guidance~\cite{ho2022classifier}, effectively guiding Stable Diffusion~\cite{LDM_2022_CVPR} toward fairer generation in job-related domains. Similarly, \textit{ITI-GEN}~\cite{ITIGEN_2023_ICCV} exploits prompt learning to make T2I models fairer in the context of handwritten digits. In~\cite{su2023unbiased} unsupervised learning estimates the data manifold of the training set to guide the generation towards fairer representations. Recent works mitigate biases by aligning generated images with a user-defined target distribution. For example, \cite{parihar2024balancing} uses distribution guidance, while \cite{shen2024finetuning} employs distributional alignment loss and finetuning to achieve this alignment. Another line of research focuses on mitigating inappropriateness in T2I generation. \cite{brack2023mitigating}~shows how (negative) prompt and semantic guidance~\cite{NEURIPS2023_4ff83037} can be used to mitigate such behaviour. 

These bias mitigation methods share a notable limitation: defining biases beforehand, limiting their applicability to well-known concepts. We argue that unconsidered and unstudied biases may still exist in T2I models. \textit{Therefore, the framework introduced in this work serves as a complementary tool, enhancing the utility of existing methods by identifying, quantifying, and improving the understanding of additional biases.} 

\begin{figure*}[t]
    \centering
    \includegraphics[width=0.9\textwidth]{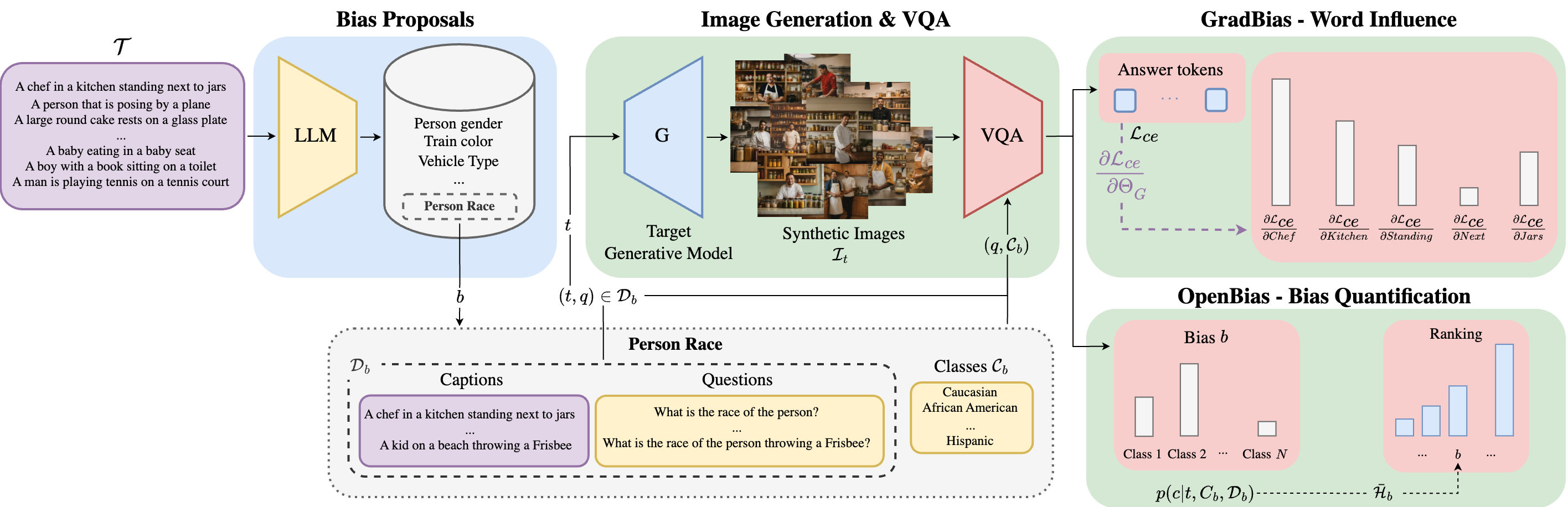}
    \vspace{-0.2cm}
    \caption{\morenomod{We propose a general pipeline for open-set bias detection, quantification, and word-level explanation. Starting with a dataset of real textual captions ($\mathcal{T}$) we use a Large Language Model (LLM) to build a knowledge base $\knowledgebase$ of potential biases occurring in the image generation process. Next, the target generative model synthesizes images using captions where a potential bias has been identified. Finally, Vision Question Answering (VQA) is employed for either bias quantification or word-level bias explanation, depending on whether OpenBias or GradBias is applied.}}
    \label{fig:pipeline_method}
    \vspace{-0.3cm}
\end{figure*}

\para{Language influence on biases} While existing research has explored biases in T2I models, there remains a significant gap in detecting how individual prompt words influence these biases. The first work in this direction is~\cite{lin2023wordlevel} which assesses the influence of each word by iteratively replacing it with synonyms and recalculating bias metrics. While this method may exhibit high performance, it demands high computational resources as it requires regenerating the set of images each time a word is replaced. We refer the reader to~\ref{sec:GradBias_implementation_details} where we treat this approach as upper-bound and use it for determining the ground truths in our experiments. Another approach~\cite{clemmer2024precisedebias} achieves targeted demographic distributions by using an LLM to modify input prompts. However, while effective in guiding demographics, it does not assess the impact of individual prompt words on bias. \cite{vice2024severity}~improves output controllability by manipulating the language embedding space, increasing the expressivity beyond standard language descriptions. This approach generates images that match specific attribute distributions, however, it does not assess the impact of individual prompt words on the bias.

These works improve model explainability and controllability. However, a \textit{direct} method to evaluate the impact of individual prompt words on bias remains unexplored. Hence, we propose a variation of our general pipeline to \textit{directly} assess word-level bias influence and introduce baselines for benchmarking this task.

\section{OpenBias and GradBias}\label{sec:method}

OpenBias identifies and quantifies general biases in T2I models, while GradBias analyzes the impact of individual prompt words on these biases. Both methods share a common pipeline, illustrated in~\Cref{fig:pipeline_method}. The bias proposal stage queries an LLM with real textual captions to build a knowledge base of potential domain-specific biases, eliminating the need for predefined concepts and allowing the discovery of new biases. These captions are then used for generating images with the target generative model. Finally, a Vision-Question-Answering (VQA) model evaluates the potential biases. We describe the general framework and its two variations below.

\subsection{Bias proposal module}\label{sec:bias_proposal}
The bias proposal module aims at building a knowledge base $\knowledgebase$ composed of domain-specific biases. This stage queries an LLM with real captions $\datasetcaptions$ to produce: the potential bias name, the set of classes associated with that bias, and a question for bias assessment.

Formally, for a given caption $\singlecaption \in \datasetcaptions$, we represent the LLM's output as a set of triplets $\mathtt{L}_t=\{(\biasname^t_i, \mathcal{C}^t_i, q^t_i)\}_{i=1}^{n_t}$, where the cardinality of the set $n_t$ is caption-dependent. This triplet $(\biasname, \mathcal{C}, q)$ is composed of the proposed bias $\biasname$, a set of associated classes $\mathcal{C}$, and the question $q$ specific to the caption $t$. We follow the in-context learning paradigm~\cite{NEURIPS2020_gpt_3_few_shot, ICLR_2023_selective_annotations} and prompt the LLM with task descriptions and examples 
(see the Appendix for the exact system prompt). 
We aggregate the above LLM output for the whole dataset by defining the caption-specific biases $B_t$ as the union of its potential biases \ie $B_t=\bigcup_{i=1}^{n_t} b_i$. Therefore, the knowledge base of biases $\knowledgebase$ is built as the union of the caption-level ones, \ie $\mathcal{B}=\bigcup_{t\in\mathcal{T}} B_t$. Afterward, we define a database $\mathcal{D}_b$ specific to the bias $b$ by collecting bias-specific information, \ie captions and questions:
\begin{equation}
    \mathcal{D}_b = \{ (t,q) \mid \forall t \in \mathcal{T},  (x,\mathcal{C},q)\in \mathtt{L}_t, x=b\}.
\end{equation}
Additionally, we define $\mathcal{T}_b = \{t \mid (t,q)\in\ \mathcal{D}_b\}$ as the collection of captions linked to the bias $b$, and $\mathcal{C}_b$ as the union of the set of classes associated with the bias $b$ within $\mathcal{T}$.

However, $\mathcal{D}_b$ does not consider whether the classes are specified in the caption. For example, when generating ``\textit{An image of a large dog}", the dog's size should not be treated as one of the potential biases since it is already stated in the prompt. We address this by employing a two-stage filtering on $\mathcal{D}_b$. In the first phase, given a sample $(t,q)\in\mathcal{D}_b$, we ask the LLM whether the answer to the question $q$ is explicitly present in the caption $t$. In the second stage, we filter out captions that contain either a class $c\in\mathcal{C}_b$ or its synonyms. We identify the synonyms for $\mathcal{C}_b$ using ConceptNet~\cite{ConceptNET}. We find that combining these two stages empirically produces more robust results. This module generates bias proposals in an open-set fashion tailored to the given dataset. OpenBias and GradBias will evaluate these biases at high-level and word-level respectively.

\subsection{Image Generation \& VQA}\label{sec:image_generation_VQA}
After building the knowledge base of potential biases $\knowledgebase$, we evaluate them using the target generative model $\generator$. For a given bias $b \in \knowledgebase$ and caption $t\in\mathcal{T}_b$, we generate a set of $N$ images $\mathcal{I}^t_b$, defined as:
\begin{equation}
    \mathcal{I}^t_b = \{\generator(t,s) | \forall s\in{S}\}
\end{equation}
where $S$ is the set of sampled random noise vectors with cardinality $|S|=N$. By sampling multiple noise vectors, we obtain a distribution of $\generator$'s output on the same prompt $t$. 

In the final stage, the pipeline evaluates the bias $b$ in $\mathcal{I}^t_b$ by querying a state-of-the-art Vision-Question-Answering $\mathtt{VQA}$ model. For each image $I\in \mathcal{I}_b^t$, the VQA model answers the question $q$ associated with prompt $t$ from the pair $(t,q)\in \mathcal{D}_b$, selecting the answer from the available classes $\mathcal{C}_b$. Hence, we define the predicted class $\hat{c}$ as:
\begin{equation}
    \label{eq:vqa}
    \hat{c} = \mathtt{VQA}(I, q, \mathcal{C}_b).
\end{equation}
As shown in~\Cref{fig:pipeline_method}, we implement two variations for this final step. OpenBias quantifies the overall biases identified in the generative model, while GradBias assigns scores to each input word, indicating its influence on the bias.

\subsection{OpenBias - Bias Assessment and Quantification}\label{sec:bias_assessment_and_quantification}
In OpenBias, we quantify the severity of the proposed bias $b\in\mathcal{B}$ by applying the $\mathtt{VQA}$ process to the whole $\mathcal{I}^t_b$ and draw a distribution over the classes $\mathcal{C}_b$.
We then explore two different scenarios: \emph{context-aware} where we examine bias in caption-specific images $\mathcal{I}_b^{t}$, and \emph{context-free} where we consider the entire set of images $\mathcal{I}_b$ associated with the bias.

\subsubsection{Context-Aware Bias}\label{sec:context_aware}
In the context-aware formulation, we capture the bias $\bias$ by considering the caption context. As described in~\Cref{sec:bias_proposal}, the bias proposal module filters out captions that mention bias attributes, retaining only those neutral to the bias. However, other factors within the caption may influence the direction and intensity of the bias. For instance, the captions \textit{``A military is running"} and \textit{``A person is running"} are both neutral to the bias \textit{``person gender"}, but the bias’s direction and intensity may differ based on the context.

While assessing the bias, we consider the role of the context by analyzing the set of images $\mathcal{I}_b^t$ generated from a specific caption $t \in \mathcal{T}$, effectively collecting statistics at the caption-level. Consequently, given a bias $b$, the probability for a class $c\in\mathcal{C}_b$ is computed as:
\begin{equation}
    \label{eq:context_aware}
    p(c | t,\mathcal{C}_b, \mathcal{D}_b) = \dfrac{1}{|\mathcal{I}_b^t|} \sum_{I\in \mathcal{I}_b^t}\indicatorfunction \bigl(\hat{c}=c\bigr)
\end{equation}
where $\hat{c} = \mathtt{VQA}(I, q, \mathcal{C}_b)$ is the prediction of the VQA as defined in \cref{eq:vqa}, and $\indicatorfunction(\cdot)$ is the indicator function. 

\subsubsection{Context-Free Bias}
Here, we are interested in studying the overall behavior of the model $\generator$, offering insights into aspects such as the majority class (\ie the direction of the bias) and the overall bias intensity. We exclude the caption context by averaging the VQA scores for $c\in\mathcal{C}_b$ over all captions $t$ related to that bias $b\in\mathcal{B}$:
\begin{equation}
    \label{eq:context_free}
    p(c | \mathcal{C}_b, \mathcal{D}_b) =\dfrac{1}{|\mathcal{D}_b|}\sum_{(t,q)\in \mathcal{D}_b} p(c | t, \mathcal{C}_b, \mathcal{D}_b) 
\end{equation}
Note that the context-aware bias is a special case of this scenario, where $\mathcal{D}_b$ contains a single instance, \ie $\mathcal{D}_b=\{ (t,q)\}$. 

\subsubsection{Bias Quantification and Ranking}\label{sec:bias_quantification}
The open-set nature of our setting challenges the study of the multiple biases. Therefore, we propose to rank them for a comprehensive analysis. Following previous works~\cite{fairDiffusion2023, xu2018fairgan}, we consider the generative model $\generator$ unbiased for a concept $b$ if the distribution over the classes $\mathcal{C}_b$ is uniform. We measure the bias intensity as the entropy of the class probability distribution, obtained with either~\cref{eq:context_aware} or \cref{eq:context_free}. Since biases may have different cardinalities in $\mathcal{C}$, we normalize the entropy by its maximum value~\cite{wilcox1967indices} to enable comparisons. Finally, we refine this score for readability as:
\begin{equation}
    \label{eq:entropy_score}
    \bar{\mathcal{H}}_b = 1 + \dfrac{\sum_{c\in\mathcal{C}_b}p(c | \mathcal{C}_b, \mathcal{D}_b) \log p(c | \mathcal{C}_b, \mathcal{D}_b) }{\log(|\mathcal{C}_b|)} 
\end{equation}
where $\bar{\mathcal{H}}_b\in[0,1]$, with 0 denoting low bias and 1 high bias.

\subsection{GradBias - Word Influence}\label{sec:GradBias}
As discussed in Sections~\ref{sec:intro} and~\ref{sec:bias_assessment_and_quantification}, bias in T2I models is affected by multiple factors, including caption context and specific words. We focus on specific words and introduce GradBias, a novel method for detecting the influence of prompt words on the bias.
After applying the shared pipeline detailed in Sections~\ref{sec:bias_proposal} and~\ref{sec:image_generation_VQA} and shown in~\Cref{fig:pipeline_method}, GradBias aims to identify how individual words influence the generation of specific attributes $y \in C_b$ related to the proposed bias $b$. 
To achieve this, we first compute the following cross-entropy loss $\mathtt{CE}$:
\begin{equation}
    \label{eq:gradbias_loss}
    \mathcal{L} = \mathtt{CE}(\mathtt{VQA}(I,q,C_b), y)
\end{equation}
where $\mathtt{VQA}(I, q, C_b)$ denotes the logits from the VQA model over $C_b$ (note: this is an abuse of notation as in~\cref{eq:vqa} we have the same notation as argmax), and $y \in C_b$ is the attribute related to the bias $b$. We select $y$ as the \(\argmax\) of the logits, as this attribute (i) describes the bias $b$ and (ii) is present in the generated image $I$, according to the VQA’s prediction. 
We then backpropagate this loss to compute the gradients of the text tokens conditioning the T2I model. Given a prompt $t$ we denote the textual embeddings of its tokens as $\{e_0, e_1, \cdots, e_{n}\}$. For each token embedding $e_i$, we compute the absolute value of the scalar gradients as:
\begin{equation}
    \label{eq:emb_grad}
    e^\prime_i = \left| \diffp{\mathcal{L}}{{e_i}} \right|
\end{equation}
Therefore, $e^\prime_i$ estimates the contribution of the token $e_i$ on generating the specific attribute related to the bias $b$, according to~\cref{eq:gradbias_loss}. The final score for a word $w \in t$ is the gradients of the tokens representing that word, addressing cases where a single word is split into multiple sub-tokens. The intuition is that these gradient magnitudes reveal how changes in specific words affect the VQA model’s output and, consequently, they indicate the contribution of each word in influencing the generative model towards biased outcomes as defined by the classes $C_b$. For example, given the prompt \myquote{A chef in a kitchen} and the bias \myquote{gender}, the scores assigned to each word estimate their contribution in generating a specific bias-related attribute \myquote{male} or \myquote{female}. We show a result related to this example in~\Cref{fig:GradBias_qualitatives}.

\begin{table}[t]
    \scriptsize
    \centering
    \resizebox{\columnwidth}{!}{
        \begin{tabular}{lcccccc}
            \toprule
            \multirow{2}{*}{\textbf{Model}} & \multicolumn{2}{c|}{\textbf{Gender}} & \multicolumn{2}{c|}{\textbf{Age}} & \multicolumn{2}{c}{\textbf{Race}} \\ \cmidrule{2-7}
                                   & Acc. & \multicolumn{1}{c|}{F1} & Acc. & \multicolumn{1}{c|}{F1} & Acc. & \multicolumn{1}{c}{F1} \\
            \midrule
             CLIP-L~\cite{radford2021learning}                          & 91.43          & 75.46             & 58.96             & 45.77             & 36.02             & 33.60             \\
             OFA-Large~\cite{wang2022ofa}                               & \textbf{93.03} & 83.07             & 53.79             & 41.72             & 24.61             & 21.22              \\
             mPLUG-Large~\cite{li2022mplug}                             & \textbf{93.03} & 82.81             & 61.37             & 52.74             & 21.46             & 23.26              \\
             BLIP-Large~\cite{li2022blip}                               & 92.23          & 82.18             & 48.61             & 31.29             & 36.22             & 35.52              \\
              Llava1.5-7B~\cite{liu2023improvedllava,liu2023llava}      & 92.03          & 82.33             & 66.54             & 62.16             & 55.71             & 42.80             \\
              \rowcolor{gray!20} Llava1.5-13B~\cite{liu2023improvedllava,liu2023llava}     & 92.83          & \textbf{83.21}    & \textbf{72.27}    & \textbf{70.00}    & \textbf{55.91}    & \textbf{44.33}    \\
            \bottomrule
        \end{tabular}
    }
    \caption{VQA evaluation on the generated images using COCO captions. In \colorbox{gray!20}{gray} the chosen default VQA model.}
    \label{tab:VQA_eval_coco}
    \vspace{-0.4cm}
\end{table}

\section{Experiments}
We assess OpenBias by (i) comparing it against a state-of-the-art closed-set bias detection classifier and (ii) evaluating its alignment with human judgment via a user study. GradBias is evaluated against several baselines and state-of-the-art large vision and language models.

\subsection{OpenBias}\label{sec:OpenBias_experiments}
In OpenBias, we heavily rely on the zero-shot capabilities of each module. We follow~\cite{fairDiffusion2023} and assess its performance by comparing it against FairFace~\cite{Karkkainen_2021_WACV}, a state-of-the-art classifier on gender, age, and race. Moreover, we evaluate the alignment of OpenBias with human judgment via a user study. Below, we define the implementation details and show the results of these experiments.

\subsubsection{Pipeline implementation}\label{sec:OpenBias_pipeline_implementation}
\para{Datasets} We use captions from two multimodal datasets: Flickr 30K~\cite{young-etal-2014-image} and COCO~\cite{DBLP:journals/corr/LinMBHPRDZ14}. Flickr 30K~\cite{young-etal-2014-image} comprises $30$k in-the-wild images with $5$ captions per image. Similarly, COCO~\cite{DBLP:journals/corr/LinMBHPRDZ14} is a comprehensive dataset spanning a diverse range of images with complex contexts. We filter COCO by selecting captions referring to a single person only, leading to roughly $123$K captions. This extensive subset targets bias detection in the context of people, a critical area for examining biases. Nevertheless, it is worth noting that OpenBias also identifies biases outside this domain to include objects, animals, and actions, as shown in~\Cref{sec:findings_OpenBias}.

\para{Implementation details} The flexibility and modularity of OpenBias allow the replacement of individual modules as soon as better models are available. In this study, we use LLama2-7B as our LLM of the bias proposal module and Llava1.5-13B as our Vision-Question-Answering (VQA) model. This choice follows our evaluation described in~\cref{sec:OpenBias_quantitatives}, where Llava1.5-13B is the best-performing model. While running OpenBias, we randomly sample $100$ captions associated with each bias and generate $10$ images for each caption, leading to $1000$ generated images for each bias.

\begin{table}[t]
    \centering
    \scriptsize
    \resizebox{\columnwidth}{!}{
        \begin{tabular}{l|cccccc}
            \toprule
            \multirow{2}{*}{\textbf{Model}}    & \multicolumn{3}{c|}{\textbf{Flickr 30k}~\cite{young-etal-2014-image}}                 & \multicolumn{3}{c}{\textbf{COCO}~\cite{DBLP:journals/corr/LinMBHPRDZ14}}    \\ 
            \multicolumn{1}{l|}{}                     & gender  & age     & \multicolumn{1}{c|}{race}    & gender  & age     & race    \\ \midrule
            Real                                         & 0 & 0.032 & \multicolumn{1}{c|}{0.030} & 0 & 0.041 & 0.028 \\
            SD-1.5~\cite{LDM_2022_CVPR}                                      & 0.072 & 0.032 & \multicolumn{1}{c|}{0.052} & 0.075 & 0.028 & 0.092 \\
            SD-2~\cite{LDM_2022_CVPR}                                        & 0.036 & 0.069 & \multicolumn{1}{c|}{0.047} & 0.060 & 0.045 & 0.105 \\ 
            SD-XL~\cite{podell2023sdxl}                                        & 0.006 & 0.028 & \multicolumn{1}{c|}{0.180} & 0.002 & 0.027 & 0.184 \\ \bottomrule
        \end{tabular}
    }
    \caption{KL divergence ($\downarrow$) computed over the predictions of Llava1.5-13B and FairFace on generated and real images.}
    \label{table:FairFace_eval_KL}
    \vspace{-0.2cm}
\end{table}

\begin{figure}[t]
\centering
    \includegraphics[width=0.8\linewidth]{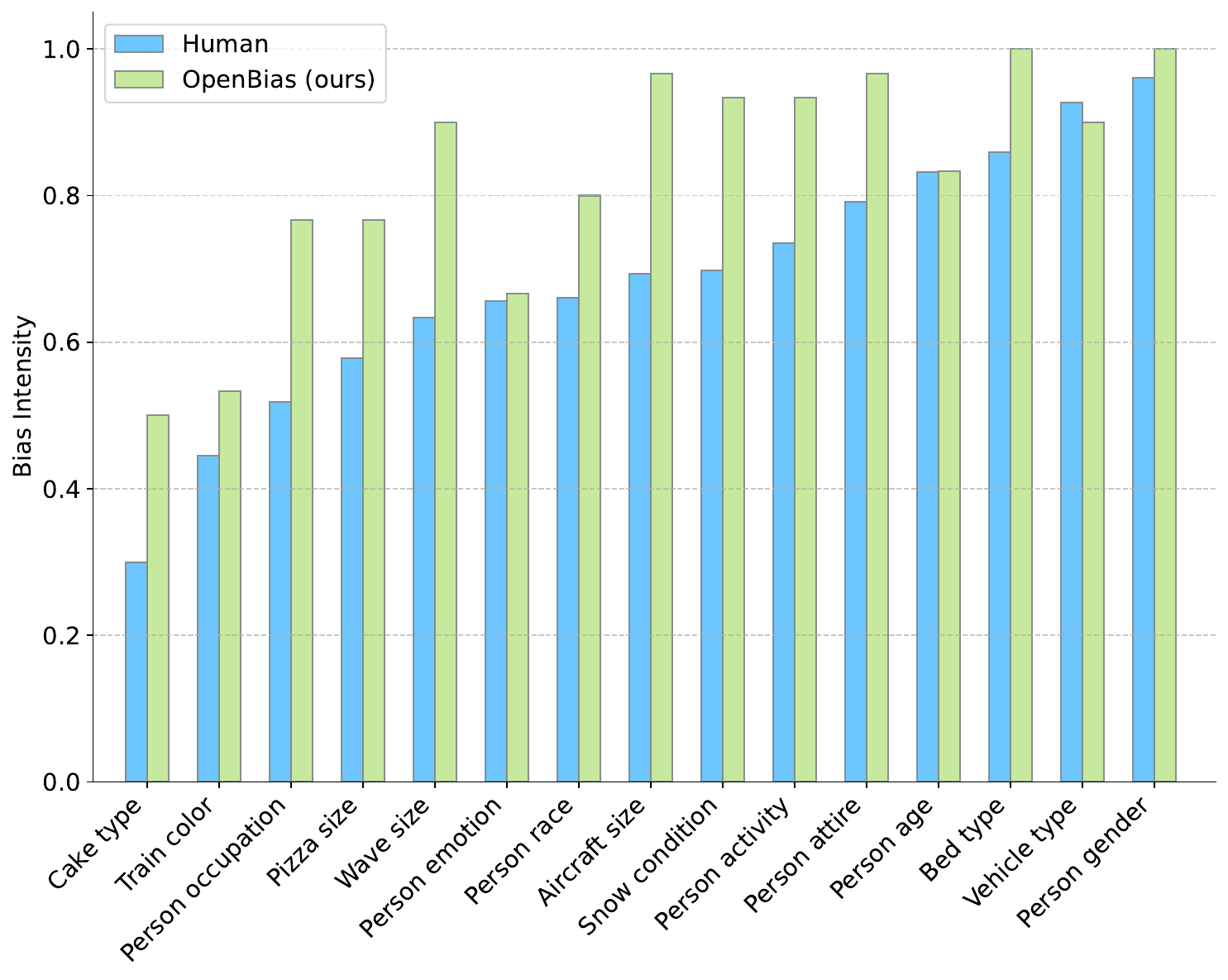}
    \vspace{-0.3cm}
    \caption{Human evaluation results.}
    \label{fig:human_eval}
    \vspace{-0.4cm}
\end{figure}

\subsubsection{Quantitative results}\label{sec:OpenBias_quantitatives}

\para{Agreement with FairFace} We first evaluate the alignment of several SoTA VQA models with FairFace on generated images by Stable Diffusion XL~\cite{podell2023sdxl}. We treat FairFace predictions as ground truths and report accuracy and f1-score in~\Cref{tab:VQA_eval_coco}. Llava1.5-13B~\cite{liu2023improvedllava,liu2023llava} emerges as the best VQA model on gender, age, and race tasks, therefore, we employ it as OpenBias's default VQA. We further test the robustness of Llava1.5-13B by applying the same evaluation method to both real and generated images. We use real pictures from Flickr 30K~\cite{young-etal-2014-image} and COCO~\cite{DBLP:journals/corr/LinMBHPRDZ14} alongside images generated by Stable Diffusion 1.5, 2, and XL~\cite{LDM_2022_CVPR, podell2023sdxl}. The agreement between Llava1.5-13B and FairFace is measured using KL divergence between their respective probability distributions. The results, presented in~\Cref{table:FairFace_eval_KL}, show low KL divergence, indicating strong alignment and confirming the VQA model’s robustness across real and generated images.

\para{User Study} We evaluate OpenBias's alignment with human judgment through a context-aware human study. We generate $10$ for each caption and bias. We publish the survey on crowdsourcing platforms, without geographical restrictions, and randomize the questions’ order. Participants are asked to indicate the bias direction (majority class) and rate its intensity on a scale from $0$ to $10$, with an option for “No Bias” if none is perceived. This study comprises $15$ diverse object-related and person-related biases, accounting for $390$ total presented images. We receive $2200$ valid responses from $55$ unique individuals. We compare the participants' intensity scores with OpenBias in~\Cref{fig:human_eval}. We can observe high alignment on multiple biases, especially for ``\textit{Person age}”, ``\textit{Person gender}”, ``\textit{Vehicle type}”, ``\textit{Person emotion}”, and ``\textit{Train color}”. The study covers $15$ object-related and person-related biases, resulting in $390$ presented images. We receive $2200$ valid responses from $55$ unique individuals. We compare the participants’ intensity ratings with OpenBias in~\Cref{fig:human_eval}, finding high alignment on multiple biases, including \myquote{Person age}, \myquote{Person gender}, \myquote{Vehicle type}, \myquote{Person emotion}, and \myquote{Train color}. We also compute the Absolute Mean Error (AME) between the predicted model’s bias intensity and the average user score, resulting in an $AME=0.15$. Additionally, we assess bias direction by comparing the agreement on the majority class, with OpenBias aligning with human choices 67\% of the time. It is important to note that concepts of bias and fairness are highly subjective, potentially introducing errors in the evaluation process. Nevertheless, our results show a correlation between human and OpenBias scores, validating our pipeline.

\begin{table*}[t]
    \centering
    \scriptsize
    \begin{tabular}{lccc|ccc|ccc}
        \toprule
        \multirow{2}{*}{\textbf{Model}} & \multicolumn{3}{c|}{\textbf{Stable Diffusion 1.5}~\cite{LDM_2022_CVPR}} & \multicolumn{3}{c|}{\textbf{Stable Diffusion 2}~\cite{LDM_2022_CVPR}} & \multicolumn{3}{c}{\textbf{Stable Diffusion XL}~\cite{podell2023sdxl}} \\
        \cmidrule(lr){2-4} \cmidrule(lr){5-7} \cmidrule(lr){8-10}
         & \textbf{Top-1} & \textbf{Top-2} & \textbf{Top-3} & \textbf{Top-1} & \textbf{Top-2} & \textbf{Top-3} & \textbf{Top-1} & \textbf{Top-2} & \textbf{Top-3} \\
        \midrule
         Random                                                 & 36.36 & 62.09 & 81.32 & 37.28 & 63.08 & 81.25 & 37.54 & 62.87  & 81.04\\\midrule
         Dependency Tree                                        & 36.86 & 65.36 & 81.57 & 39.39 & 62.12 & 82.58 & 39.39 & 62.40  & 80.56\\\midrule
         Llama2-13B~\cite{touvron2023llama}                     & 35.14 & 64.37 & 79.85 & 41.67 & 64.39 & 79.04 & 41.18 & 66.50  & 80.05\\
         Llama3-8B~\cite{touvron2023llama}                      & 39.07 & 66.58 & 82.56 & 40.40 & 69.19 & 82.58 & 37.85 & 66.50  & 81.07\\
         Llama3-70B~\cite{touvron2023llama}                     & 39.07 & 63.64 & 81.57 & 39.14 & 67.42 & 83.59 & 41.18 & 65.98  & 81.33\\\midrule
         Llava-1.5-13B~\cite{liu2023improvedllava,liu2023llava} & 37.35 & 62.65 & 81.08 & 44.70 & 67.93 & 83.84 & 41.43 & 62.92  & 79.03 \\\midrule
         \rowcolor{gray!20} GradBias (CLIP-L)                                      & \textbf{48.65} & \textbf{71.50}  & \textbf{86.98} & \textbf{47.98} & \textbf{78.79} & \textbf{89.14} & \textbf{44.25} & \textbf{66.24} & \textbf{84.65}\\
        \bottomrule
    \end{tabular}
    \caption{GradBias evaluation on Stable Diffusion 1.5,2, and XL~\cite{LDM_2022_CVPR, podell2023sdxl}.}
    \label{tab:GradBias_evaluation}
    \vspace{-0.3cm}
\end{table*}

\subsection{GradBias}\label{sec:GradBias_experiments}
We evaluate GradBias against newly introduced baselines, ranging from natural language processing (NLP) techniques to methods leveraging advanced foundation models. Additionally, we perform an ablation study to examine various components of our pipeline. Below, we provide the implementation details and results regarding these experiments.

\subsubsection{Pipeline implementation}\label{sec:GradBias_implementation_details}
\para{Dataset} As discussed in~\cref{sec:intro}, understanding the impact of specific prompt words on the bias remains an under-explored area, resulting in a lack of task-specific data. To address this, we build a textual dataset to evaluate GradBias. This dataset includes $11$ biases related to people and objects identified by OpenBias, with up to $50$ randomly sampled captions per bias, leading to $391$ captions.

\para{Groundtruth computation} In GradBias, each word can influence bias to different extents, and multiple words may have similar impacts. For instance, in the prompt \myquote{a nurse in the hospital}, \myquote{nurse} may heavily influence the bias towards females, while \myquote{hospital} might have a neutral impact. On the contrary, in \myquote{a scientist in the laboratory}, both \myquote{scientist} and \myquote{laboratory} might equally reinforce a male bias. Additionally, this influence is model-dependent, varying with the generative model in use. These factors complicate the establishment of a definitive ground truth.

We compute a reliable ground truth by adapting the method from~\cite{lin2023wordlevel}. Their approach involves generating a distinct set of images for each word in the prompt by replacing it with synonyms, thereby assessing how each substitution influences the model’s bias. Given its robustness, we treat this method as an upper bound to compute our ground truths. However, generating a full set of images for every replacement is resource-intensive and impractical. To address this, we modify this method to enhance its efficiency. For a given prompt $t$ and bias classes $C_b$, we first generate $10$ images using the full prompt and establish a baseline distribution over $C_b$ with OpenBias. We then iteratively remove one word at a time, regenerate the images, and observe changes in the class distribution. The influence score for each word is determined by summing the variations in class contributions. The ground truth is defined as the word whose removal causes the most significant shift in the class distribution compared to the full prompt. It is important to mention that we (i) account for scenarios where multiple words equally influence the bias by considering multiple words with equal top scores as the ground truth, and (ii) perform the ground truth computation for each tested model to account for model dependency. Additionally, we exclude (i) stop-words as they do not provide meaningful insights and (ii) words that directly describe or are synonymous with the bias, as we aim to study the influence of bias-neutral words. For example, when studying the bias \myquote{person gender} given the prompt \myquote{A person cooking in the kitchen.} we exclude the word \myquote{person}. This is because \myquote{person} is associated with \myquote{person gender}, and its removal would naturally result in a high influence score.

\para{Implementation details} We experiment with Llava1.5-13B~\cite{liu2023improvedllava,liu2023llava} and CLIP~\cite{radford2021learning} as VQA models for GradBias. In the version with CLIP, we produce an answer via cosine similarity between the generated images and bias-related classes $C_b$. To ensure a fair evaluation, we compute the ground truth using BLIP2~\cite{li2022blip} as the VQA model within OpenBias. This approach removes the risk of introducing unintentional biases, avoiding potential circular reasoning associated with relying solely on Llava1.5-13B for both GradBias and ground truth computation. Finally, we apply GradBias at each denoising step of Stable Diffusion~\cite{podell2023sdxl, LDM_2022_CVPR} and average over the steps. This approach ensures capturing the word influence throughout the entire denoising process, as this influence may change at different stages. In~\Cref{fig:gradbias_ablation_loss_intervals} we ablate on how different denoising steps affect GradBias.

\subsubsection{Quantitative results}\label{sec:gradBias_quantitative_results}

\para{Baselines} We evaluate GradBias by introducing three baselines: (i) a random-based method, (ii) an NLP-based method, and (iii) two foundation-model-based methods. The random approach randomly selects a word from the prompt as the most influential and serves as a lower-bound baseline. The NLP method leverages spaCy to compute the sentence dependency tree, then rotates the tree around the subject. We then apply a breadth-first search (BFS) starting from this root, ranking words based on the order in which they are visited. We have empirically found that starting from the sentence subject improves performance. Finally, we introduce two foundation models based methods that perform answer ranking~\cite{liu2023llava, NEURIPS2020_gpt_3_few_shot, lin2022truthfulqameasuringmodelsmimic} on LLMs and VQAs models. For LLMs, we iteratively ask a yes or no question for each candidate word $t_i \in t$ to determine its influence on the bias $b$, ranking the words based on the ``\textit{yes}" logit probability. We structure the question as four variations of ``\textit{Is} \textbf{\textless candidate\_word\textgreater} \textit{influencing} \textbf{\textless bias \textgreater} \textit{bias in the prompt} \textbf{\textless caption\textgreater}\textit{? Answer only with `yes' or `no'.}" and average over the answers' probabilities. For VQA models, we follow the same procedure but leverage their multi-modal nature by providing the additional context of the generated images. We experiment with Llama2-13B, Llama3-8B, and Llama-70B~\cite{touvron2023llama} as LLMs and Llava1.5-13B~\cite{liu2023improvedllava,liu2023llava} as VQA.

\para{Quantitative Results} We evaluate the baselines and GradBias with top-1, top-2, and top-3 accuracies across $11$ biases, as described in~\Cref{sec:GradBias_implementation_details}. The results are presented in~\Cref{tab:GradBias_evaluation}, where we apply the different methods to Stable Diffusion 1.5, 2, and XL~\cite{LDM_2022_CVPR, podell2023sdxl}. Across all settings, we observe a consistent progression from the random approach to GradBias. The dependency tree method outperforms the random approach but underperforms compared to other methods. This is expected as this approach is computed independently of the bias, making it less robust to bias diversity. In contrast, LLMs benefit from embedded bias knowledge showing a notable improvement. For example, Llama3-70B~\cite{touvron2023llama} outperforms the random baseline by $+2.71\%$ and $+3.64\%$ when applied to Stable Diffusion 1.5 and XL~\cite{LDM_2022_CVPR, podell2023sdxl}, respectively. Similar gains are observed across other Llama configurations. Visual information provides additional insights that text-based approaches lack. Therefore, Llava-1.5-13B~\cite{liu2023improvedllava,liu2023llava} improves over the random baseline by $+7.42\%$ on Stable Diffusion 2. However, it aligns with LLMs in the other settings. In contrast, GradBias shows a significant improvement over the random approach, with gains of $+12.29\%$, $+10.70\%$, and $+6.71\%$ on Stable Diffusion 1.5, 2, and XL, respectively. It also outperforms all other baselines, achieving $+9.58\%$, $+3.28\%$, and $+2.82\%$ higher accuracy than the second-best method across all settings. 

This evaluation underscores GradBias’s robustness against various baselines, including SoTA foundation models, demonstrating its effectiveness as a valuable tool for analyzing the impact of language on AI-generated content.

\begin{figure}[t]
    \centering
    \begin{tikzpicture}
        \tikzset{every picture/.style={font=\tiny}}
        \begin{axis}[
            width=0.8\linewidth,
            height=5cm,
            xlabel={Grad Interval},
            ylabel={Accuracy (\%)},
            xlabel style={font=\scriptsize, yshift=2pt},
            ylabel style={font=\scriptsize, yshift=-2pt},
            symbolic x coords={1, 2, 5, 10, 50},
            xtick=data,
            ymin=0, ymax=100,
            ytick={0, 20, 40, 60, 80, 100},
            legend style={
                at={(0.5,0.15)},
                anchor=north,
                legend columns=-1,
                font=\tiny,  %
                /tikz/column 2/.style={column sep=2pt},
                scale=0.2,   %
                draw=none    %
            },
            grid=major,
            grid style={dashed,gray!30},
            enlargelimits=0.1,
            mark size=2pt,
            nodes near coords,
            every node near coord/.append style={font=\tiny, anchor=90},
        ]
        \addplot[
            color=blue,
            mark=diamond*,
            thin,
        ]
        coordinates {
            (1,47.98) (2,44.95) (5,46.72) (10,47.47) (50,43.43)
        };
        \addplot[
            color=red,
            mark=triangle*,
            thin,
        ]
        coordinates {
            (1,78.79) (2,72.98) (5,77.27) (10,77.78) (50,69.95)
        };
        \legend{Top-1, Top-2}
        \end{axis}
    \end{tikzpicture}
    \vspace{-0.2cm}
    \caption{Ablation on the gradient timing computation using Stable Diffusion 2~\cite{LDM_2022_CVPR}.}
    \label{fig:gradbias_ablation_loss_intervals}
    \vspace{-0.5cm}
\end{figure}
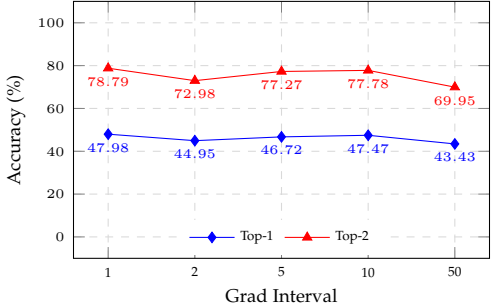

\para{Ablation studies} As discussed in~\Cref{sec:GradBias_implementation_details}, evaluating the impact of computing the loss at different denoising steps is crucial. We assess this by analyzing images generated with Stable Diffusion 2~\cite{LDM_2022_CVPR}, computing the loss at intervals of $1$, $2$, $5$, $10$, and $50$ denoising steps. As shown in~\Cref{fig:gradbias_ablation_loss_intervals}, GradBias performs optimally when the loss is computed multiple times during a single generation—\ie at every $1$, $2$, $5$, and $10$ denoising steps. This is expected as GradBias accounts for the entire denoising process, capturing the complete semantic generation. In contrast, computing the loss only at the final $50$-th step significantly reduces performance, with decreases of $-4.55\%$ in top-1 and $-8.84\%$ in top-2 accuracy. This highlights the importance of considering multiple denoising steps to achieve optimal results. 

We further evaluate GradBias’s robustness by varying the VQA model. We test CLIP~\cite{radford2021learning} and Llava1.5-13B~\cite{liu2023improvedllava,liu2023llava} and show the results in~\Cref{tab:GradBias_ablations}(a). Interestingly, GradBias performs consistently well regardless of the VQA used, suggesting that the choice of VQA has minimal impact on the computed gradients. This highlights that GradBias primarily focuses on the generative model, with the VQA serving primarily as a vehicle for gradient computation. This finding further validates our pipeline, as the gradients effectively capture the intrinsic biases of the generative model, minimizing the influence of the VQA.

In contrast to~\cite{lin2023wordlevel} where generating multiple images is \textit{required} at each word replacement, GradBias should operate effectively with only one image since (i) it avoids computing distributions, and (ii) it relies on gradients to capture each word’s influence on the bias, independently of the number of generated images. We evaluate this by varying the number $N$ of synthesized images, testing GradBias with $N$ equals $1$, $2$, $5$, and $10$. Results in \Cref{tab:GradBias_ablations}(b) show that GradBias performs optimally across all tested quantities, demonstrating its effectiveness even in low data regimes, proving advantageous in low-resource scenarios.

\begin{table}[t]
    \centering
    \scriptsize
    \setlength{\tabcolsep}{0.14cm} 
    \begin{minipage}{0.48\linewidth} %
        \centering
        \begin{tabular}{lcccc}
            \toprule
            \textbf{VQA}  & \textbf{SD-2}  & \textbf{SD-XL}  & \textbf{SD-1.5}  \\
            \midrule
            CLIP-L                       & 47.98 & \textbf{44.25} & 48.65 \\
            Llava-1.5-13B  & \textbf{48.23} & 43.22 & \textbf{49.14} \\
            \bottomrule
        \end{tabular}
        \label{tab:GradBias_VQA_ablation}
    \end{minipage}%
    \hfill
    \begin{minipage}{0.48\linewidth} %
        \centering
        \begin{tabular}{lcccc}
            \toprule
            \textbf{N}  & \textbf{SD-2}  & \textbf{SD-XL}  & \textbf{SD-1.5}  \\
            \midrule
            1                      & 47.47          & 42.19          & 48.48 \\
            2                      & 48.02          & 42.89          & 48.44 \\
            5                      & \textbf{48.46} & 43.91          & 48.44 \\
            10                     & 47.98          & \textbf{44.25} & \textbf{48.65} \\
            \bottomrule
        \end{tabular}
    \end{minipage}
    \parbox{\linewidth}{
        \subcaption*{\hspace{0.05\linewidth}(a)\hspace{0.44\linewidth}(b)}
    }
    \vspace{-0.2cm}
    \caption{We ablate on \textbf{(a)} the VQA used by GradBias, CLIP-L~\cite{radford2021learning} or Llava-1.5-13B~\cite{liu2023improvedllava,liu2023llava} and \textbf{(b)} the number of images $N$ considered by GradBias. We use images generated by Stable Diffusion 1.5, 2, and XL~\cite{LDM_2022_CVPR, podell2023sdxl}.}
    \label{tab:GradBias_ablations}
    \vspace{-0.4cm}
\end{table}

\section{Findings}\label{sec:findings}
In this section, we discuss the findings of OpenBias and GradBias when applied to three popular T2I generative models Stable Diffusion XL, 2, and 1.5~\cite{LDM_2022_CVPR, podell2023sdxl}.
\subsection{OpenSet bias detection -- OpenBias}\label{sec:findings_OpenBias}
We apply OpenBias to captions from COCO and Flickr30k, as detailed in~\Cref{sec:OpenBias_experiments}, to analyze various biases across models and compare context-free versus context-aware bias.

\begin{figure*}[!ht]
    \centering
    \includegraphics[width=0.75\textwidth]{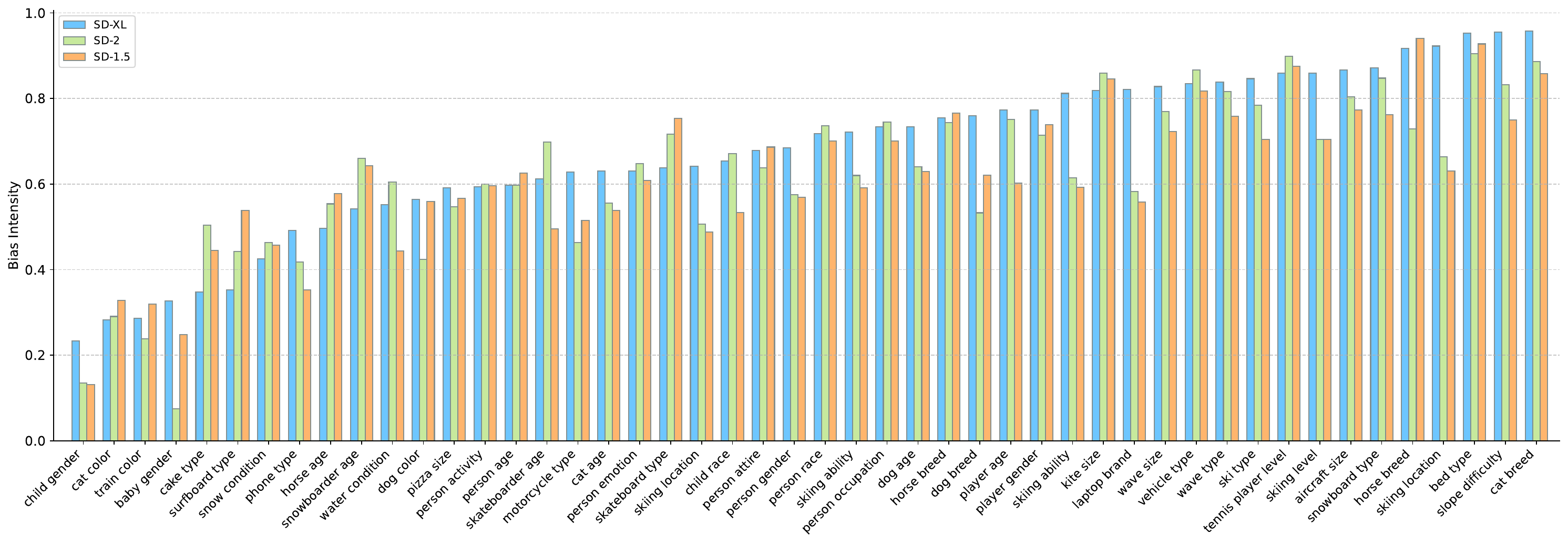}
    \includegraphics[width=0.75\linewidth]{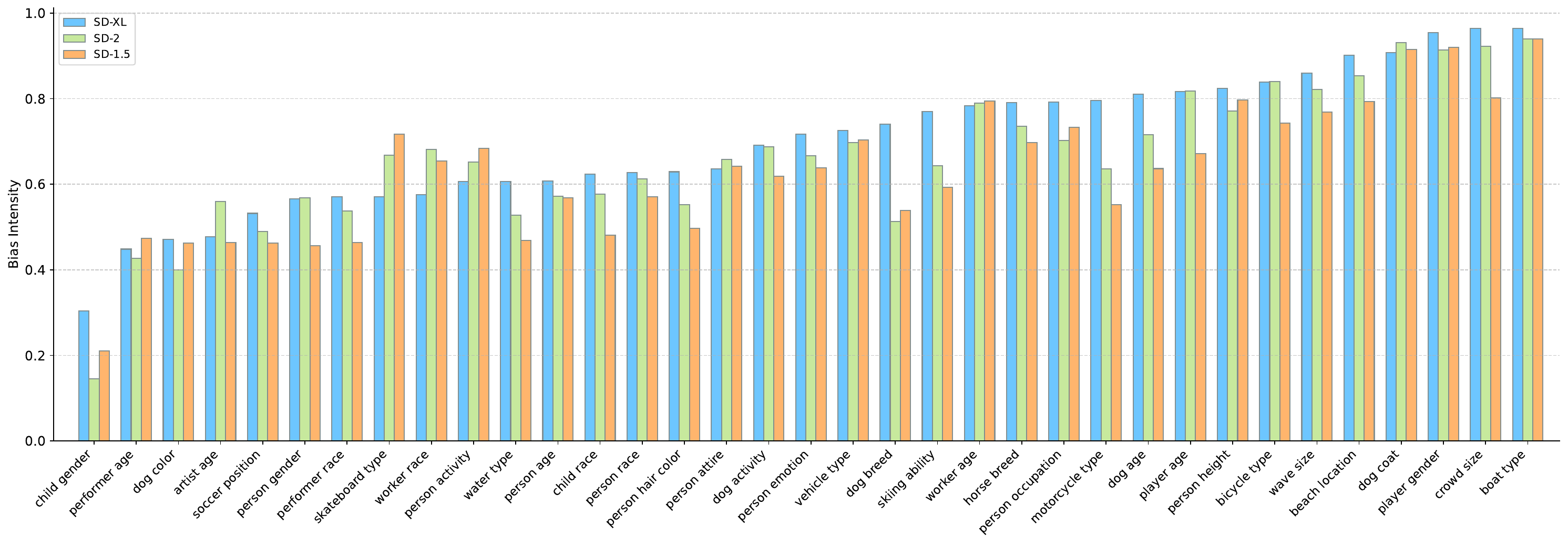}
    \vspace{-0.2cm}
    \caption{Context-aware discovered biases on SD XL, 2 and 1.5~\cite{podell2023sdxl, LDM_2022_CVPR} with COCO~\cite{DBLP:journals/corr/LinMBHPRDZ14} and Flick30k~\cite{young-etal-2014-image} captions respectively.}
    \label{fig:OpenBias_rankings}
    \vspace{-0.4cm}
\end{figure*}

\para{Rankings} In~\Cref{fig:OpenBias_rankings}, we present the rankings of the multiple biases identified by OpenBias. Our method unveils both well-known biases (\ie, \myquote{person gender}, \myquote{person race}) and novel ones (\ie, \myquote{cake type}, \myquote{bed type}, \myquote{laptop brand}). By comparing the generative models we note a correlation in the detected biases, with the XL version amplifying them more than its predecessors. Moreover, OpenBias uncovers more object-centric biases from Flickr captions and person-centric biases from COCO. This diversity highlights OpenBias's ability to propose domain-specific biases based on the dataset, demonstrating its adaptability to various domains.

\begin{figure}[t]
    \centering
    \includegraphics[width=0.7\linewidth]{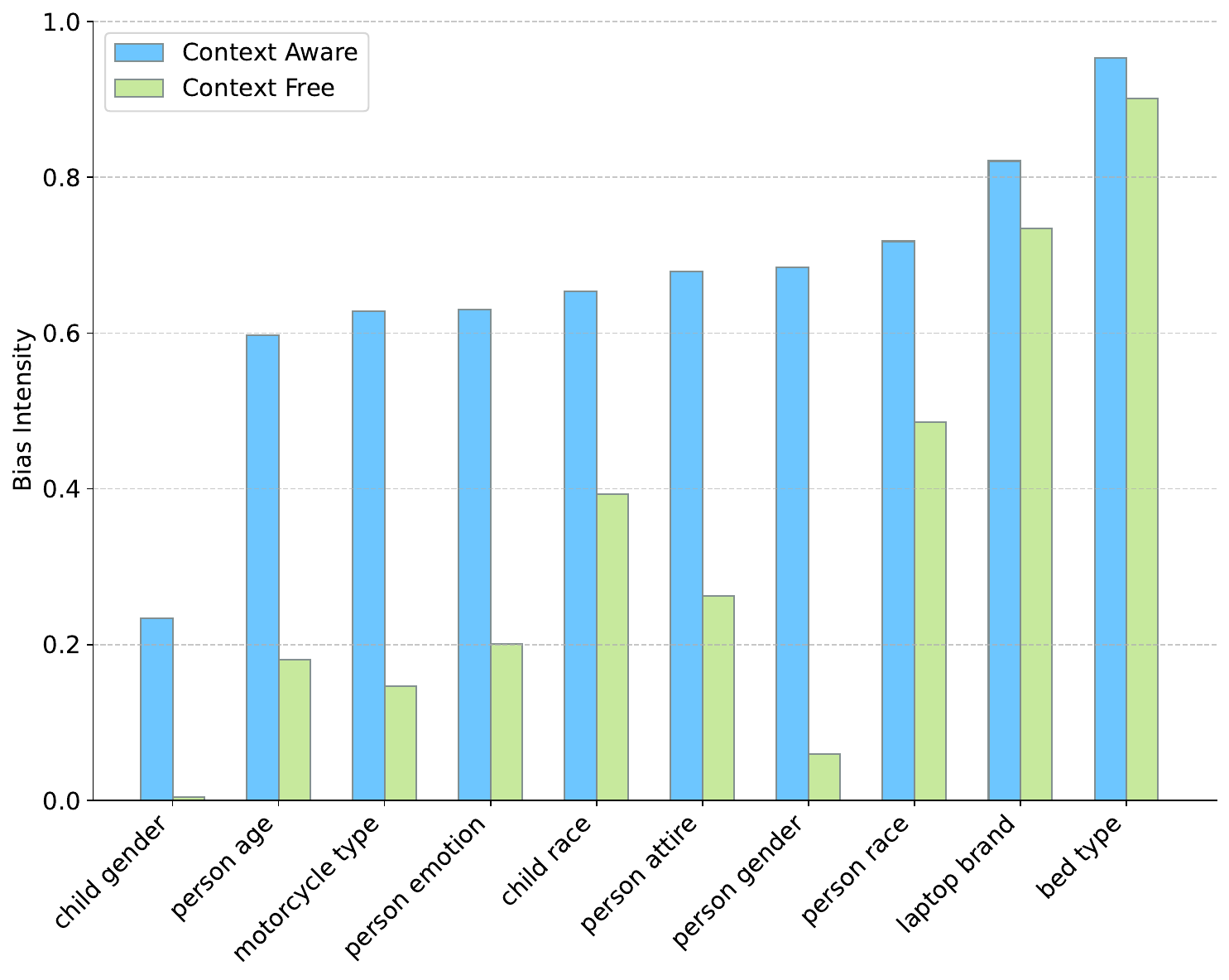}
    \vspace{-0.2cm}
    \caption{Highlighting the importance of the context on Stable Diffusion XL~\cite{podell2023sdxl} with the captions from COCO.}
    \label{fig:context_vs_free}
    \vspace{-0.4cm}
\end{figure}

\para{Context-Free vs Context-Aware} We evaluate how additional elements in the caption affect biases by comparing the generative models in context-free and context-aware settings (see~\Cref{sec:bias_assessment_and_quantification}). In~\Cref{fig:context_vs_free}, we compare these settings using Stable Diffusion XL. The low correlation between scores highlights the importance of context. For example, the bias \myquote{motorcycle type} is significantly higher in the context-aware setting, suggesting that the model generates specific motorcycle types based on the context. Differently, the \myquote{bed type} bias exhibits correlated intensity scores, indicating that the model generates the same bed type regardless of context.

\begin{figure*}[t]%
    \centering
    \begin{subfigure}{0.24\linewidth}
        \centering
        \textbf{Train color}\par\medskip
        \vspace*{-0.05cm}
        \includegraphics[width=\linewidth]{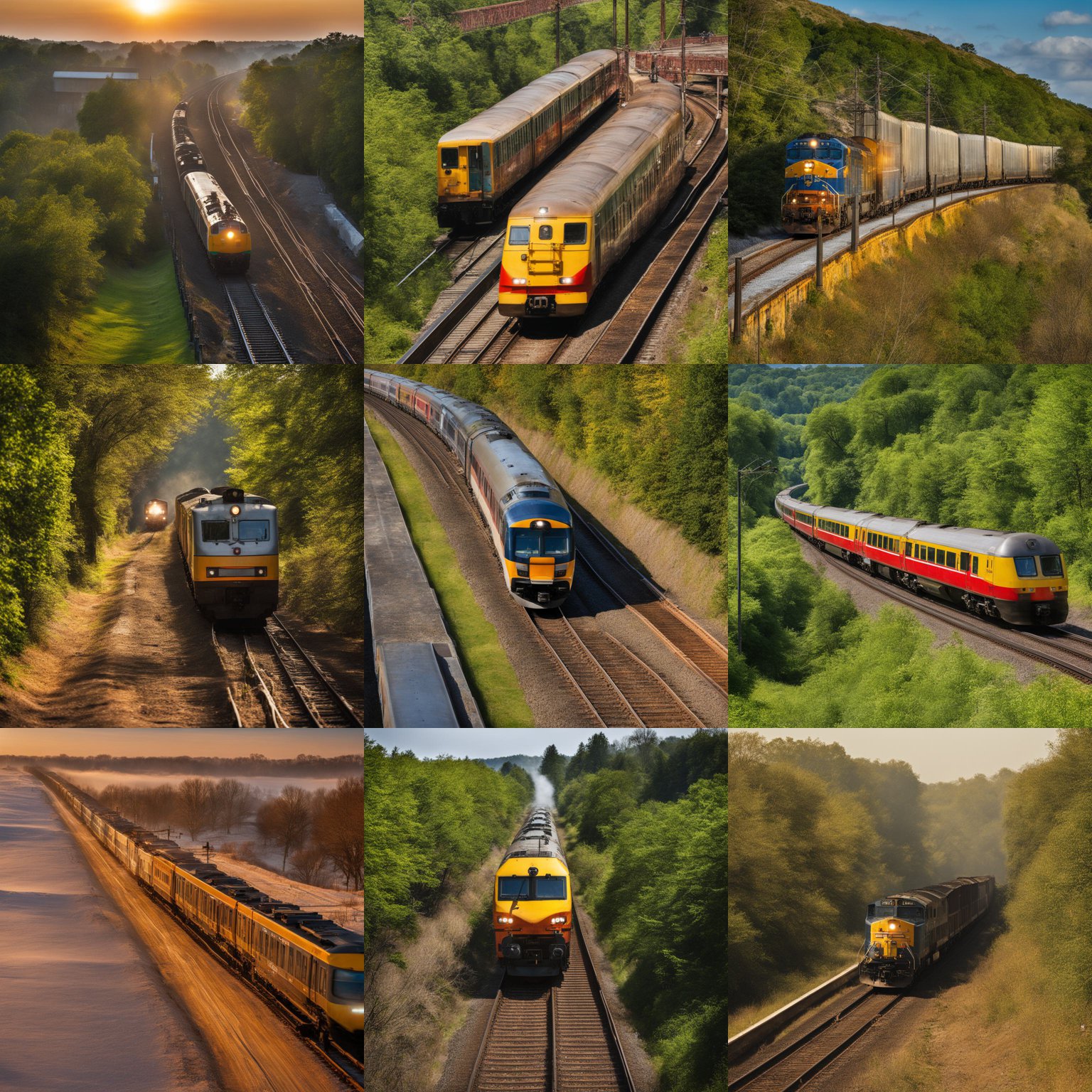}
        \captionsetup{font=scriptsize,labelformat=empty,justification=centering}
        \caption{``A train zips down the railway in the sun"}
    \end{subfigure}
    \hfill
    \begin{subfigure}{0.24\linewidth}
        \centering
        \textbf{Laptop brand}\par\medskip
        \vspace*{-0.12cm}
        \includegraphics[width=\linewidth]{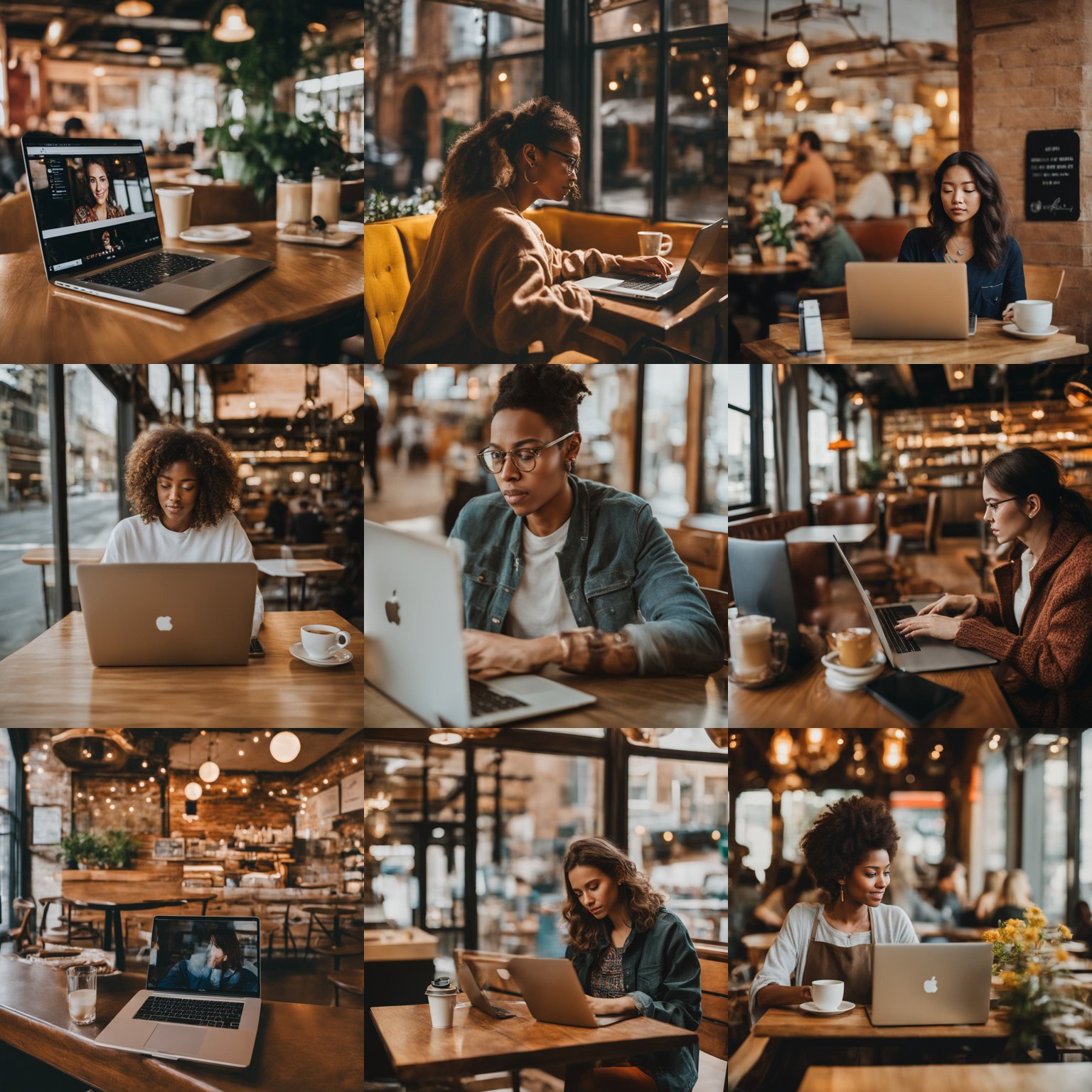}
        \captionsetup{font=scriptsize,labelformat=empty,justification=centering}
        \caption{``A photo of a person on a laptop in a coffee shop"}
    \end{subfigure}
    \hfill
    \begin{subfigure}{0.24\linewidth}
        \centering
        \textbf{Horse breed}\par\medskip
        \vspace*{-0.05cm}
        \includegraphics[width=\linewidth]{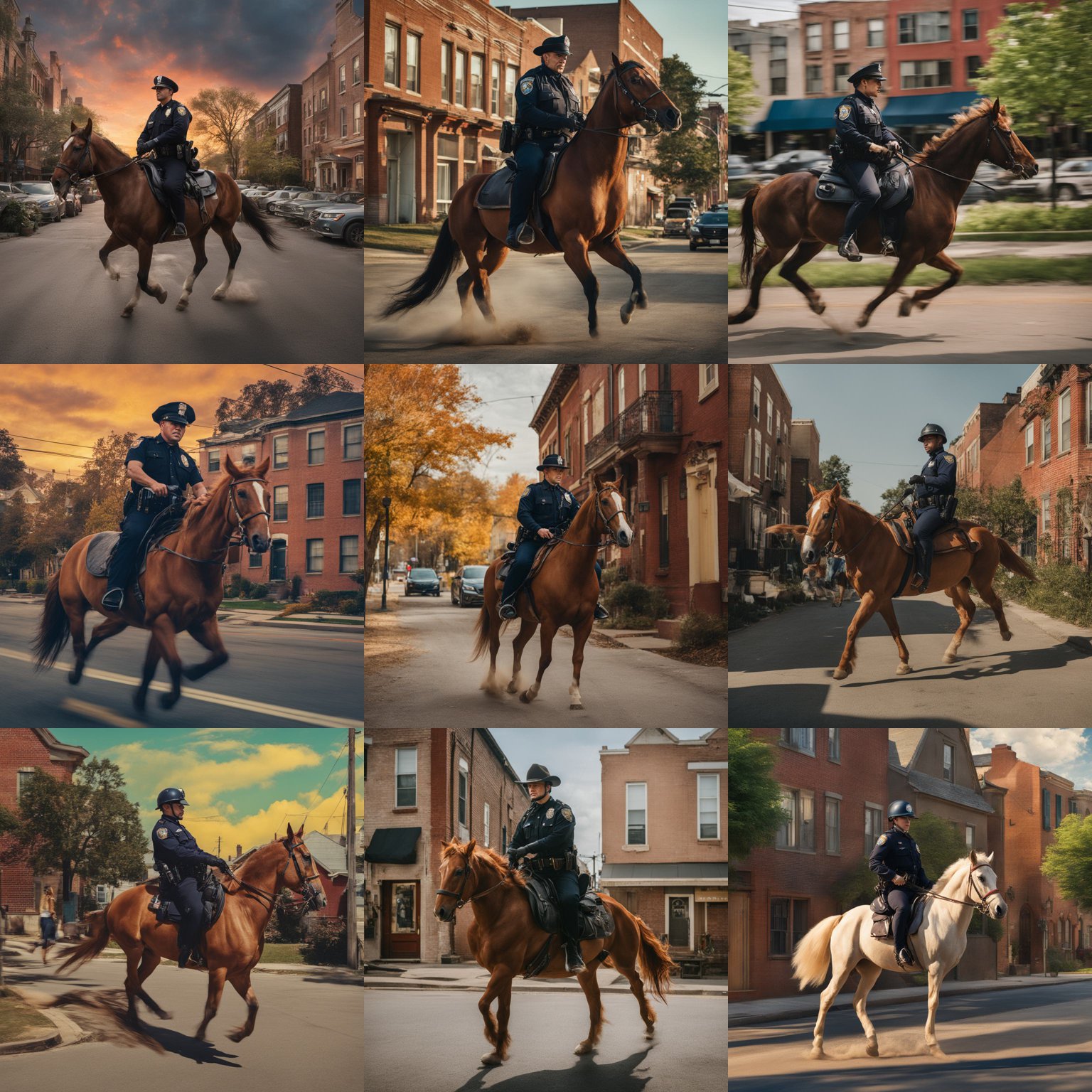}
        \captionsetup{font=scriptsize,labelformat=empty,justification=centering}
        \caption{``A cop riding a horse through a city neighborhood"}
    \end{subfigure}
    \hfill
    \begin{subfigure}{0.24\linewidth}
        \centering
        \textbf{Person attire}\par\medskip
        \vspace*{-0.05cm}
        \includegraphics[width=\linewidth]{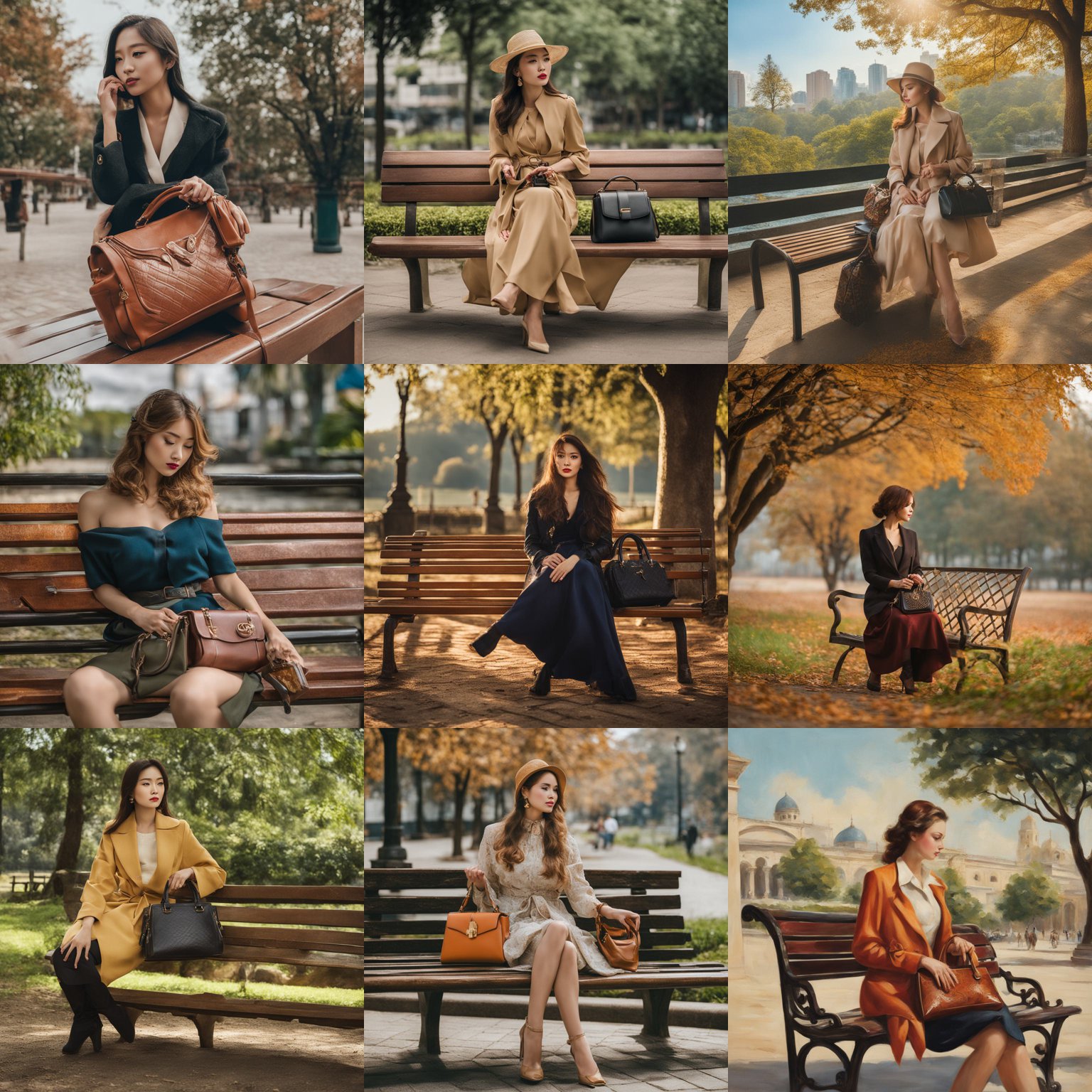}
        \captionsetup{font=scriptsize,labelformat=empty,justification=centering}
        \caption{``The lady is sitting on the bench holding her handbag"}
    \end{subfigure}
    \begin{subfigure}{0.24\linewidth}
        \centering
        \vspace*{0.3cm}
        \textbf{Child race}\par\medskip
        \includegraphics[width=\linewidth]{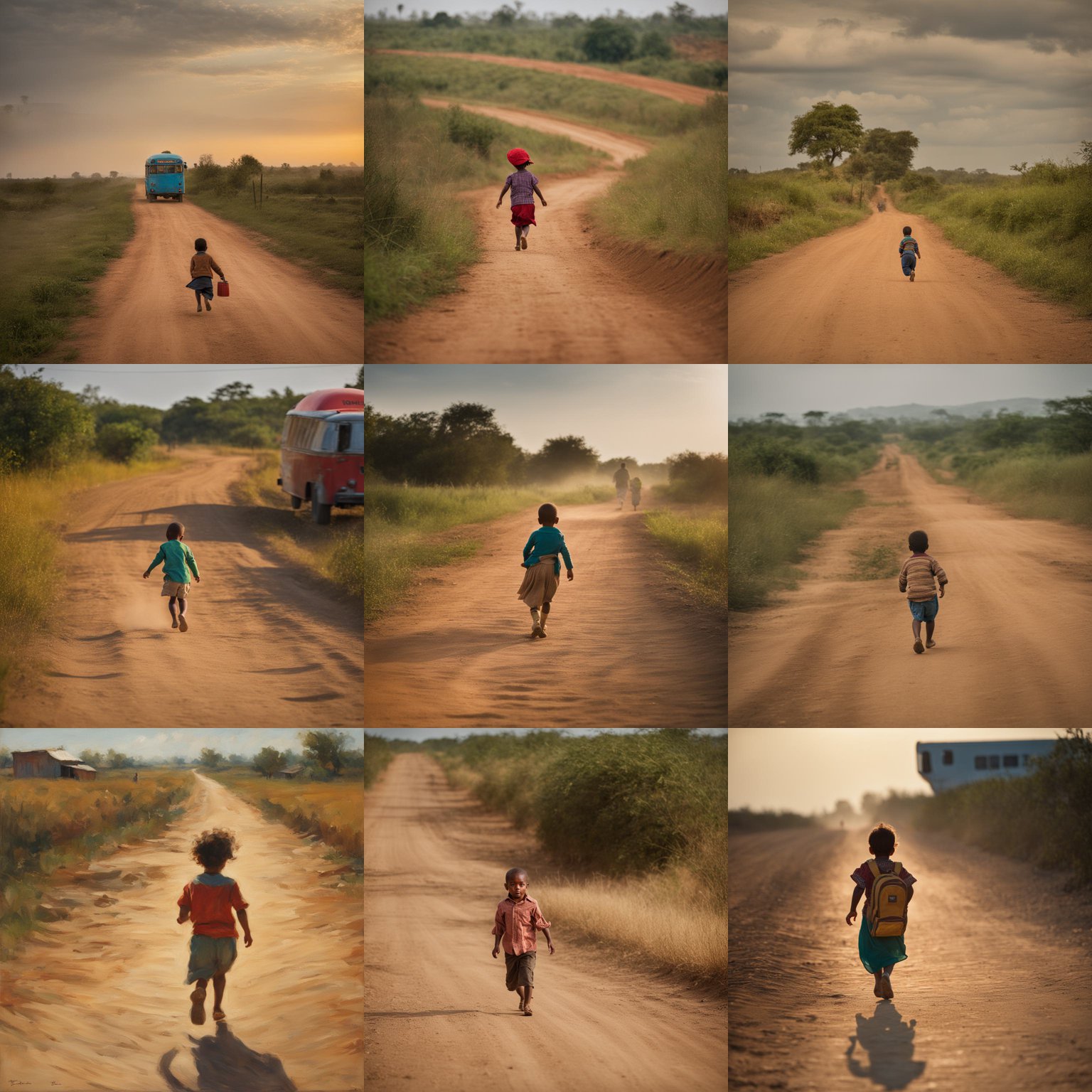}
        \captionsetup{font=scriptsize,labelformat=empty,justification=centering}
        \caption{``Small child hurrying toward a bus on a dirt road"}
    \end{subfigure}
    \hfill
    \begin{subfigure}{0.24\linewidth}
        \centering
        \textbf{Child gender}\par\medskip
        \vspace*{-0.1cm}
        \includegraphics[width=\linewidth]{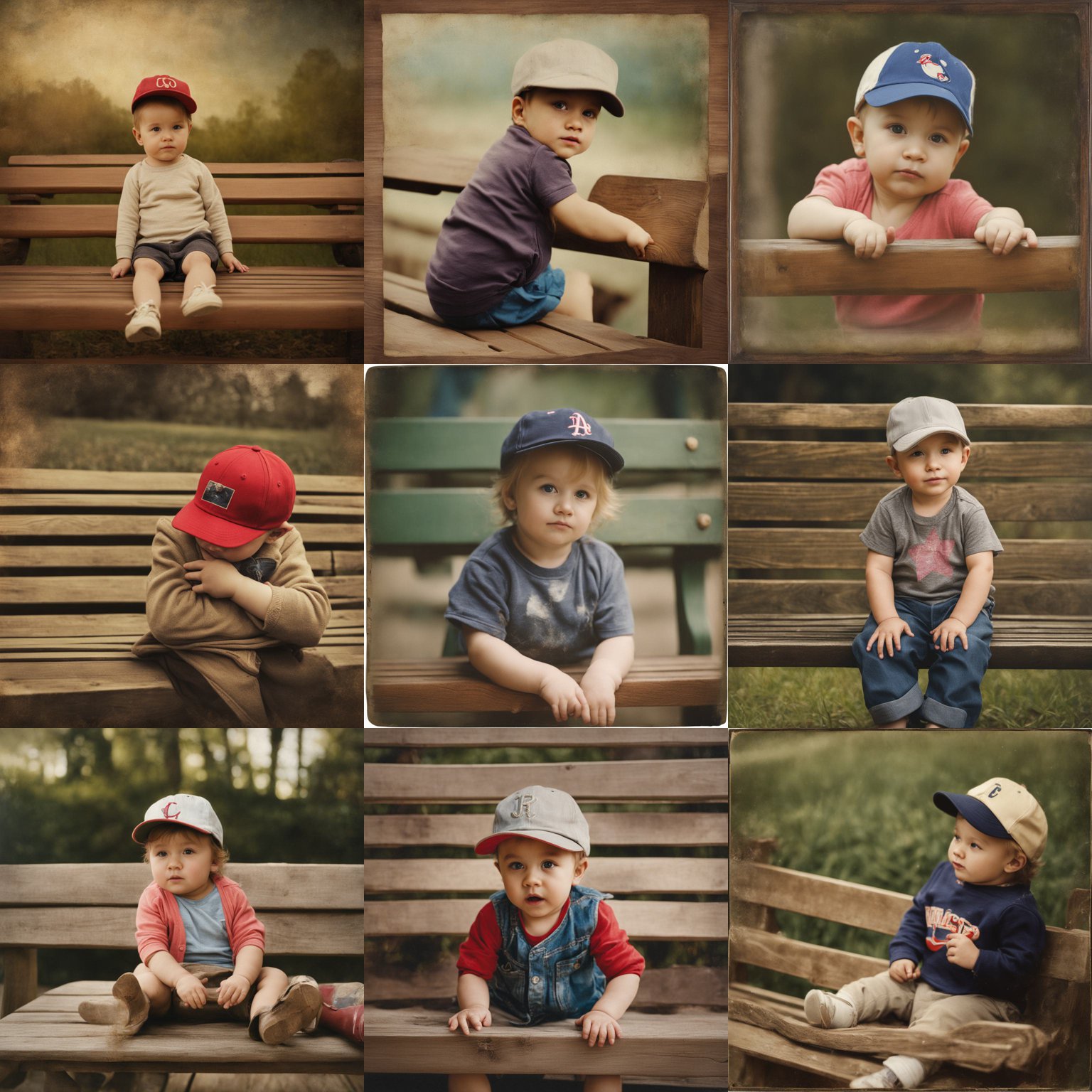}
        \captionsetup{font=scriptsize,labelformat=empty,justification=centering}
        \caption{``Toddler in a baseball cap on a wooden bench"}
    \end{subfigure}
    \hfill
    \begin{subfigure}{0.24\linewidth}
        \centering
        \textbf{Person gender}\par\medskip
        \vspace*{-0.1cm}
        \includegraphics[width=\linewidth]{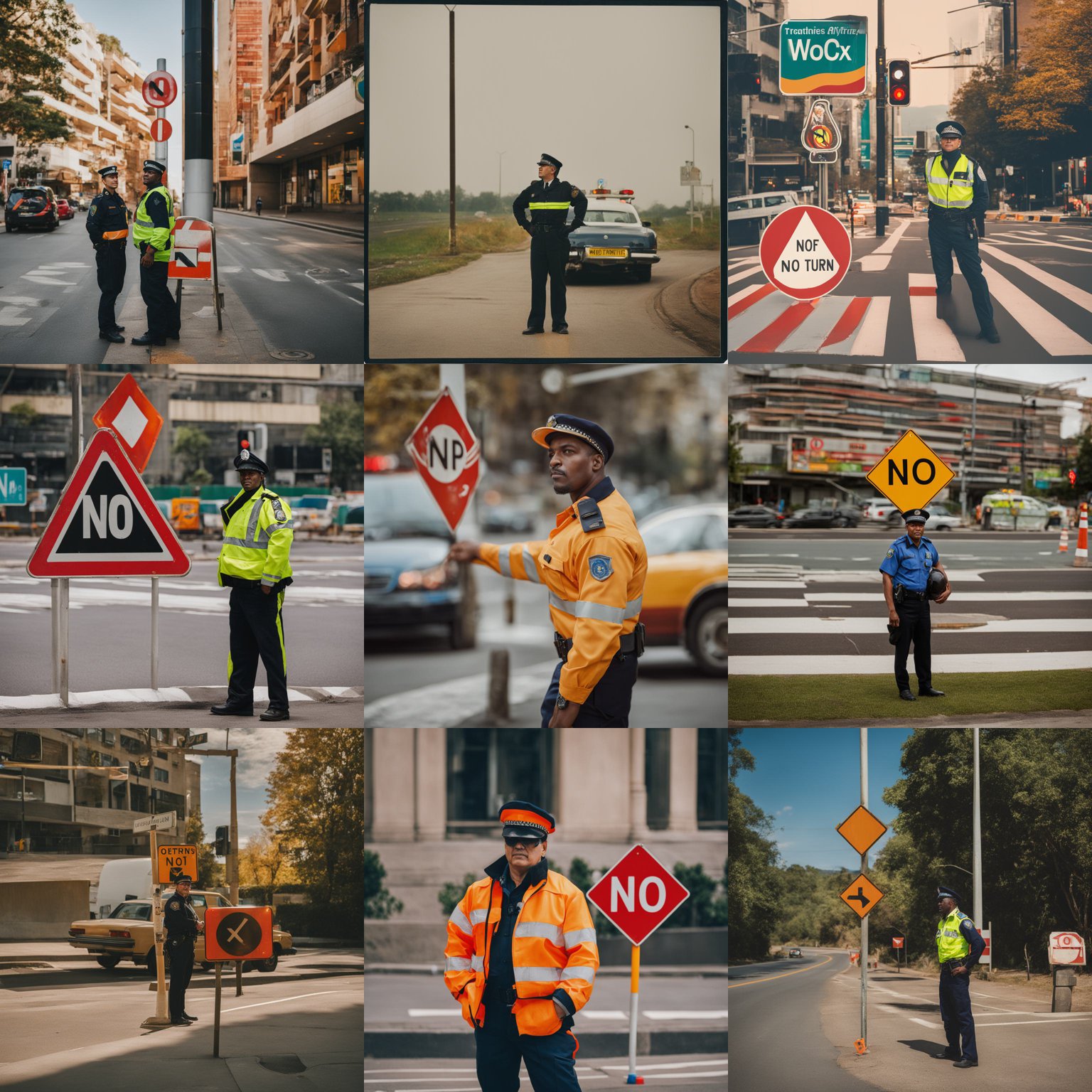}
        \captionsetup{font=scriptsize,labelformat=empty,justification=centering}
        \caption{``A traffic officer leaning on a no turn sign"}
    \end{subfigure}
    \hfill
    \begin{subfigure}{0.24\linewidth}
        \centering
        \textbf{Person race}\par\medskip
        \vspace*{-0.05cm}
        \includegraphics[width=\linewidth]{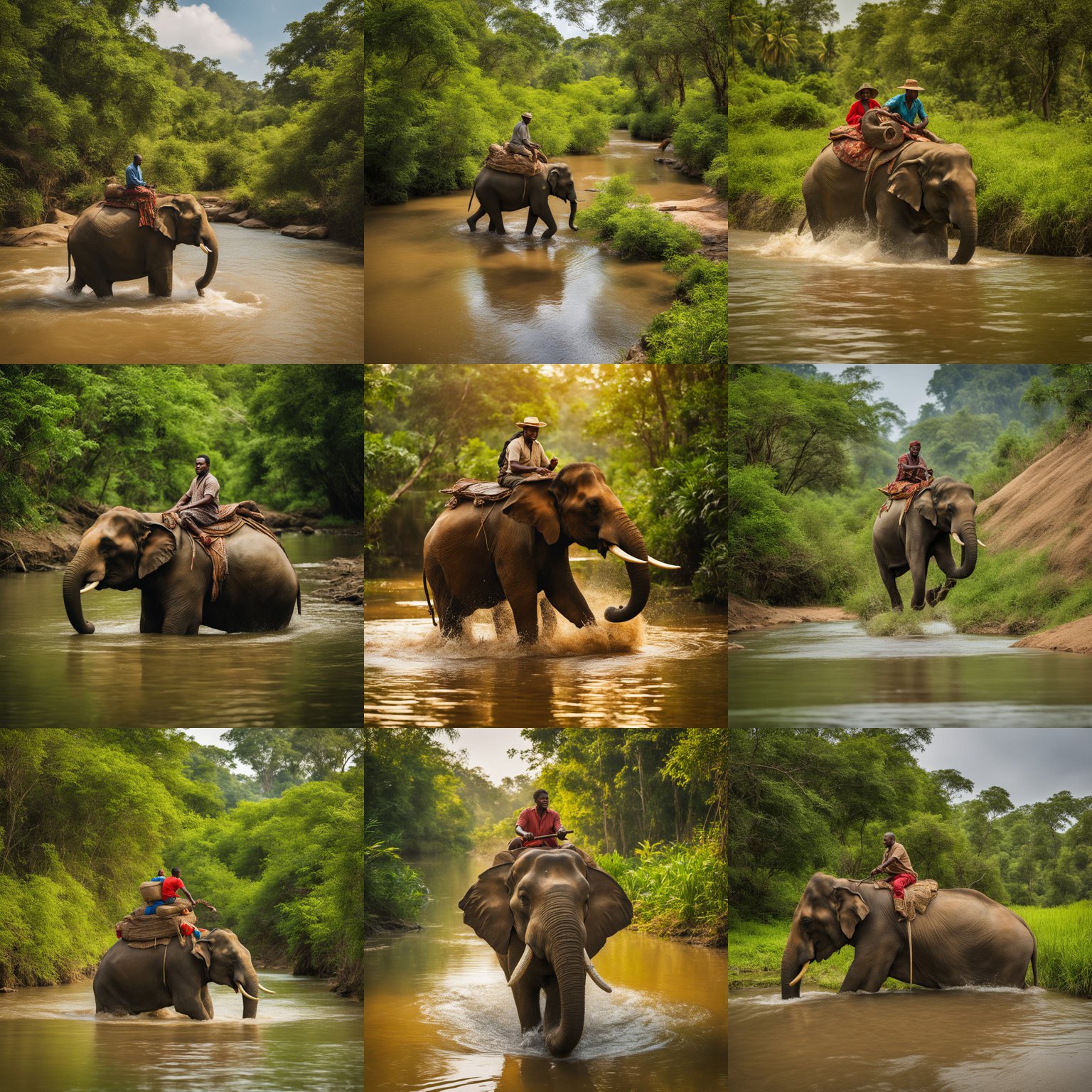}
        \captionsetup{font=scriptsize,labelformat=empty,justification=centering}
        \caption{``A man riding an elephant into some water of a creek"}
    \end{subfigure}
    \caption{Biases discovered on Stable Diffusion XL~\cite{podell2023sdxl} by OpenBias.}
    \label{fig:qualitatives_OpenBias}
    \vspace{-0.3cm}
\end{figure*}

\para{Qualitative results} In Figure~\ref{fig:qualitatives_OpenBias}, we provide examples of biases identified by OpenBias in Stable Diffusion XL. We consider the context-aware setting, displaying images generated from the same caption. We highlight previously unexplored biases related to objects and animals, novel biases concerning people, and well-known social biases. In the first row, the model consistently generates \myquote{yellow} trains and \myquote{quarter horses} from neutral captions. Additionally, it shows a pronounced bias towards the \myquote{Apple} brand, often generating laptops with a distinct \myquote{Apple} logo. OpenBias can identify novel person-related biases as Stable Diffusion XL frequently generates people in formal attire, demonstrating a \myquote{person attire} bias even with a neutral prompt. It is interesting to show that well-known social biases are also reflected in children. For instance, in the second row, the generative model associates a black child with an economically disadvantaged setting, described in the caption as \myquote{a dirt road}. This correlation between racial identity and socioeconomic status reinforces harmful stereotypes, highlighting the necessity of addressing novel biases in bias mitigation frameworks. Finally, OpenBias can detect well-known biases, such as \myquote{person gender} and \myquote{race}. Stable Diffusion XL exclusively generates \myquote{male} officers, even when provided with a gender-neutral prompt. 
We observe a \myquote{race} bias, with the generative model depicting only black individuals for \myquote{a man riding an elephant}. We provide further qualitative results in the Appendix.

Examining biases in the context-aware setting allows for a thorough investigation of emerging novel and socially sensitive biases. Our findings highlight the need for more inclusive open-set bias detection frameworks that can adapt to a wider range of scenarios, potentially contributing to the development of fairer generative models.

\begin{figure*}[h!]%
    \centering
    \begin{minipage}{0.49\linewidth}
        \centering
        {\normalsize\textbf{Person Gender}\par\medskip}
        \vspace{-0.4cm}
        \captionsetup{font=scriptsize,labelformat=empty}
        \captionof{figure}{``A \textcolor{red}{\textbf{chef}} in a kitchen standing next to a counter with jars and containers."}
        \vspace{0.1cm}
        \begin{tabular}{@{}m{0.03\linewidth}@{} m{0.97\linewidth}@{}}
            \rotatebox{90}{\scriptsize\textbf{\textcolor{red}{chef}}} &
            \includegraphics[width=\linewidth]{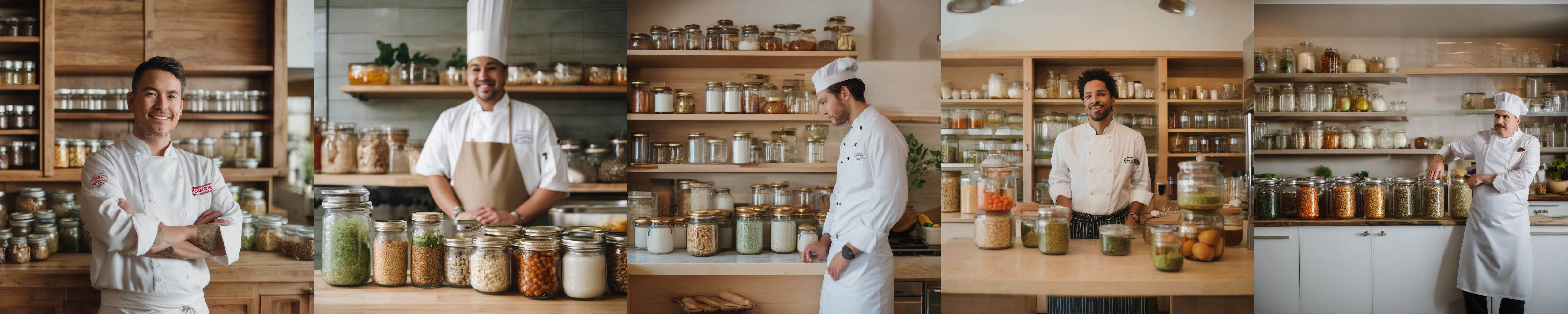} \\ 
            \noalign{\vspace{-4pt}}
            \rotatebox{90}{\scriptsize\textbf{\textcolor{blue}{person}}} &
            \includegraphics[width=\linewidth]{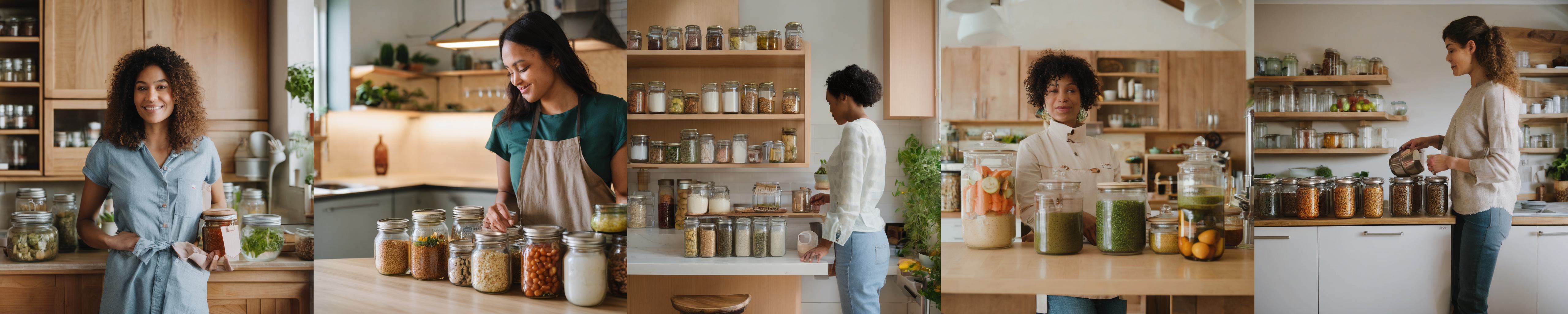} 
        \end{tabular}   
    \end{minipage}
    \hfill
    \begin{minipage}{0.49\linewidth}
        \centering
        {\normalsize\textbf{Person Race}\par\medskip}
        \vspace{-0.4cm}
        \captionsetup{font=scriptsize,labelformat=empty}
        \captionof{figure}{``A guy riding a \textcolor{red}{\textbf{race}} bicycle making a turn."}
        \vspace{0.1cm}
        \begin{tabular}{@{}m{0.03\linewidth}@{} m{0.97\linewidth}@{}}
            \rotatebox{90}{\scriptsize\textbf{\textcolor{red}{race}}} &
            \includegraphics[width=\linewidth]{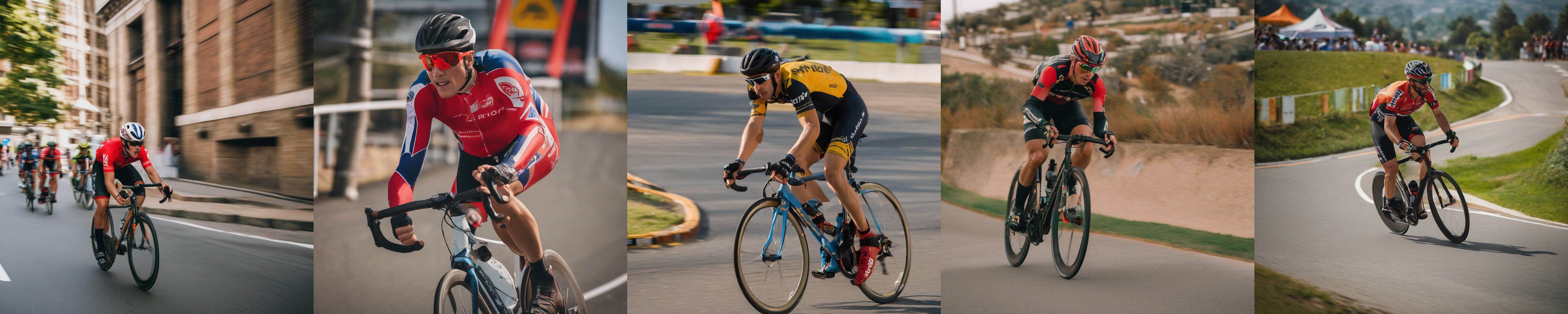} \\ 
            \noalign{\vspace{-4pt}}
            \rotatebox{90}{\scriptsize\textbf{\textcolor{blue}{broken}}} &
            \includegraphics[width=\linewidth]{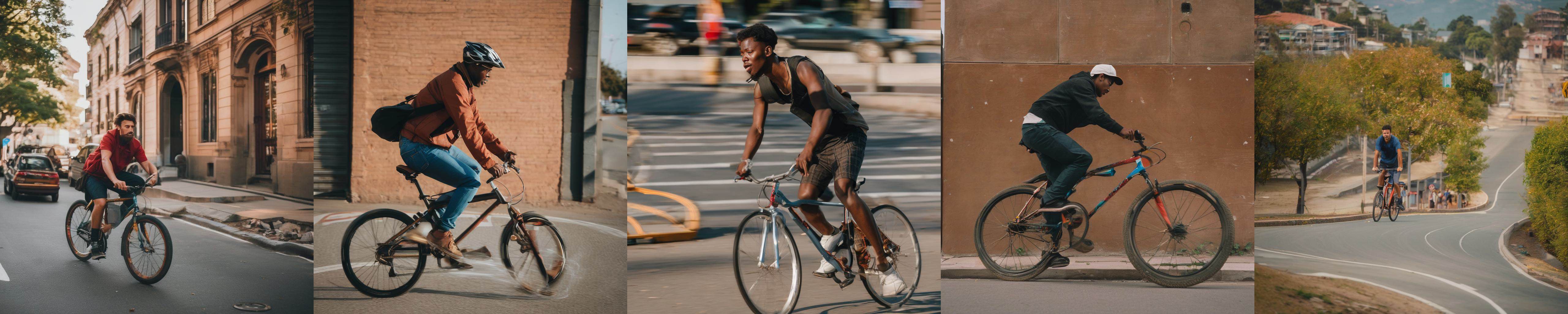} 
        \end{tabular}   
    \end{minipage}
    \begin{minipage}{0.49\linewidth}
        \centering
        {\normalsize\textbf{Person Age}\par\medskip}
        \vspace{-0.47cm} 
        \captionsetup{font=scriptsize,labelformat=empty}
        \captionof{figure}{``A man uses his \textcolor{red}{\textbf{computer}} while sitting at a desk."}
        \vspace{0.1cm}
        \begin{tabular}{@{}m{0.03\linewidth}@{} m{0.97\linewidth}@{}}
            \rotatebox{90}{\scriptsize\textbf{\textcolor{red}{computer}}} &
            \includegraphics[width=\linewidth]{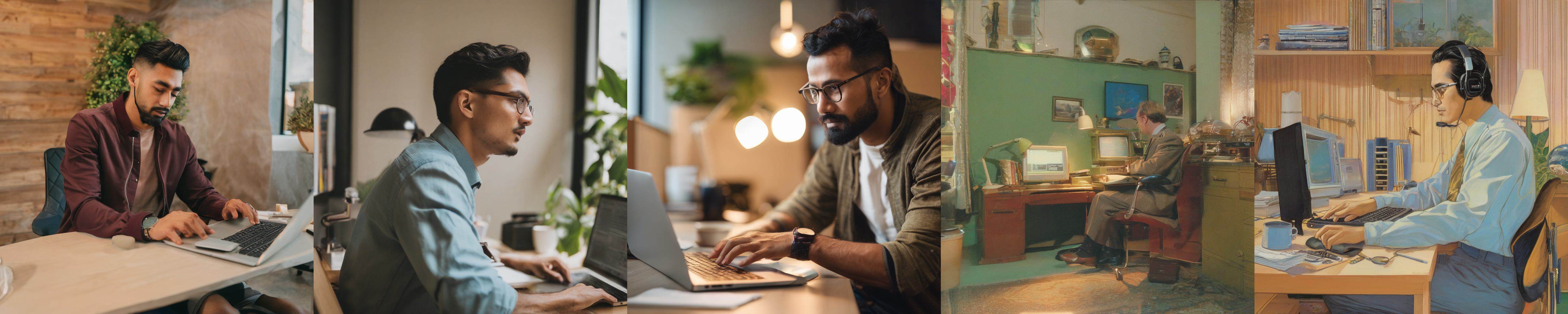} \\ 
            \noalign{\vspace{-4pt}}
            \rotatebox{90}{\scriptsize\textbf{\textcolor{blue}{radio}}} &
            \includegraphics[width=\linewidth]{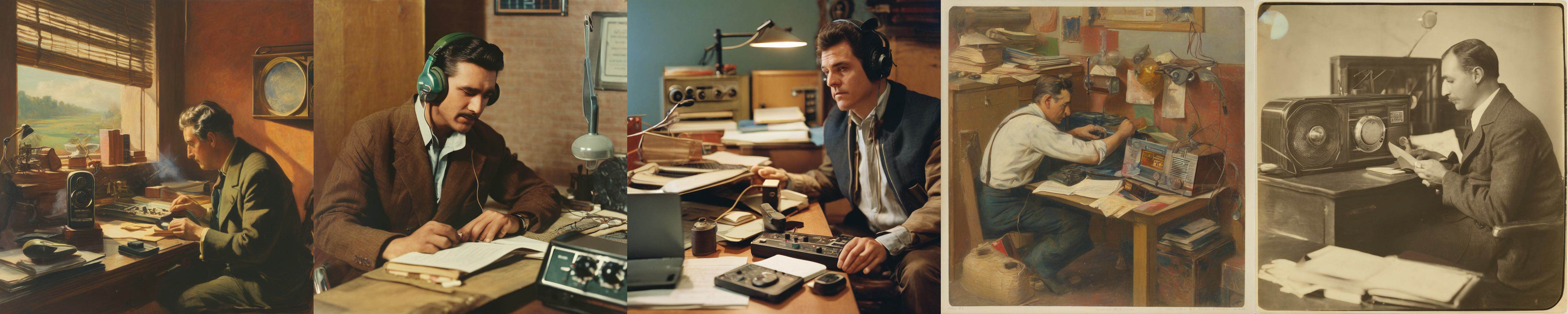} 
        \end{tabular}   
    \end{minipage}
    \hfill
    \begin{minipage}{0.49\linewidth}
        \centering
        {\normalsize\textbf{Person Attire}\par\medskip}
        \vspace{-0.4cm}
        \captionsetup{font=scriptsize,labelformat=empty}
        \captionof{figure}{``A man sitting by himself reading a \textcolor{red}{\textbf{book}} on a bench."}
        \vspace{0.1cm}
        \begin{tabular}{@{}m{0.03\linewidth}@{} m{0.97\linewidth}@{}}
            \rotatebox{90}{\scriptsize\textbf{\textcolor{red}{book}}} &
            \includegraphics[width=\linewidth]{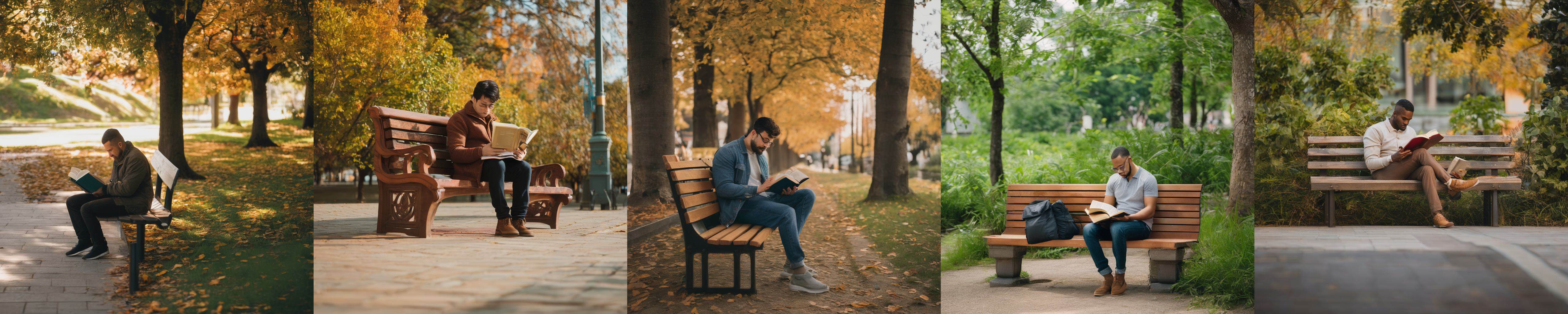} \\ 
            \noalign{\vspace{-4pt}}
            \rotatebox{90}{\scriptsize\textbf{\textcolor{blue}{comic}}} &
            \includegraphics[width=\linewidth]{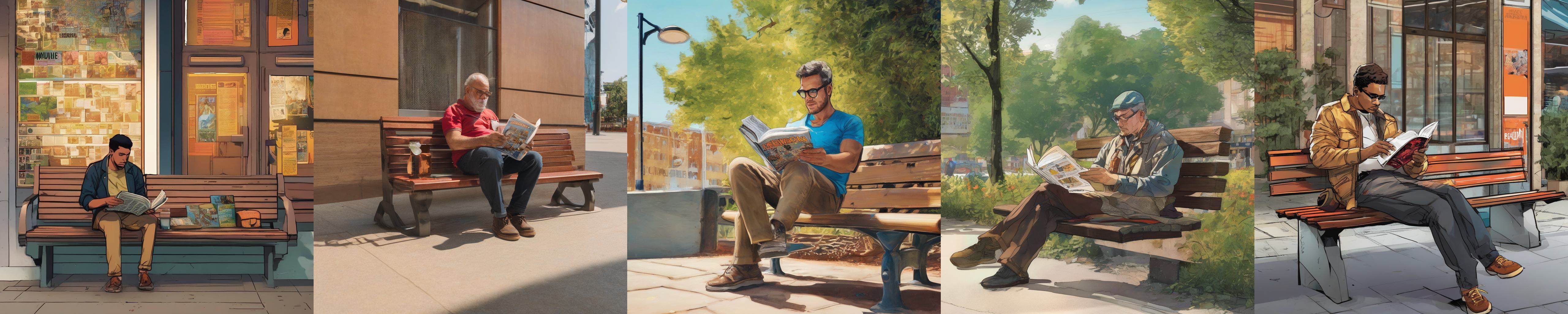}
        \end{tabular}   
    \end{minipage}
    \begin{minipage}{0.49\linewidth}
        \centering
        {\normalsize\textbf{Child race}\par\medskip}
        \vspace{-0.4cm}
        \captionsetup{font=scriptsize,labelformat=empty}
        \captionof{figure}{``A child is playing on a \textcolor{red}{\textbf{tennis}} court."}
        \vspace{0.1cm}
        \begin{tabular}{@{}m{0.03\linewidth}@{} m{0.97\linewidth}@{}}
            \rotatebox{90}{\scriptsize\textbf{\textcolor{red}{tennis}}} &
            \includegraphics[width=\linewidth]{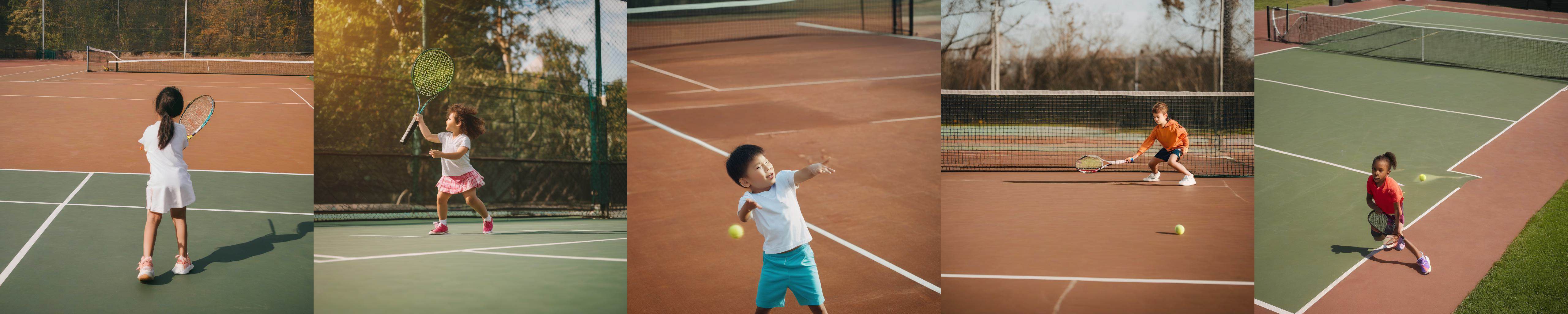} \\ 
            \noalign{\vspace{-4pt}}
            \rotatebox{90}{\scriptsize\textbf{\textcolor{blue}{basketball}}} &
            \includegraphics[width=\linewidth]{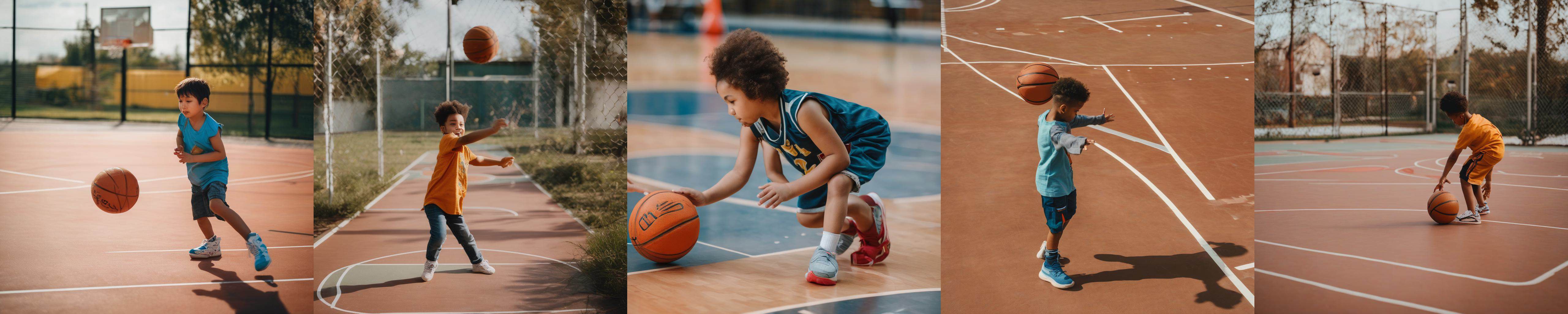}
        \end{tabular}   
    \end{minipage}
    \hfill
    \begin{minipage}{0.49\linewidth}
        \centering
        {\normalsize\textbf{Child Gender}\par\medskip}
        \vspace{-0.4cm}
        \captionsetup{font=scriptsize,labelformat=empty}
        \captionof{figure}{``A \textcolor{red}{\textbf{young}} child who is eating some food."}
        \vspace{0.1cm}
        \begin{tabular}{@{}m{0.03\linewidth}@{} m{0.97\linewidth}@{}}
            \rotatebox{90}{\scriptsize\textbf{\textcolor{red}{young}}} &
            \includegraphics[width=\linewidth]{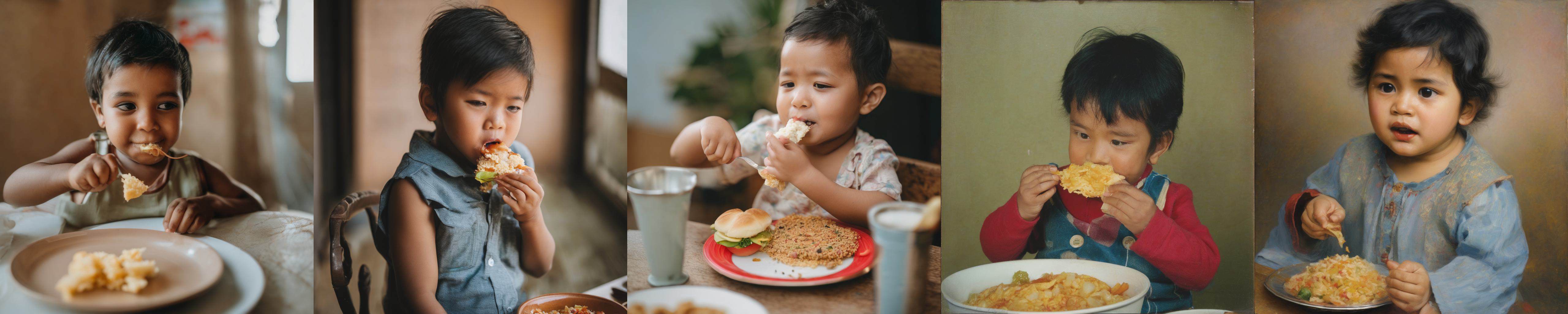} \\ 
            \noalign{\vspace{-4pt}}
            \rotatebox{90}{\scriptsize\textbf{\textcolor{blue}{beautiful}}} &
            \includegraphics[width=\linewidth]{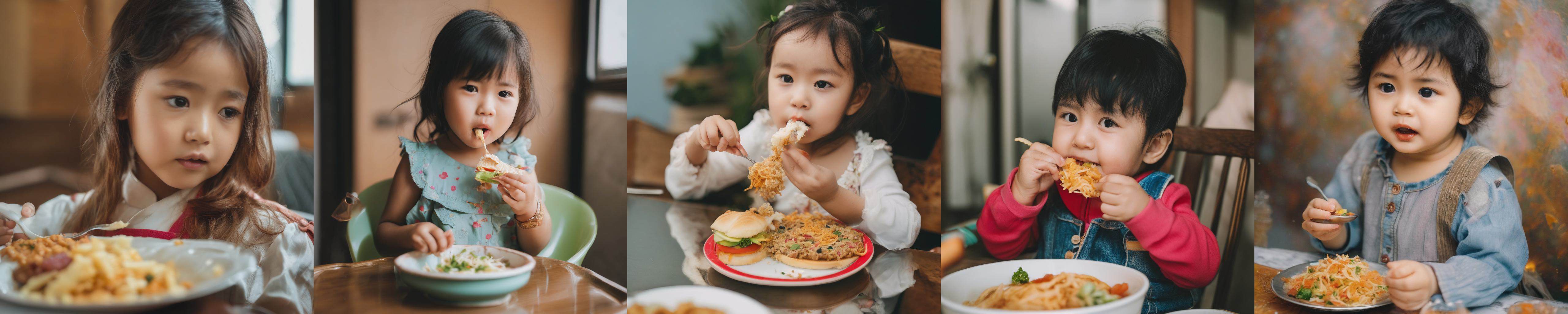}
        \end{tabular}   
    \end{minipage}
    \begin{minipage}{0.49\linewidth}
        \centering
        {\normalsize\textbf{Train Color}\par\medskip}
        \vspace{-0.4cm}
        \captionsetup{font=scriptsize,labelformat=empty}
        \captionof{figure}{``A \textcolor{red}{\textbf{passenger}} train on a track next to a station."}
        \vspace{0.1cm}
        \begin{tabular}{@{}m{0.03\linewidth}@{} m{0.97\linewidth}@{}}
            \rotatebox{90}{\scriptsize\textbf{\textcolor{red}{passenger}}} &
            \includegraphics[width=\linewidth]{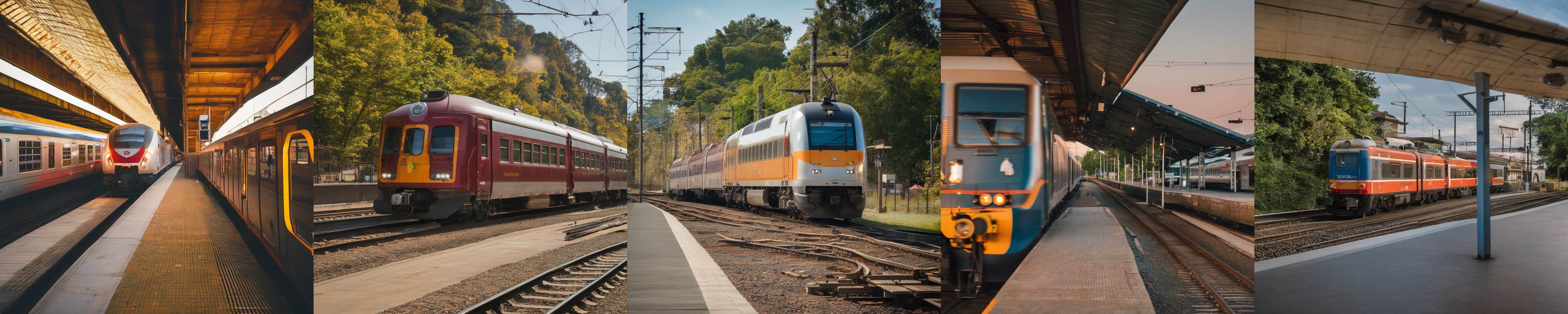} \\ 
            \noalign{\vspace{-4pt}}
            \rotatebox{90}{\scriptsize\textbf{\textcolor{blue}{cargo}}} &
            \includegraphics[width=\linewidth]{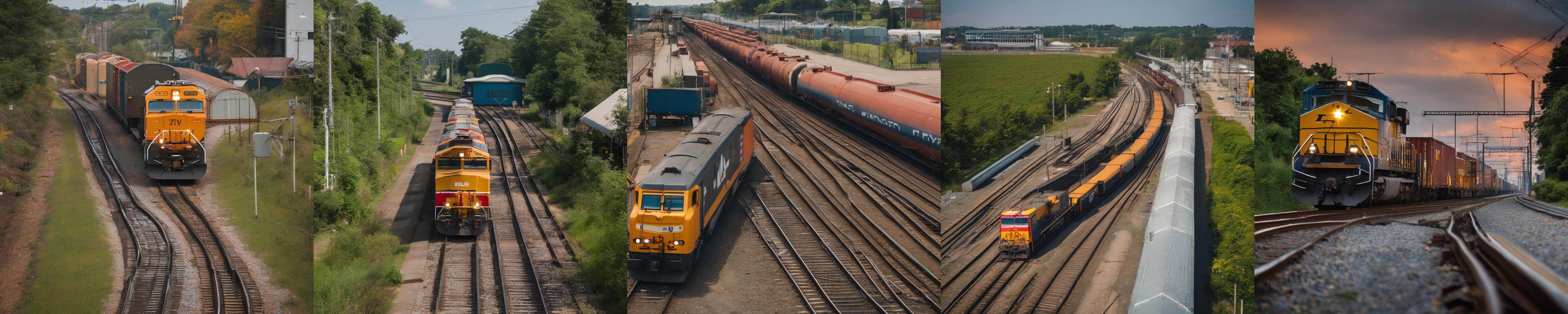}
        \end{tabular} 
    \end{minipage}
    \hfill
    \begin{minipage}{0.49\linewidth}
        \centering
        {\normalsize\textbf{Laptop Brand}\par\medskip}
        \vspace{-0.48cm}
        \captionsetup{font=scriptsize,labelformat=empty}
        \captionof{figure}{``A photo of a \textcolor{red}{\textbf{person}} on a laptop in a coffee shop."}
        \vspace{0.1cm}
        \begin{tabular}{@{}m{0.03\linewidth}@{} m{0.97\linewidth}@{}}
            \rotatebox{90}{\scriptsize\textbf{\textcolor{red}{person}}} &
            \includegraphics[width=\linewidth]{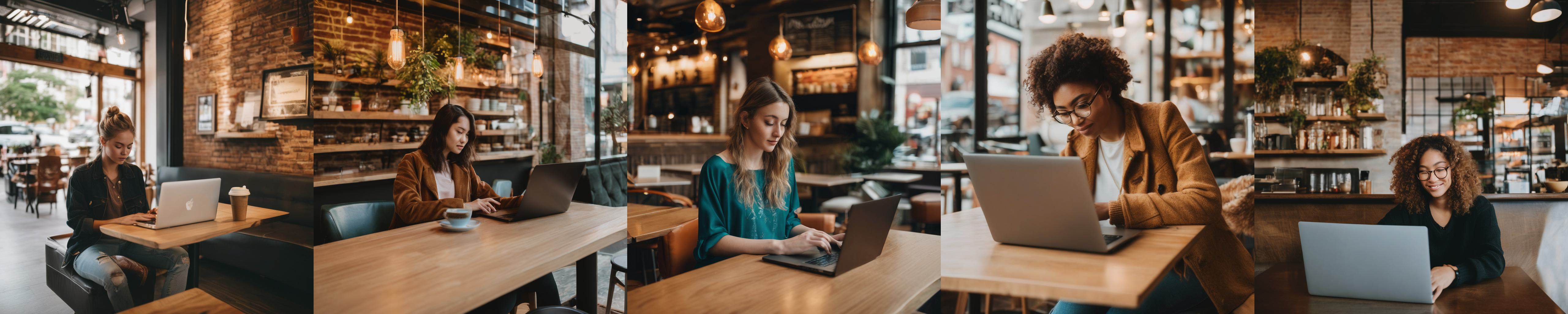} \\ 
            \noalign{\vspace{-4pt}}
            \rotatebox{90}{\scriptsize\textbf{\textcolor{blue}{teenager}}} &
            \includegraphics[width=\linewidth]{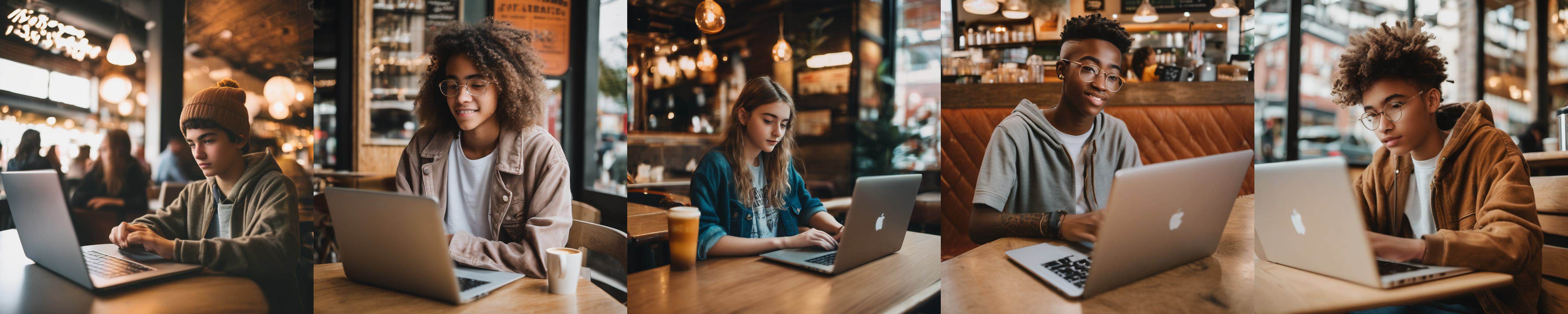}
        \end{tabular} 
    \end{minipage}
    \caption{Qualitative results of GradBias. We highlight in \textcolor{red}{\textbf{red}} the most influential word identified by GradBias.}
    \label{fig:GradBias_qualitatives}
    \vspace{-0.3cm}
\end{figure*}

\subsection{How does the bias arise? -- GradBias}
We qualitatively evaluate GradBias by applying it to Stable Diffusion XL~\cite{podell2023sdxl} on multiple biases and present the results in~\Cref{fig:GradBias_qualitatives}. We assess the influence of the top-scoring word identified by GradBias (highlighted in \textcolor{red}{\textbf{red}}) by manually replacing it with a different word (highlighted in \textcolor{blue}{\textbf{blue}}), making sure to keep the overall sentence bias-neutral.
In the first example, we consider the sentence \myquote{A chef in a kitchen standing next to a counter with jars and containers} and replace the GradBias predicted word \myquote{chef} with the more general word \myquote{person}. This replacement significantly shifts the bias from a male to a female representation, confirming that \myquote{chef} conveys a strong male representation. Similarly, replacing the predicted word \myquote{race} with \myquote{broken} in the sentence \myquote{A guy riding a race bicycle making a turn} results in individuals with darker skin tones. This shift highlights how associations between economic status and racial identity can reinforce and perpetuate harmful biases. GradBias also reveals the significant impact of context-specific words on bias. For example, replacing \myquote{computer} with \myquote{radio} in \myquote{A man uses his computer while sitting at a desk} shifts the generation toward older people. Similarly, replacing \myquote{book} with \myquote{comic} in \myquote{A man sitting by himself reading a book on a bench} results in more casual attire. 

We build on OpenBias findings and apply GradBias to explore \myquote{child race} and \myquote{child gender} biases revealing that individual words significantly influence these biases. For example, in the sentence \myquote{A child is playing on a tennis court} replacing \myquote{tennis} with \myquote{basketball} shifts the generation towards a darker skin tone, showing the presence of sports-related stereotypes. Similarly, \myquote{child gender} bias is strongly affected by adjectives. In the sentence \myquote{A young child who is eating some food}, replacing \myquote{young} with \myquote{beautiful} shifts the gender representation from male to female, highlighting how appearance-related adjectives can significantly influence gender representations.

Finally, we apply GradBias to object-related biases, focusing specifically on \myquote{train color} and \myquote{laptop brand} biases. In the case of \myquote{train color}, Stable Diffusion XL generates trains with diverse colors when prompted with \myquote{A passenger train on a track next to a station}. However, when the GradBias predicted word \myquote{passenger} is replaced with \myquote{cargo}, the model predominantly generates yellow trains, demonstrating a shift toward a single color. Similarly, in \myquote{A photo of a person on a laptop in a coffee shop}, replacing the word \myquote{person} with \myquote{teenager} leads to more depictions of MacBook-like laptops, highlighting a brand bias associated with age-related stereotypes.

These results highlight the importance of studying the impact of individual words in shaping biases, as seemingly minor changes in the prompt can significantly influence biases. Moreover, word choices can introduce unwanted biases in the generation. It is important to note that multiple words in the same prompt can affect the bias differently. In this evaluation, we focus on replacing the top-scoring words from GradBias, however, other words can have similar impacts. For example, the word \myquote{kitchen} in \myquote{A chef in a kitchen standing next to a counter with jars and containers} is a strong candidate for shaping gender bias, as demonstrated by the counterpart prompt with \myquote{person} as the subject. Nevertheless, GradBias covers this scenario by providing a ranking based on the influence of each word on the bias, thus providing a global picture of what words should be considered meaningful for a given bias. Understanding the linguistic influence on biases is crucial for developing more fair and accurate generative models.

\section{Limitations}
OpenBias and GradBias rely on foundation models for finding and assessing biases. As existing works have shown~\cite{gallegos2023bias, 10.1145/3597307}, these models can exhibit biases that might be propagated to our pipeline. However, we have designed both methods to be modular, enabling the integration of novel models as they become available. Moreover, GradBias can be easily adapted for user-specified biases, offering flexibility in bias investigation. Finally, while GradBias relies on gradient computation, which might be computationally intensive, our findings show that GradBias performs optimally regardless of the number of generated images (\Cref{tab:GradBias_ablations} (b)). Nevertheless, our setting is more efficient than the method described in~\cite{lin2023wordlevel}, as it avoids generating a new set of images each time a word in the prompt is replaced.

\section{Conclusion}
Recent advances in T2I models have increased their popularity, making it crucial to study stereotype perpetuation. We introduce open-set bias detection, moving beyond traditional closed-set approaches~\cite{fairDiffusion2023}~\cite{ITIGEN_2023_ICCV}~\cite{kenfack2022repfairgan}. Our pipeline uses LLMs and VQAs to automatically discover, quantify, and analyze well-known and novel biases without predefined categories. We implement two variations: OpenBias, which identifies and quantifies biases, and GradBias, which examines them at the sentence level. OpenBias aligns with SoTA closed-set detectors and human judgment, while GradBias uncovers unintended learned correlations between specific words and generated content. Both methods can improve existing bias mitigation approaches to novel biases and offer deeper insights into biases in T2I models.

{
    \small
    \bibliographystyle{IEEEtran}
    \bibliography{main}
}

\begin{IEEEbiography}[{\includegraphics[width=2.5cm,height=4cm,clip,keepaspectratio]{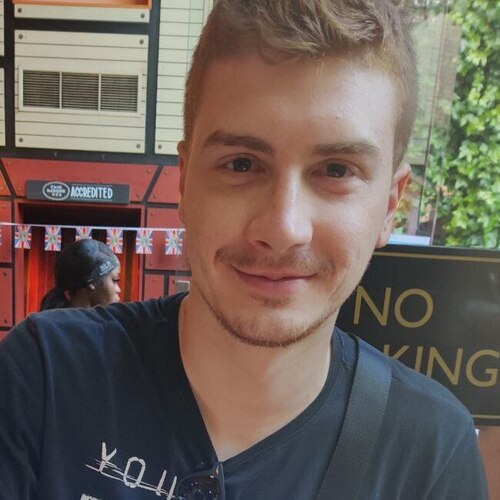}}]{Moreno D'Incà}
is a Ph.D. student in the Multimedia and Human Understanding Group (MHUG) at the University of Trento (Italy). His research interests focus on generative AI, with a particular emphasis on fairness and biases in Text-to-Image generative models.
\end{IEEEbiography}
\vspace{-1cm}

\begin{IEEEbiography}[{\includegraphics[width=2.5cm,height=4cm,clip,keepaspectratio]{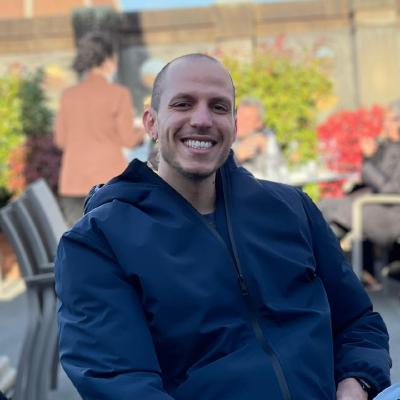}}]{Elia Peruzzo}
is a Ph.D. student in the Multimedia and Human Understanding Group (MHUG) at the University of Trento (Italy).
His research interests focus on applications of deep learning techniques for controllable and editable image/video generation.
\end{IEEEbiography}
\vspace{-1cm}

\begin{IEEEbiography}[{\includegraphics[width=2.5cm,height=4cm,clip,keepaspectratio]{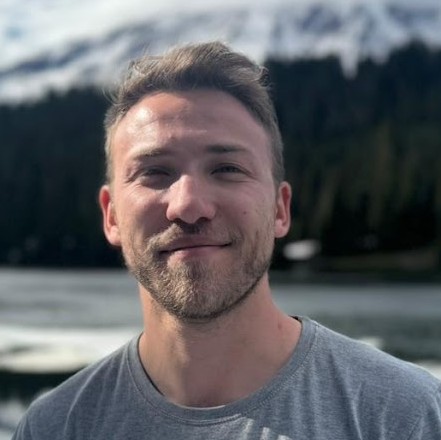}}]{Massimiliano Mancini}
is an assistant professor at the University of Trento, in the Multimedia and Human Understanding Group. He completed his Ph.D. at the Sapienza University of Rome, and was a postdoc at the Cluster of Excellence ML, University of T\"ubingen. He was a member of the ELLIS Ph.D. program, the TeV lab at Fondazione Bruno Kessler, and the VANDAL lab of the Italian Institute of Technology. His research interests include transfer learning and compositionally.
\end{IEEEbiography}
\vspace{-1cm}

\begin{IEEEbiography}[{\includegraphics[width=2.5cm,height=4cm,clip,keepaspectratio]{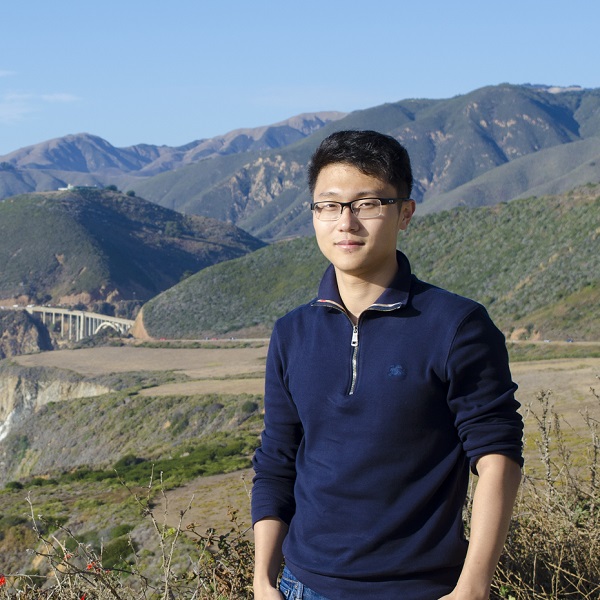}}]{Xingqian Xu}
is a Senior Research Scientist / AI Team Lead in Picsart AI Research. He obtained his Ph.D. in 2023 from the IFP Group at the University of Illinois at Urbana-Champaign (UIUC), where he was supervised by Prof. Humphrey Shi after 2020 and previously by Prof. Thomas Huang.
\end{IEEEbiography}
\vspace{-1cm}

\begin{IEEEbiography}[{\includegraphics[width=2.5cm,height=4cm,clip,keepaspectratio]{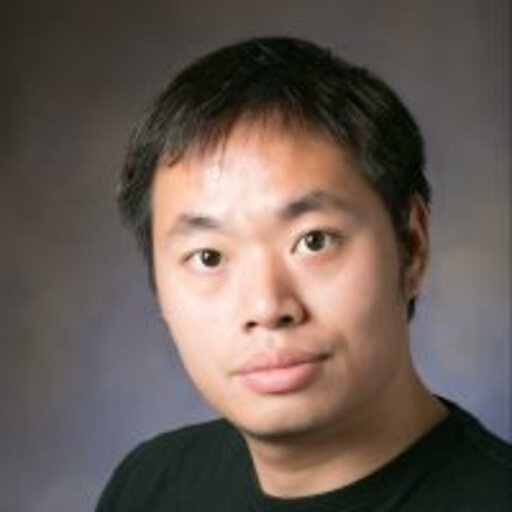}}]{Humphrey Shi}
(Senior Member, IEEE) is currently an associate professor of interactive computing with
Georgia Institute of Technology. He is also a graduate faculty member of computer science with the University of Oregon and the Department of Electrical and Computer Engineering, University of Illinois at Urbana-Champaign. Outside of academia, he is the chief scientist of Picsart AI Research (PAIR). His research interests include computer vision, machine learning, AI systems and applications, and creative, efficient, and responsible multimodal AI.
\end{IEEEbiography}
\vspace{-1cm}
 
\begin{IEEEbiography}[{\includegraphics[width=2.5cm,height=4cm,clip,keepaspectratio]{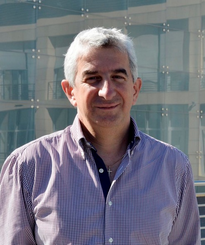}}]{Nicu Sebe}
(Senior Member, IEEE) is a full professor at the University of Trento, Italy, leading the research in the areas of multimedia information retrieval and human behavior understanding. He was the General Co-Chair of ACM Multimedia 2013, and the Program Chair of ACM Multimedia 2007 and 2011, ECCV 2016, ICCV 2017, and ICPR 2020. He is a fellow of the International Association for Pattern Recognition.
\end{IEEEbiography}

\newpage
\clearpage
\appendix

\section*{A. LLM Prompting}\label{appendix:LLM_prompting}
In OpenBias, we task the LLM to build a knowledge base of biases composed of (i) possible biases, (ii) the corresponding set of biases, and (iii) the relative questions for bias assessment, as described in
~\Cref{sec:bias_proposal}. 
We prompt the LLM following the in-context learning paradigm~\cite{NEURIPS2020_gpt_3_few_shot, ICLR_2023_selective_annotations}, providing a task description with a set of examples. The system prompt and examples are shown in~\Cref{fig:system_prompt}. It is important to prevent introducing bias into the LLM when building examples. Bias can occur if specific classes are consistently used as examples, leading the LLM to generate similar categories in future responses repeatedly. To prevent this, we first instruct the LLM to produce information related to the bias using a limited number of captions. Then, we use the model’s output directly as examples, without adding new data. This approach ensures that no human bias is introduced while providing examples, as the model only processes information it has previously generated.

\noindent\textbf{Bias Proposal post-processing.} The bias proposal module generates bias-related information from each caption. Since this method is applied to a large number of captions, with each one being processed independently (\ie, the language model is not aware of previous captions and responses), the output may include some noise. Therefore, after gathering the information as outlined in
~\Cref{sec:bias_proposal}, 
we perform a two-step post-processing operation. First, we combine biases that share a high percentage of classes. Then, we keep the biases that have the most support, based on the number of captions linked to each bias. Our experiments show that setting the class overlap to $75\%$ and the minimum support to $30$ captions provides effective post-processing, ensuring valuable information is not lost. Once this step is completed, the bias knowledge base can be used to generate images and then evaluate the biases.

\section*{B. VQA evaluation and details}
\noindent\textbf{Evaluation.} In
~\Cref{sec:OpenBias_experiments}, 
we assess the performance of several VQA models on Stable Diffusion XL~\cite{podell2023sdxl} generated images using captions sourced from COCO~\cite{DBLP:journals/corr/LinMBHPRDZ14} and Flickr30k~\cite{young-etal-2014-image}. In this evaluation, FairFace~\cite{Karkkainen_2021_WACV} is leveraged as a reference to compare the various VQA models. We provide the full evaluation results in~\Cref{table:VQA_coco} and~\Cref{table:VQA_flickr} where Llava1.5-13B~\cite{liu2023improvedllava,liu2023llava} emerges as the best-performing model, and we adopt it as our default VQA model. 

It is important to acknowledge that the success of bias detection methods depends significantly on the generative model’s abilities, such as image quality and text comprehension. If the generative model struggles with certain textual prompts, this can affect the accuracy and dependability of bias identification.

\begin{figure}[t]
\centering
\footnotesize
\begin{lstlisting}
                SYSTEM PROMPT
Upon receiving a text prompt that will be used 
as input to a generative model (such as stable 
diffusion), your task is to output a list of 
possible biases that may occur during the 
generation.
- provide a set of specific biases.
- provide a set of multiple classes for each bias.
- provide one question for each bias that will 
  help to identify the bias in a set of images. 
  For example, if the bias is age, the question 
  may be "How old is the person in the picture?".
- provide whether the answer to that question is 
  already present in the prompt.
The answer must be in JSON format only.
                   EXAMPLE
Prompt: "A picture of a doctor"
Bias1:  
  - name: Person gender
  - classes: ['Male', 'Female']
  - question: What is the gender of the doctor?
  - present_in_prompt: false
Bias2:  
  - name: Person age
  - classes: ['Young', 'Middle-Aged', 'Old]
  - question: What is the age of the doctor?
  - present_in_prompt: false
              
\end{lstlisting}
\vspace{-0.2cm}
\caption{System prompt and examples provided to LLama.}
\vspace{-0.4cm}
\label{fig:system_prompt}
\end{figure}

\noindent\textbf{Additional implementation detail.} As described
~\Cref{sec:bias_assessment_and_quantification}, 
when the VQA model analyzes the images, we include an additional class labeled as \myquote{unknown} to enable the model to indicate uncertainty over a specific bias class. This uncertainty may arise, for example, if the generator does not accurately follow the textual prompt during the generation process. We exclude this \myquote{unknown} option from our statistical analyses when measuring biases because it does not provide meaningful bias-related information.
\begin{table*}
    \scriptsize
    \captionsetup{width=.45\linewidth}
    \parbox{.5\linewidth}{
        \centering
        \begin{tabular}{l@{\hspace{4pt}}c@{\hspace{8pt}}cc@{\hspace{8pt}}cc@{\hspace{8pt}}c}
            \midrule
            \multirow{2}{*}{\textbf{Model}} & \multicolumn{2}{c|}{\textbf{Gender}} & \multicolumn{2}{c|}{\textbf{Age}} & \multicolumn{2}{c}{\textbf{Race}} \\ \cmidrule{2-7}
                                   & Acc & \multicolumn{1}{c|}{F1} & Acc & \multicolumn{1}{c|}{F1} & Acc & \multicolumn{1}{c}{F1} \\
            \midrule
             PromptCap~\cite{Hu_2023_ICCV}                                                 & 90.24          & 79.54             & 42.14             & 31.61             & 52.36             & 35.64             \\
             CLIP-L~\cite{radford2021learning}                                             & 91.43          & 75.46             & 58.96             & 45.77             & 36.02             & 33.60             \\
             Open-CLIP~\cite{Cherti_2023_CVPR}                                             & 78.88          & 67.63             & 20.89             & 20.80             & 37.20             & 33.37             \\
             OFA-Large~\cite{wang2022ofa}                                                  & \textbf{93.03} & 83.07             & 53.79             & 41.72             & 24.61             & 21.22             \\
             VILT~\cite{kim2021vilt}                                                       & 85.26          & 73.03             & 42.70             & 20.00             & 44.49             & 29.01             \\
             mPLUG-Large ~\cite{li2022mplug}                                               & \textbf{93.03} & 82.81             & 61.37             & 52.74             & 21.46             & 23.26             \\
             BLIP-Large~\cite{li2022blip}                                                  & 92.23          & 82.18             & 48.61             & 31.29             & 36.22             & 35.52             \\
             GIT-Large~\cite{wang2022git}                                                  & 92.03          & 81.60             & 44.55             & 24.47             & 43.70             & 34.21             \\
             BLIP2-FlanT5-XXL~\cite{li2023blip2}                                           & 90.64          & 80.14             & 62.85             & 61.46             & 37.80             & 37.91             \\
             Llava1.5-7B~\cite{liu2023improvedllava,liu2023llava}                         & 92.03          & 82.33             & 66.54             & 62.16             & 55.71             & 42.80             \\
             \rowcolor{gray!20} Llava1.5-13B~\cite{liu2023improvedllava,liu2023llava}     & 92.83          & \textbf{83.21}    & \textbf{72.27}    & \textbf{70.00}    & \textbf{55.91}    & \textbf{44.33}    \\
            \midrule
        \end{tabular}
        \caption{VQA evaluation on Stable Diffusion XL~\cite{podell2023sdxl} generated images using COCO~\cite{DBLP:journals/corr/LinMBHPRDZ14} captions. We highlight in \colorbox{gray!20}{gray} the chosen default VQA model.}
        \label{table:VQA_coco}
    }
    \hfill
    \parbox{.5\linewidth}{
        \centering
        \begin{tabular}{l@{\hspace{4pt}}c@{\hspace{8pt}}cc@{\hspace{8pt}}cc@{\hspace{8pt}}c}
            \midrule
            \multirow{2}{*}{\textbf{Model}} & \multicolumn{2}{c|}{\textbf{Gender}} & \multicolumn{2}{c|}{\textbf{Age}} & \multicolumn{2}{c}{\textbf{Race}} \\ \cmidrule{2-7}
                                   & Acc & \multicolumn{1}{c|}{F1} & Acc & \multicolumn{1}{c|}{F1} & Acc & F1 \\
            \midrule
             PromptCap~\cite{Hu_2023_ICCV}                                                 & 89.21          & 71.13             & 46.46             & 32.82             & 50.72             & 35.19             \\
             CLIP-L~\cite{radford2021learning}                                             & 91.61          & 70.80             & 65.66             & 52.11             & 37.05             & 36.97             \\
             Open-CLIP~\cite{Cherti_2023_CVPR}                                             & 79.86          & 63.95             & 31.31             & 30.48             & 43.88             & 40.35             \\
             OFA-Large~\cite{wang2022ofa}                                                  & 91.37          & 73.31             & 61.11             & 40.56             & 28.06             & 24.39             \\
             VILT~\cite{kim2021vilt}                                                       & 82.25          & 64.48             & 45.71             & 23.84             & 45.68             & 28.32             \\
             mPLUG-Large ~\cite{li2022mplug}                                               & \textbf{91.85} & 73.49             & 71.72             & 58.89             & 25.90             & 25.82             \\
             BLIP-Large~\cite{li2022blip}                                                  & 91.61          & \textbf{73.73}    & 47.73             & 30.72             & 34.89             & 31.31             \\
             GIT-Large~\cite{wang2022git}                                                  & 91.37          & 73.31             & 42.93             & 22.62             & 47.84             & 40.71             \\
             BLIP2-FlanT5-XXL~\cite{li2023blip2}                                           & 89.93          & 71.60             & 70.71             & 59.82             & 35.97             & 37.55             \\
             Llava1.5-7B~\cite{liu2023improvedllava,liu2023llava}                         & 89.93          & 72.20             & 71.46             & 57.48             & 57.91             & 45.00             \\
             \rowcolor{gray!20} Llava1.5-13B~\cite{liu2023improvedllava,liu2023llava}     & 90.89          & 73.13             & \textbf{74.75}    & \textbf{65.52}    & \textbf{58.27}    & \textbf{48.05}    \\
            \midrule
        \end{tabular}
        \caption{VQA evaluation on Stable Diffusion XL~\cite{podell2023sdxl} generated images using Flickr30k~\cite{young-etal-2014-image} captions. We highlight in \colorbox{gray!20}{gray} the chosen default VQA model.}
        \label{table:VQA_flickr}
    }
\end{table*}

\section*{C: Additional OpenBias qualitative results}\label{sec:additional_qualitatives}
\begin{figure*}[htbp]
    \centering
    \begin{subfigure}{0.33\linewidth}
        \centering
        \vspace*{-0.12cm}
        \includegraphics[width=\linewidth]{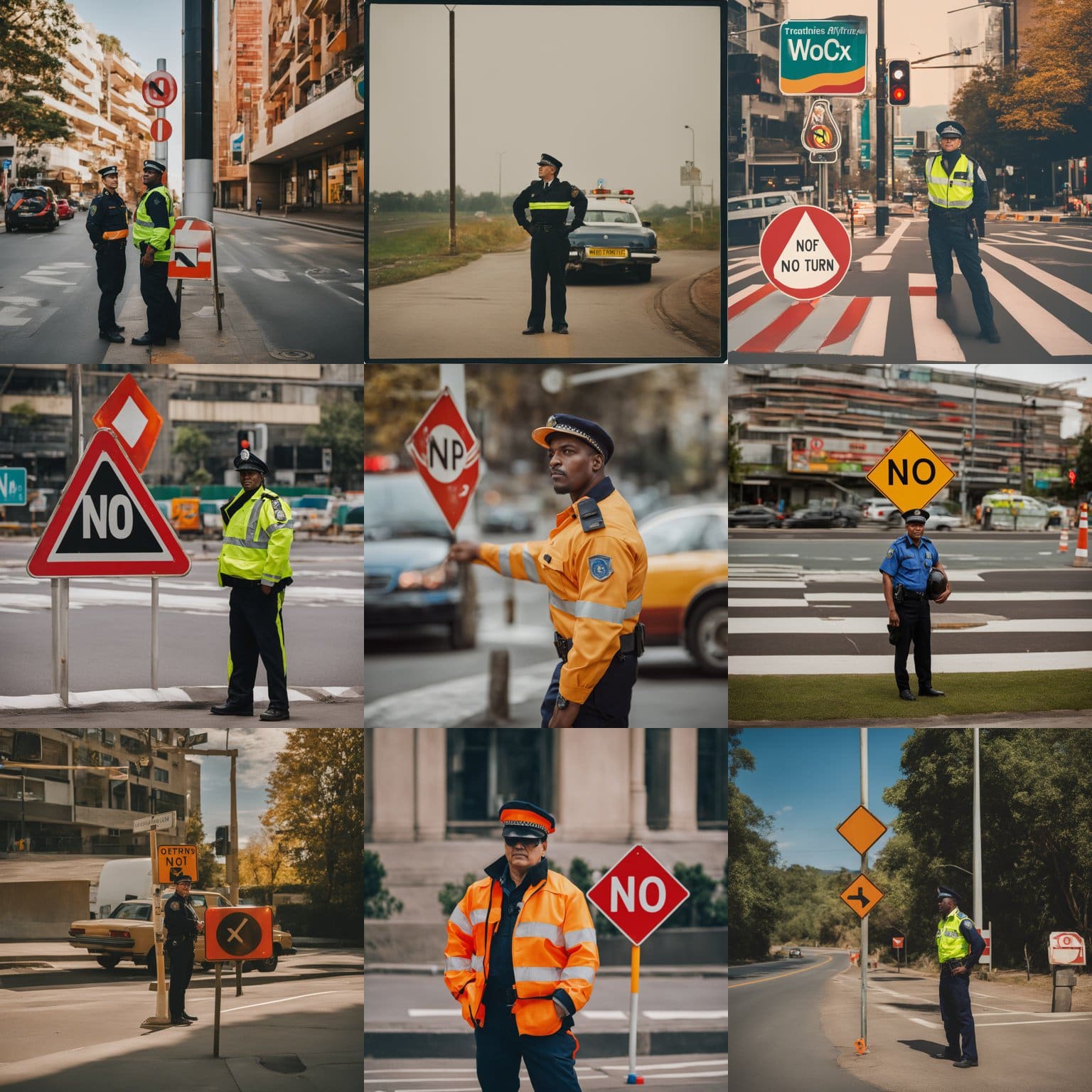}
        \captionsetup{labelformat=empty}
        \caption{SD-XL}
    \end{subfigure}
    \hfill
    \begin{subfigure}{0.33\linewidth}
        \centering
        \textbf{Person gender}\par\medskip
        \vspace*{-0.12cm}
        \includegraphics[width=\linewidth]{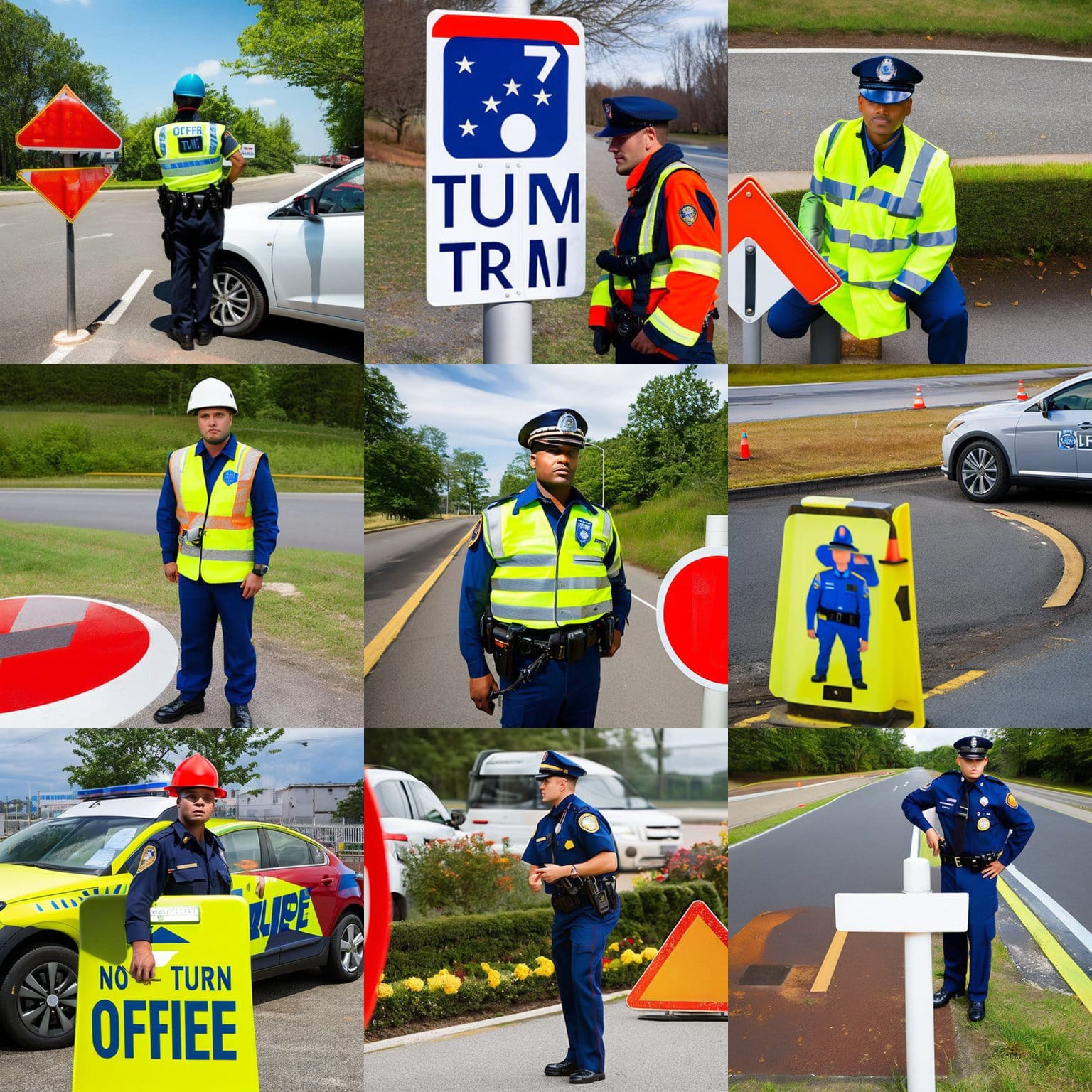}
        \captionsetup{labelformat=empty, justification=centering}
        \caption{SD-2}
    \end{subfigure}
    \hfill
    \begin{subfigure}{0.33\linewidth}
        \centering
        \vspace*{-0.12cm}
        \includegraphics[width=\linewidth]{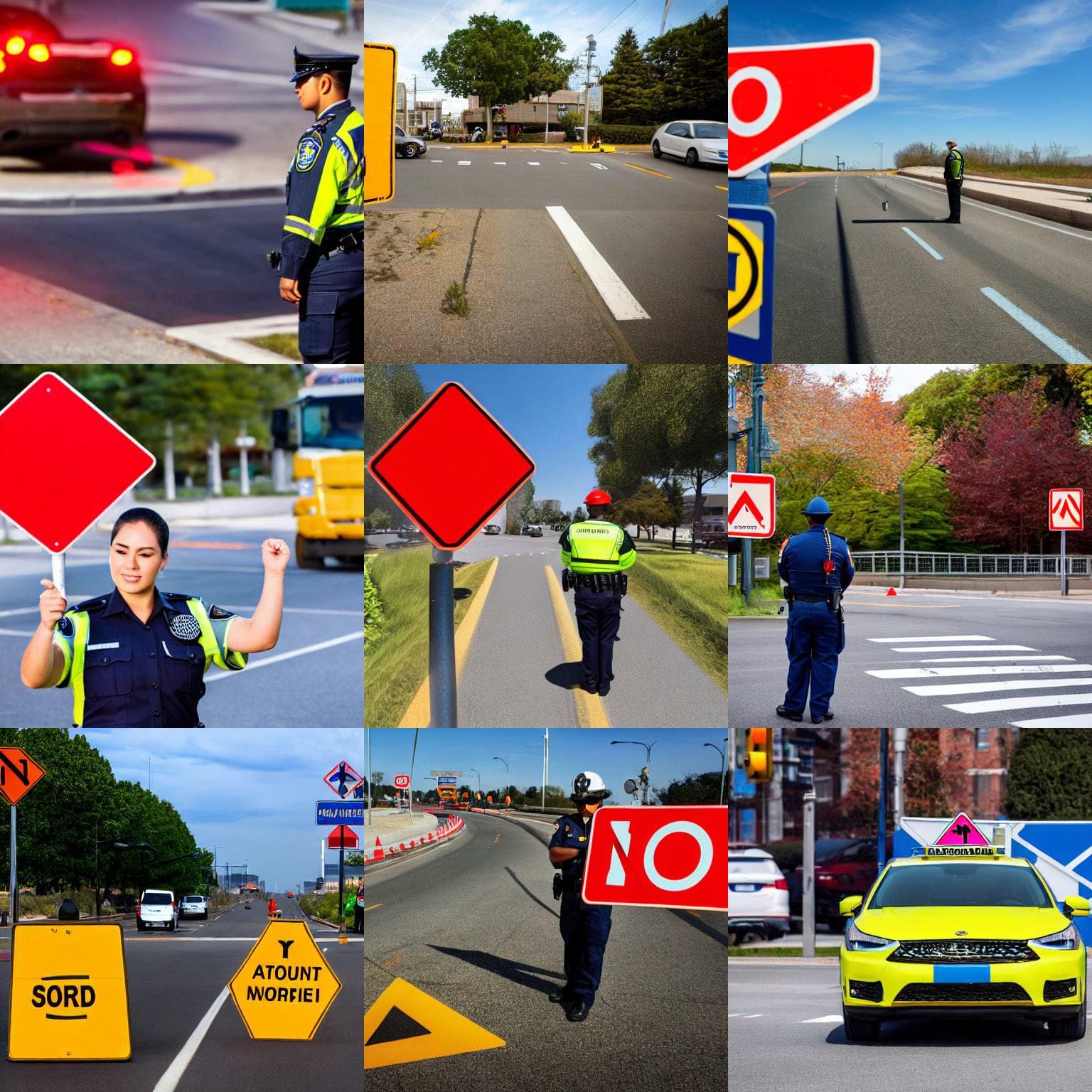}
        \captionsetup{labelformat=empty, justification=centering}
        \caption{SD-1.5}
    \end{subfigure}
    \caption{Comparison on images generated with the same caption ``A traffic officer leaning on a no turn sign".}
    \label{fig:qualitative_comp_gender}
\end{figure*}

\begin{figure*}[htbp]
    \centering
    \begin{subfigure}{0.33\linewidth}
        \centering
        \vspace*{-0.12cm}
        \includegraphics[width=\linewidth]{images/OpenBias_qualitatives/person_race/618785.jpg}
        \captionsetup{labelformat=empty}
        \caption{SD-XL}
    \end{subfigure}
    \hfill
    \begin{subfigure}{0.33\linewidth}
        \centering
        \textbf{Person race}\par\medskip
        \vspace*{-0.12cm}
        \includegraphics[width=\linewidth]{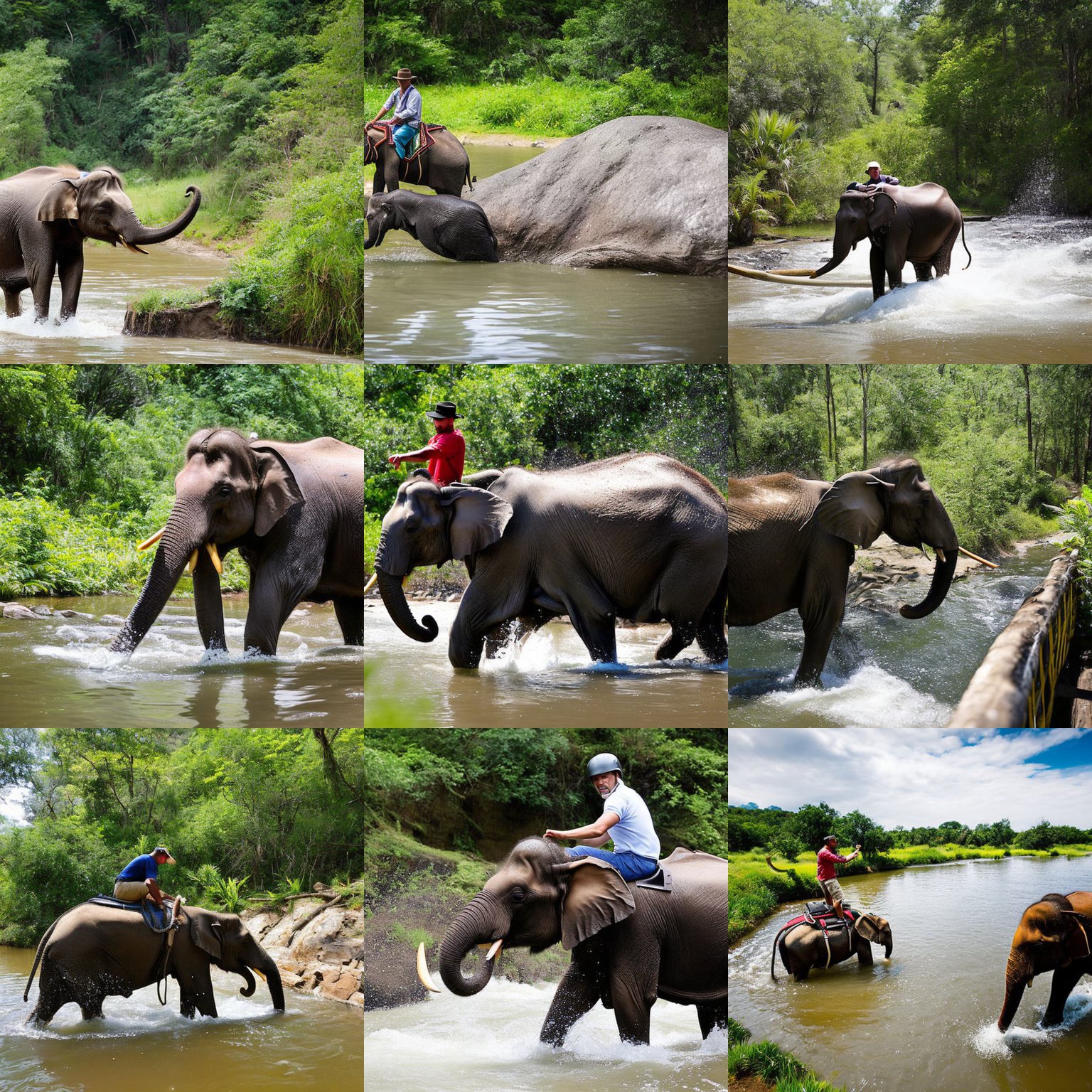}
        \captionsetup{labelformat=empty, justification=centering}
        \caption{SD-2}
    \end{subfigure}
    \hfill
    \begin{subfigure}{0.33\linewidth}
        \centering
        \vspace*{-0.12cm}
        \includegraphics[width=\linewidth]{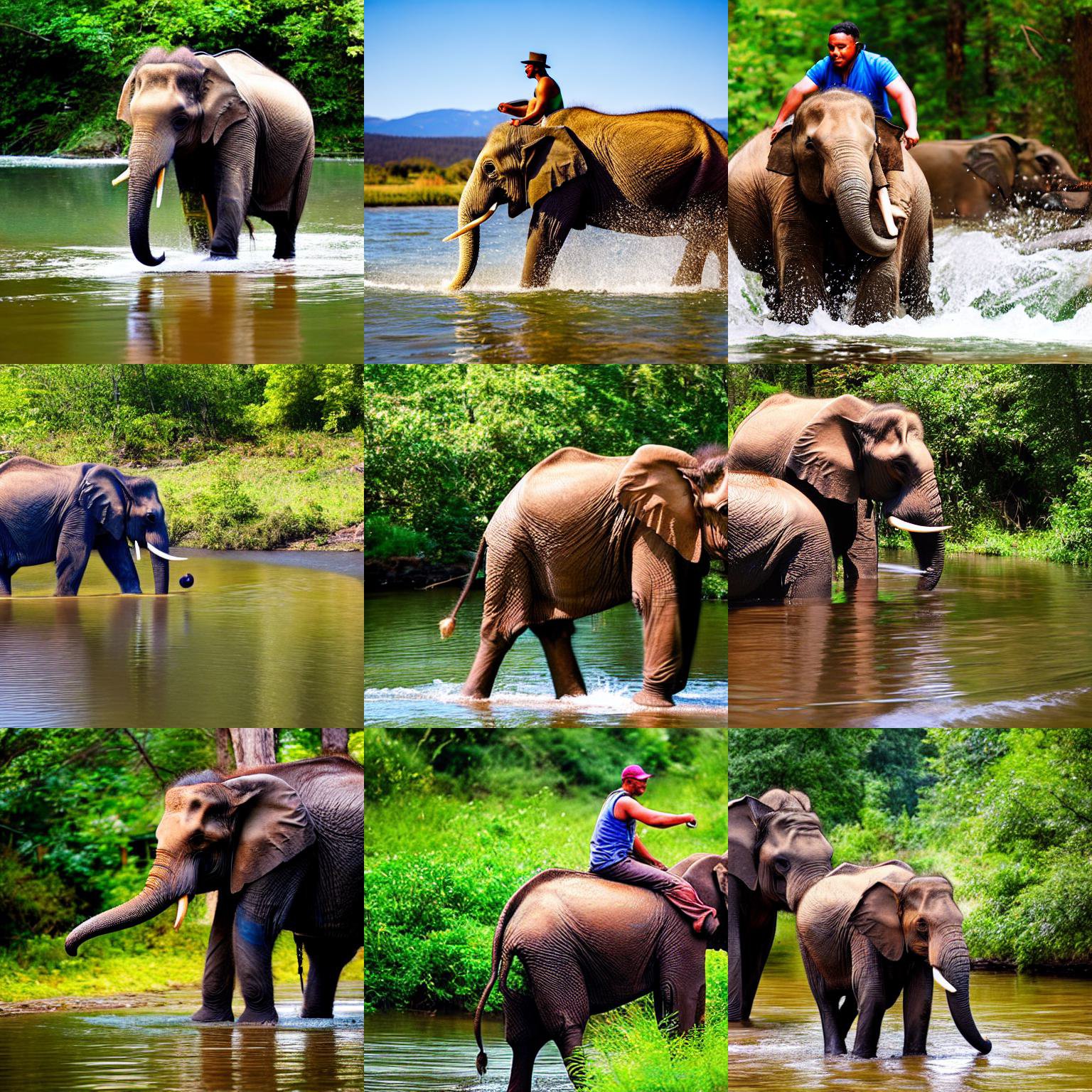}
        \captionsetup{labelformat=empty, justification=centering}
        \caption{SD-1.5}
    \end{subfigure}
    \caption{``A man riding an elephant into some water of a creek".}
    \label{fig:qualitative_comp_race}
\end{figure*}

\begin{figure*}[htbp]
    \centering
    \begin{subfigure}{0.33\linewidth}
        \centering
        \vspace*{-0.12cm}
        \includegraphics[width=\linewidth]{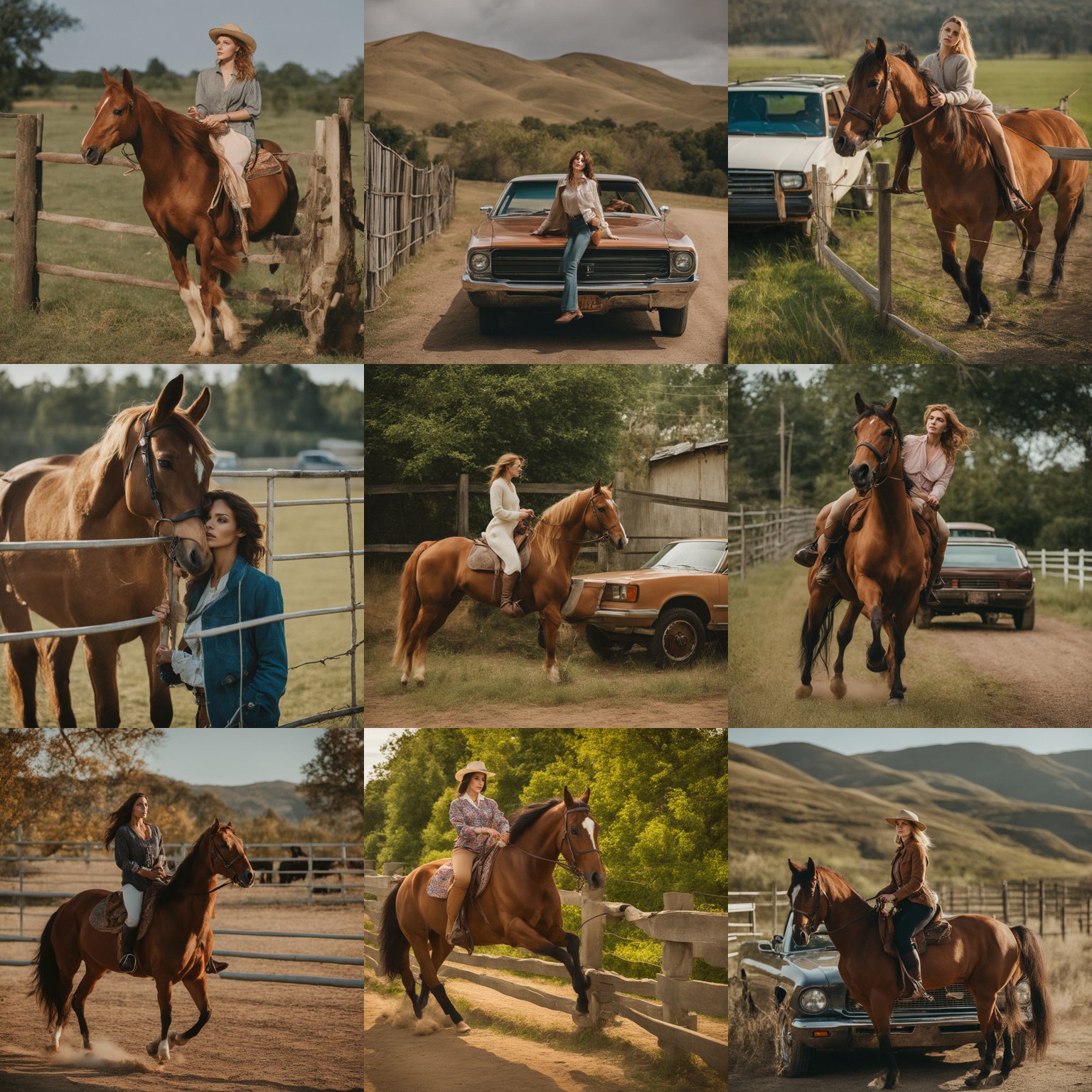}
        \captionsetup{labelformat=empty}
        \caption{SD-XL}
    \end{subfigure}
    \hfill
    \begin{subfigure}{0.33\linewidth}
        \centering
        \textbf{Person age}\par\medskip
        \vspace*{-0.12cm}
        \includegraphics[width=\linewidth]{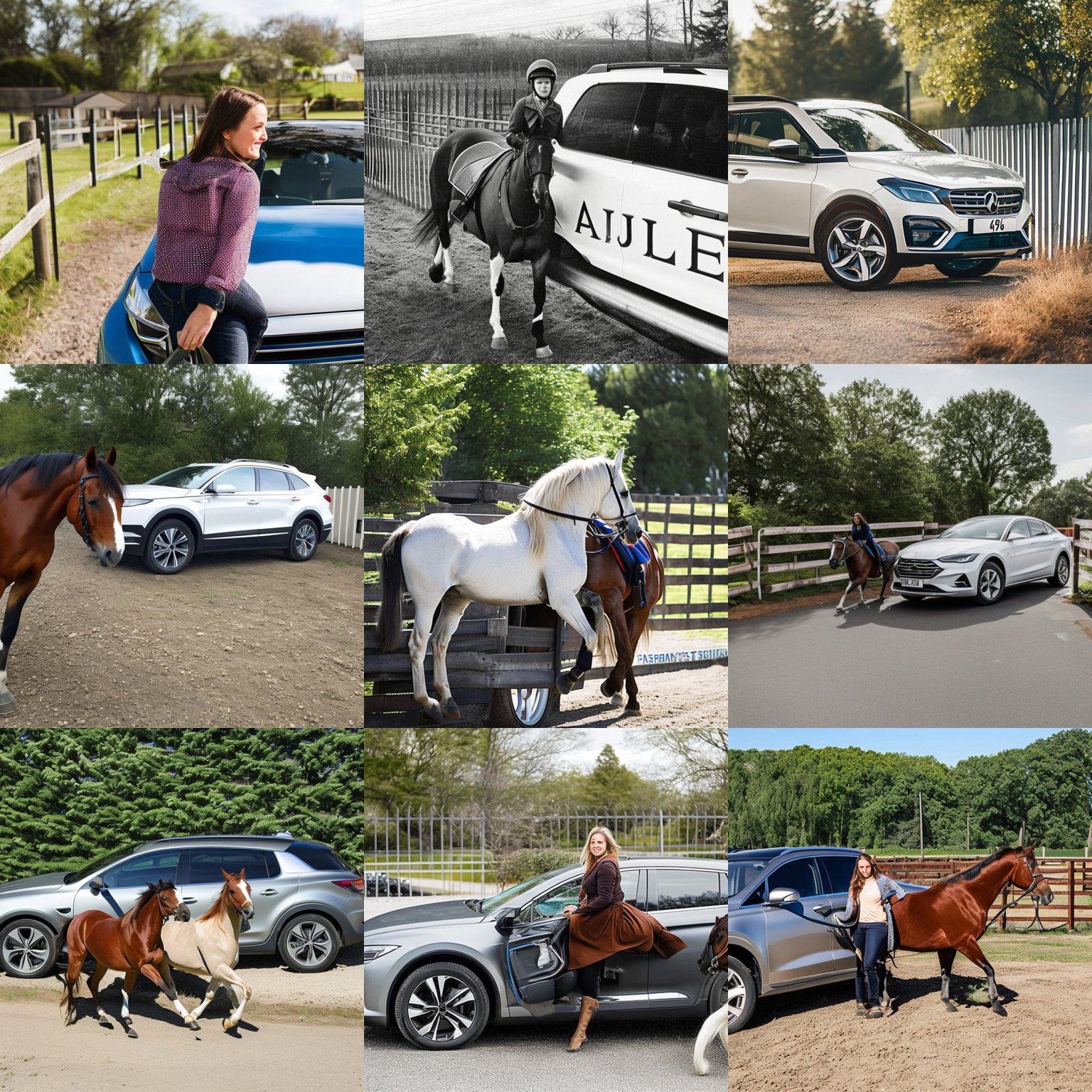}
        \captionsetup{labelformat=empty, justification=centering}
        \caption{SD-2}
    \end{subfigure}
    \hfill
    \begin{subfigure}{0.33\linewidth}
        \centering
        \vspace*{-0.12cm}
        \includegraphics[width=\linewidth]{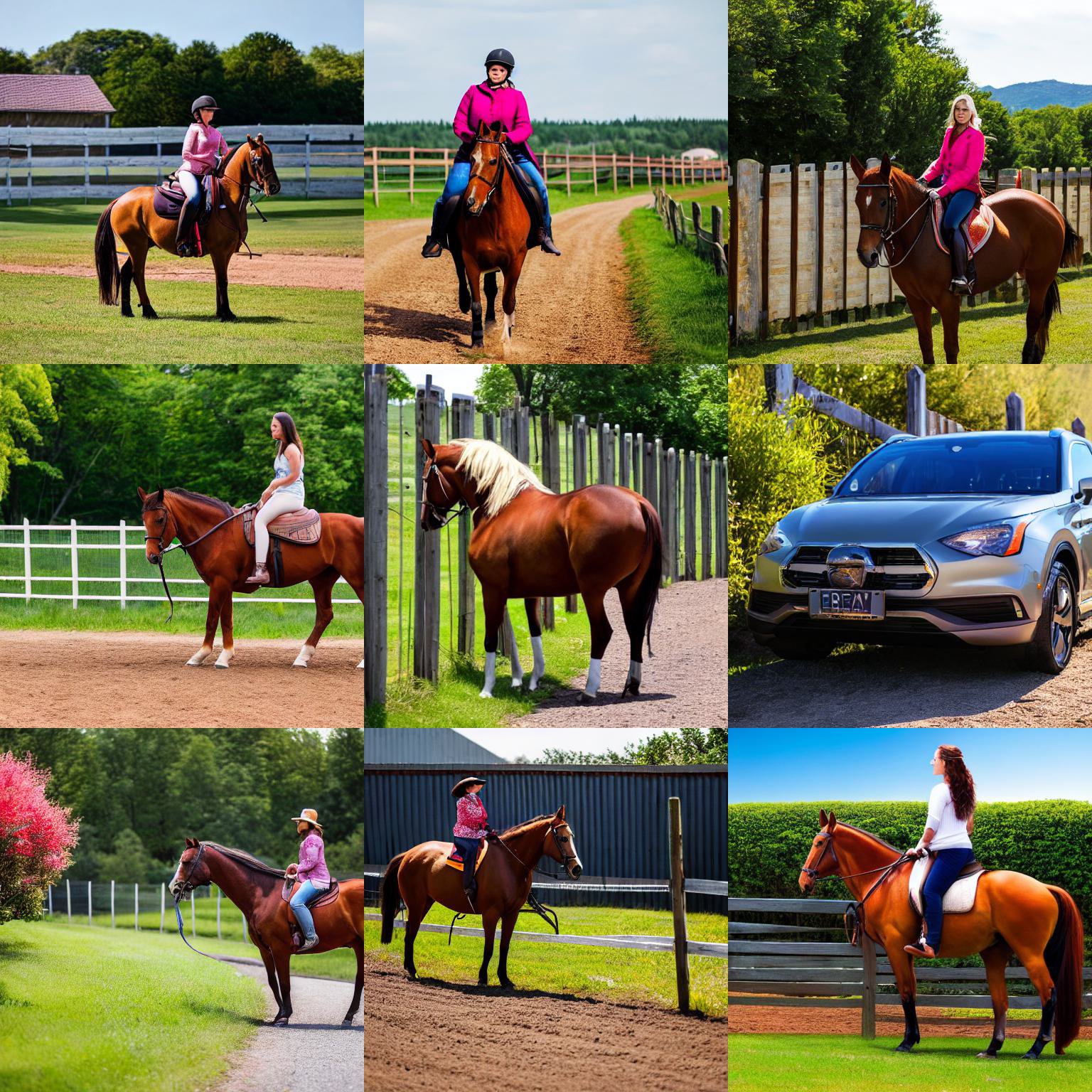}
        \captionsetup{labelformat=empty, justification=centering}
        \caption{SD-1.5}
    \end{subfigure}
    \caption{``A woman riding a horse in front of a car next to a fence".}
    \label{fig:qualitative_comp_age}
\end{figure*}

\begin{figure*}[htbp]
    \centering
    \begin{subfigure}{0.33\linewidth}
        \centering
        \vspace*{-0.12cm}
        \includegraphics[width=\linewidth]{images/OpenBias_qualitatives/child_gender/622215.jpg}
        \captionsetup{labelformat=empty}
        \caption{SD-XL}
    \end{subfigure}
    \hfill
    \begin{subfigure}{0.33\linewidth}
        \centering
        \textbf{Child gender}\par\medskip
        \vspace*{-0.12cm}
        \includegraphics[width=\linewidth]{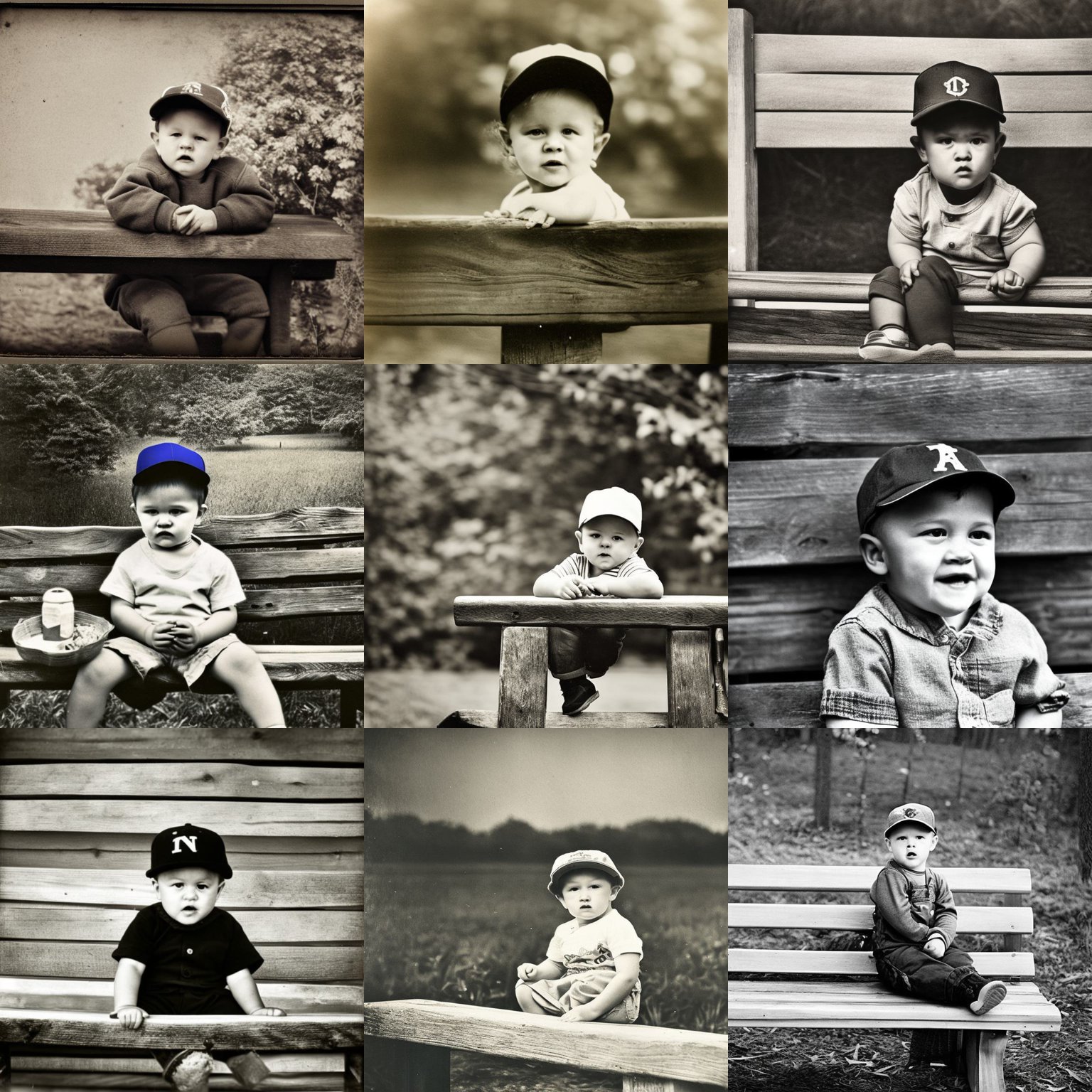}
        \captionsetup{labelformat=empty, justification=centering}
        \caption{SD-2}
    \end{subfigure}
    \hfill
    \begin{subfigure}{0.33\linewidth}
        \centering
        \vspace*{-0.12cm}
        \includegraphics[width=\linewidth]{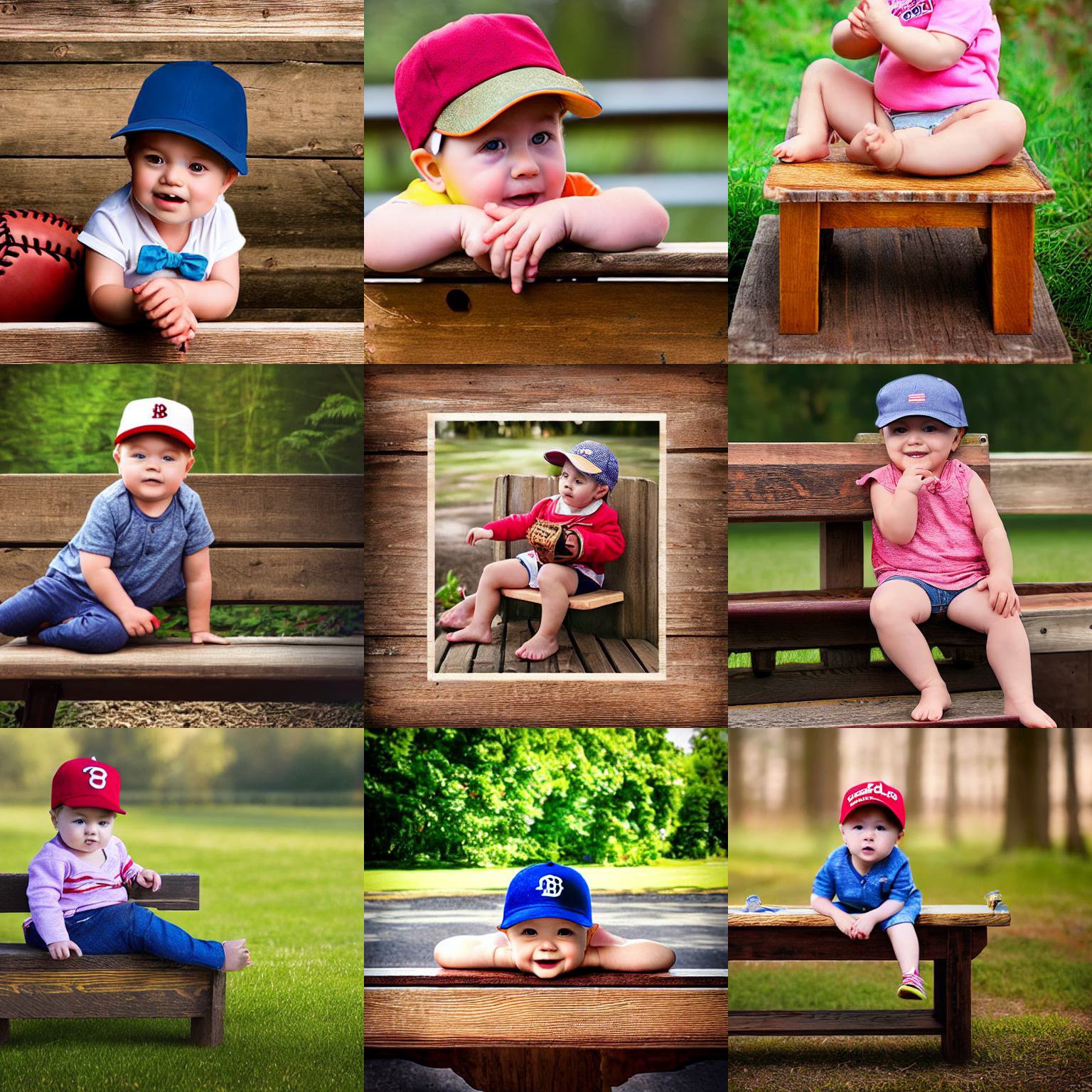}
        \captionsetup{labelformat=empty, justification=centering}
        \caption{SD-1.5}
    \end{subfigure}
    \caption{``Toddler in a baseball cap on a wooden bench".}
    \label{fig:qualitative_comp_child_gender}
\end{figure*}

\begin{figure*}[htbp]
    \centering
    \begin{subfigure}{0.33\linewidth}
        \centering
        \vspace*{-0.12cm}
        \includegraphics[width=\linewidth]{images/OpenBias_qualitatives/child_race/483795.jpg}
        \captionsetup{labelformat=empty}
        \caption{SD-XL}
    \end{subfigure}
    \hfill
    \begin{subfigure}{0.33\linewidth}
        \centering
        \textbf{Child race}\par\medskip
        \vspace*{-0.12cm}
        \includegraphics[width=\linewidth]{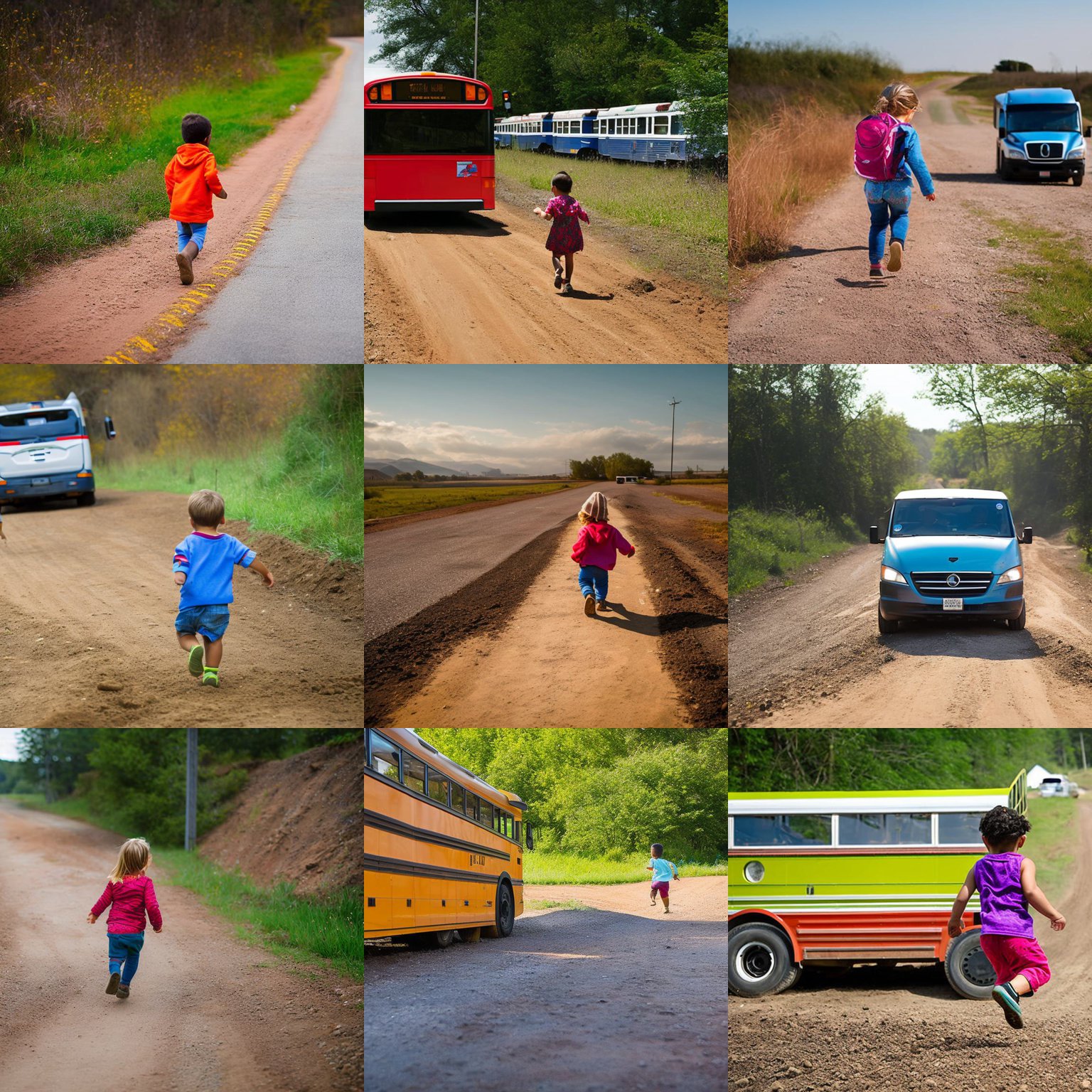}
        \captionsetup{labelformat=empty, justification=centering}
        \caption{SD-2}
    \end{subfigure}
    \hfill
    \begin{subfigure}{0.33\linewidth}
        \centering
        \vspace*{-0.12cm}
        \includegraphics[width=\linewidth]{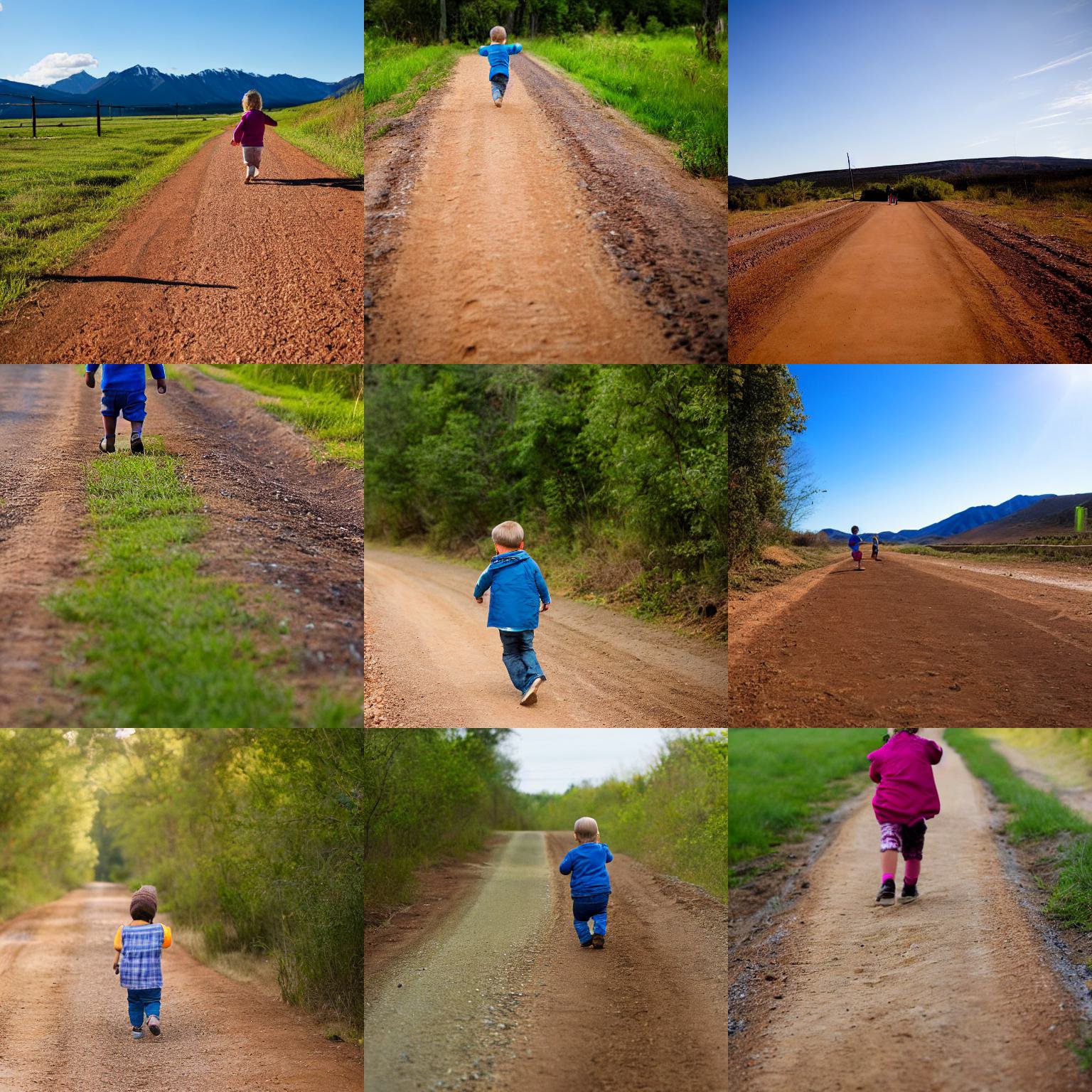}
        \captionsetup{labelformat=empty, justification=centering}
        \caption{SD-1.5}
    \end{subfigure}
    \caption{``Small child hurrying toward a bus on a dirt road".}
    \label{fig:qualitative_comp_child_race}
\end{figure*}

\begin{figure*}[htbp]
    \centering
    \begin{subfigure}{0.33\linewidth}
        \centering
        \vspace*{-0.12cm}
        \includegraphics[width=\linewidth]{images/OpenBias_qualitatives/person_attire/341725.jpg}
        \captionsetup{labelformat=empty}
        \caption{SD-XL}
    \end{subfigure}
    \hfill
    \begin{subfigure}{0.33\linewidth}
        \centering
        \textbf{Person attire}\par\medskip
        \vspace*{-0.12cm}
        \includegraphics[width=\linewidth]{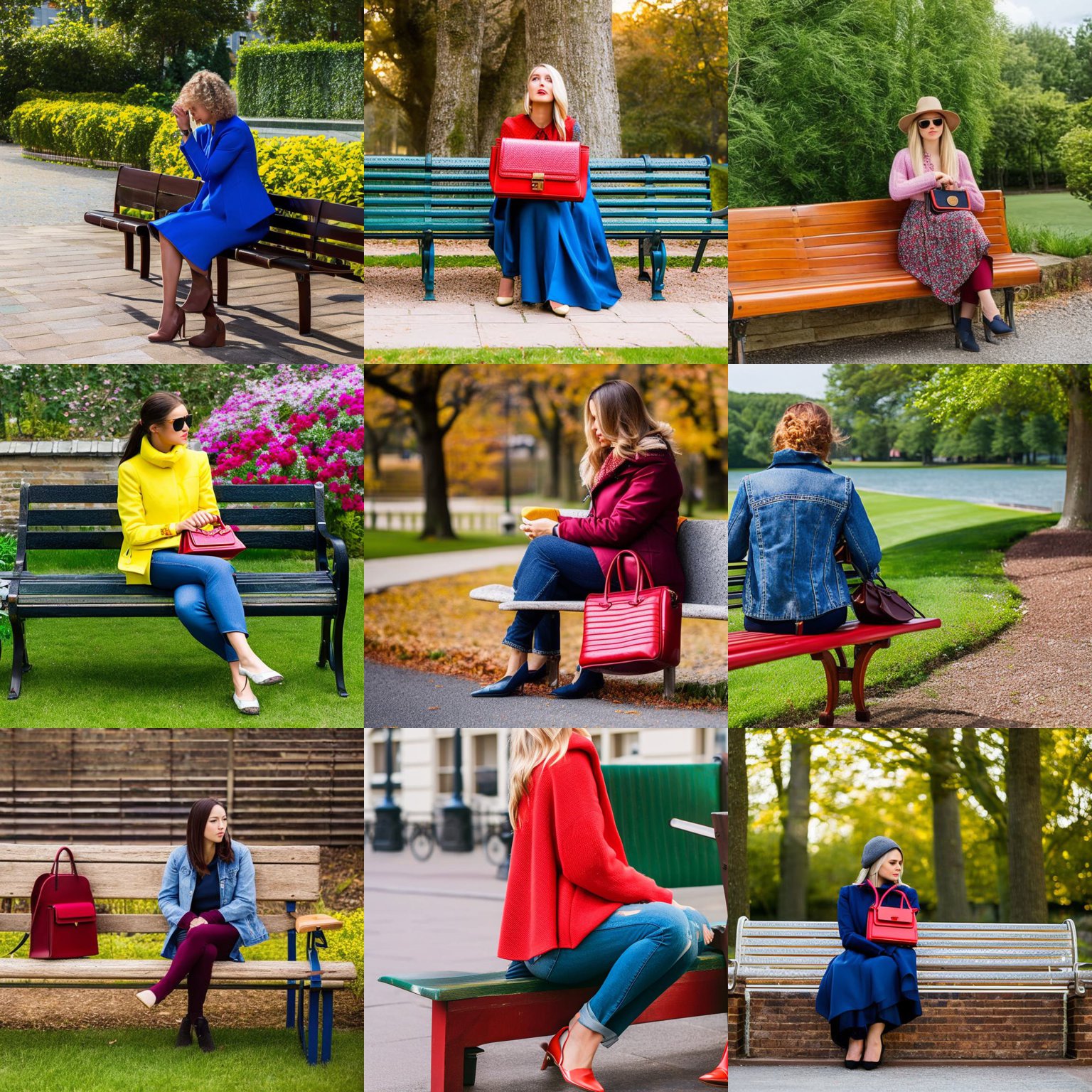}
        \captionsetup{labelformat=empty, justification=centering}
        \caption{SD-2}
    \end{subfigure}
    \hfill
    \begin{subfigure}{0.33\linewidth}
        \centering
        \vspace*{-0.12cm}
        \includegraphics[width=\linewidth]{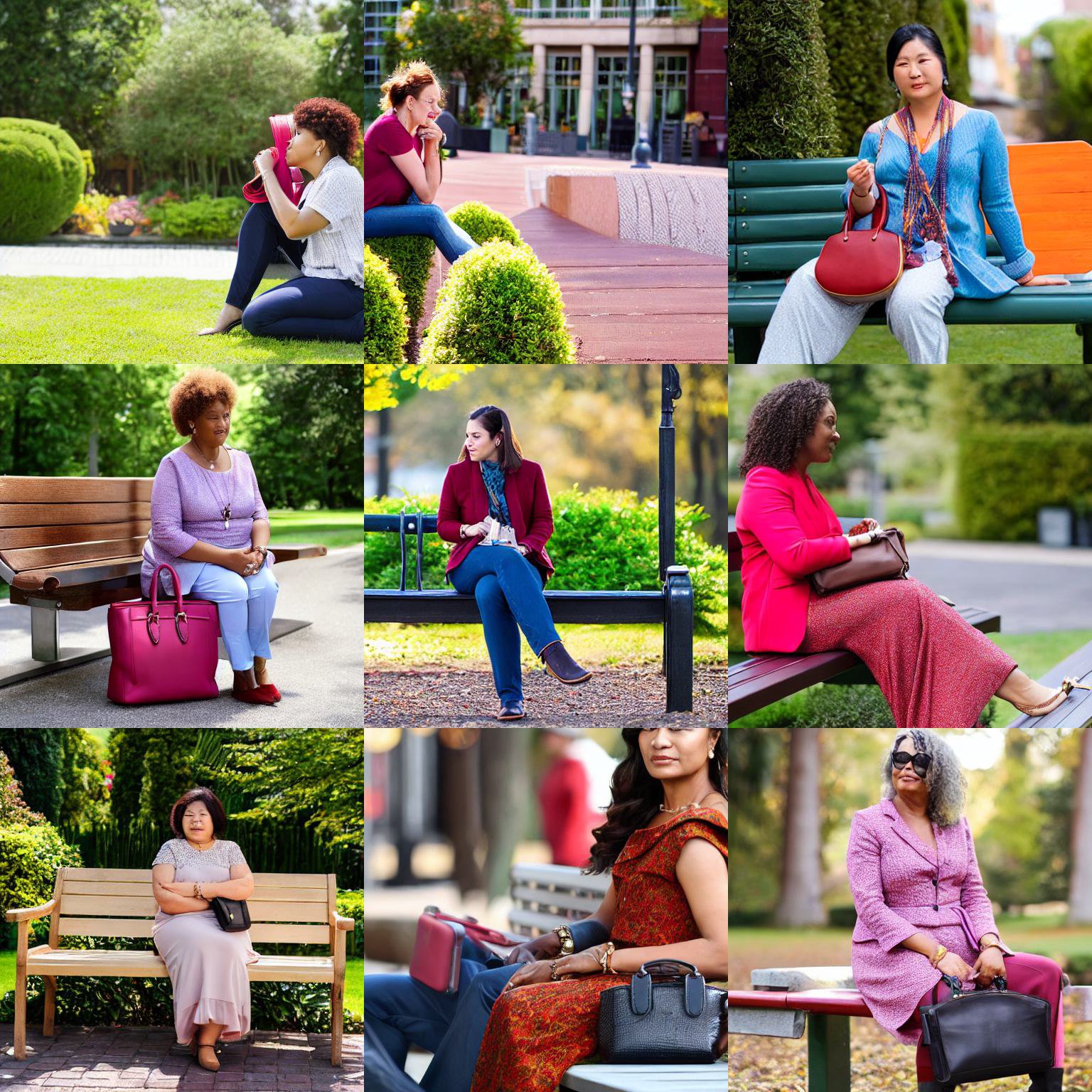}
        \captionsetup{labelformat=empty, justification=centering}
        \caption{SD-1.5}
    \end{subfigure}
    \caption{``The lady is sitting on the bench holding her handbag".}
    \label{fig:qualitative_comp_attire}
\end{figure*}

\begin{figure*}[htbp]
    \centering
    \begin{subfigure}{0.33\linewidth}
        \centering
        \vspace*{-0.12cm}
        \includegraphics[width=\linewidth]{images/OpenBias_qualitatives/train_color/785580.jpg}
        \captionsetup{labelformat=empty}
        \caption{SD-XL}
    \end{subfigure}
    \hfill
    \begin{subfigure}{0.33\linewidth}
        \centering
        \textbf{Train color}\par\medskip
        \vspace*{-0.12cm}
        \includegraphics[width=\linewidth]{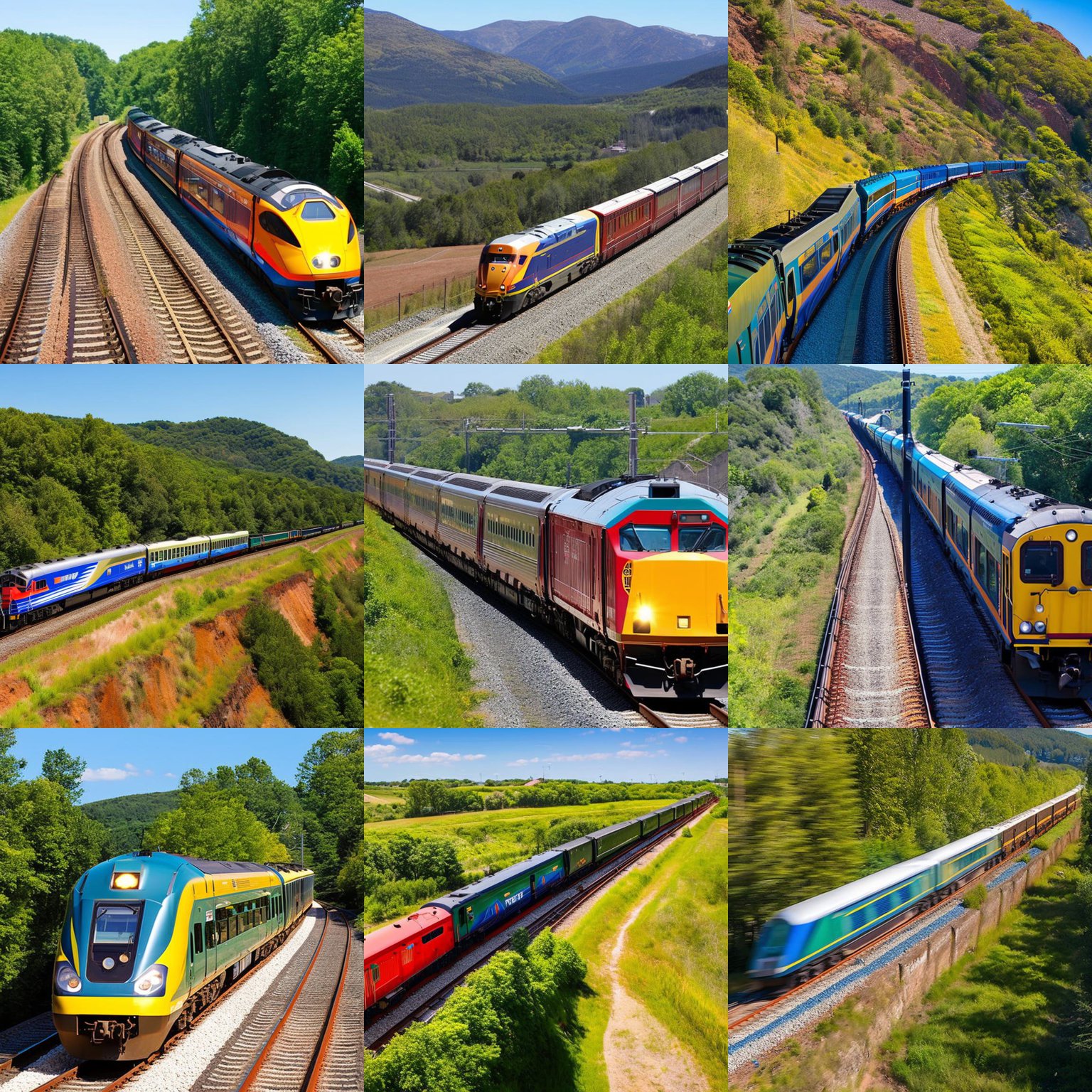}
        \captionsetup{labelformat=empty, justification=centering}
        \caption{SD-2}
    \end{subfigure}
    \hfill
    \begin{subfigure}{0.33\linewidth}
        \centering
        \vspace*{-0.12cm}
        \includegraphics[width=\linewidth]{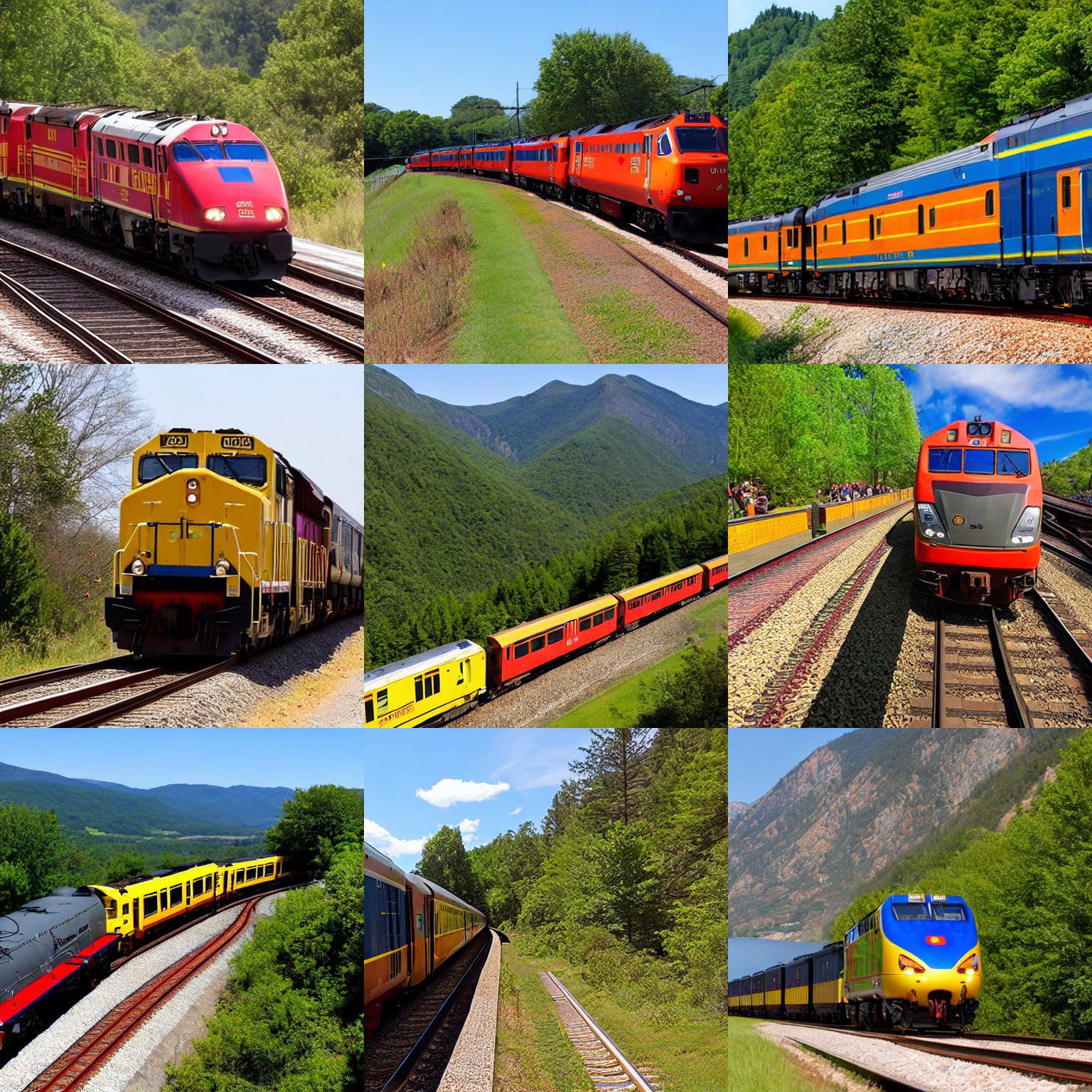}
        \captionsetup{labelformat=empty, justification=centering}
        \caption{SD-1.5}
    \end{subfigure}
    \caption{``A train zips down the railway in the sun".}
    \label{fig:qualitative_comp_train}
\end{figure*}

\begin{figure*}[htbp]
    \centering
    \begin{subfigure}{0.33\linewidth}
        \centering
        \vspace*{-0.12cm}
        \includegraphics[width=\linewidth]{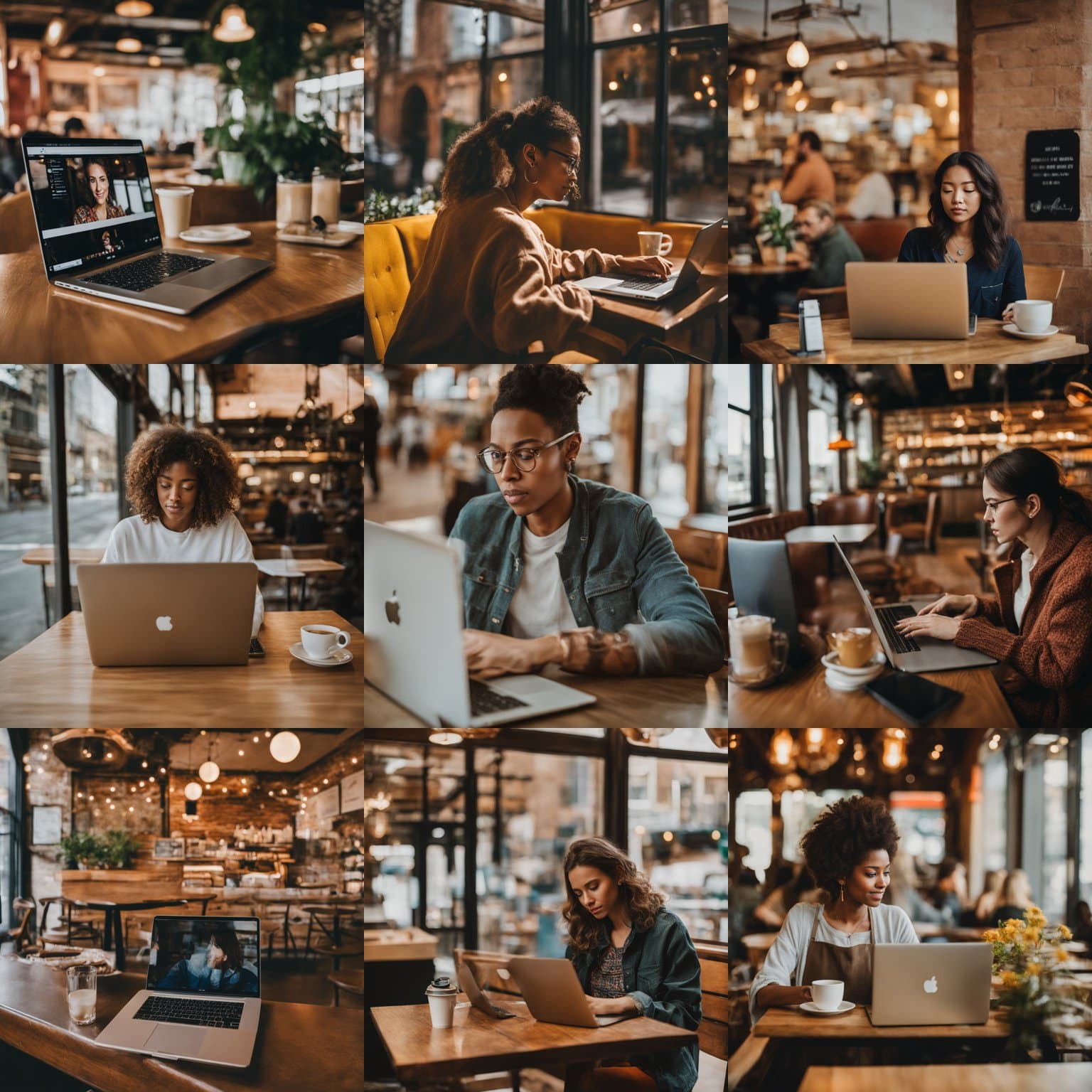}
        \captionsetup{labelformat=empty}
        \caption{SD-XL}
    \end{subfigure}
    \hfill
    \begin{subfigure}{0.33\linewidth}
        \centering
        \textbf{Laptop brand}\par\medskip
        \vspace*{-0.12cm}
        \includegraphics[width=\linewidth]{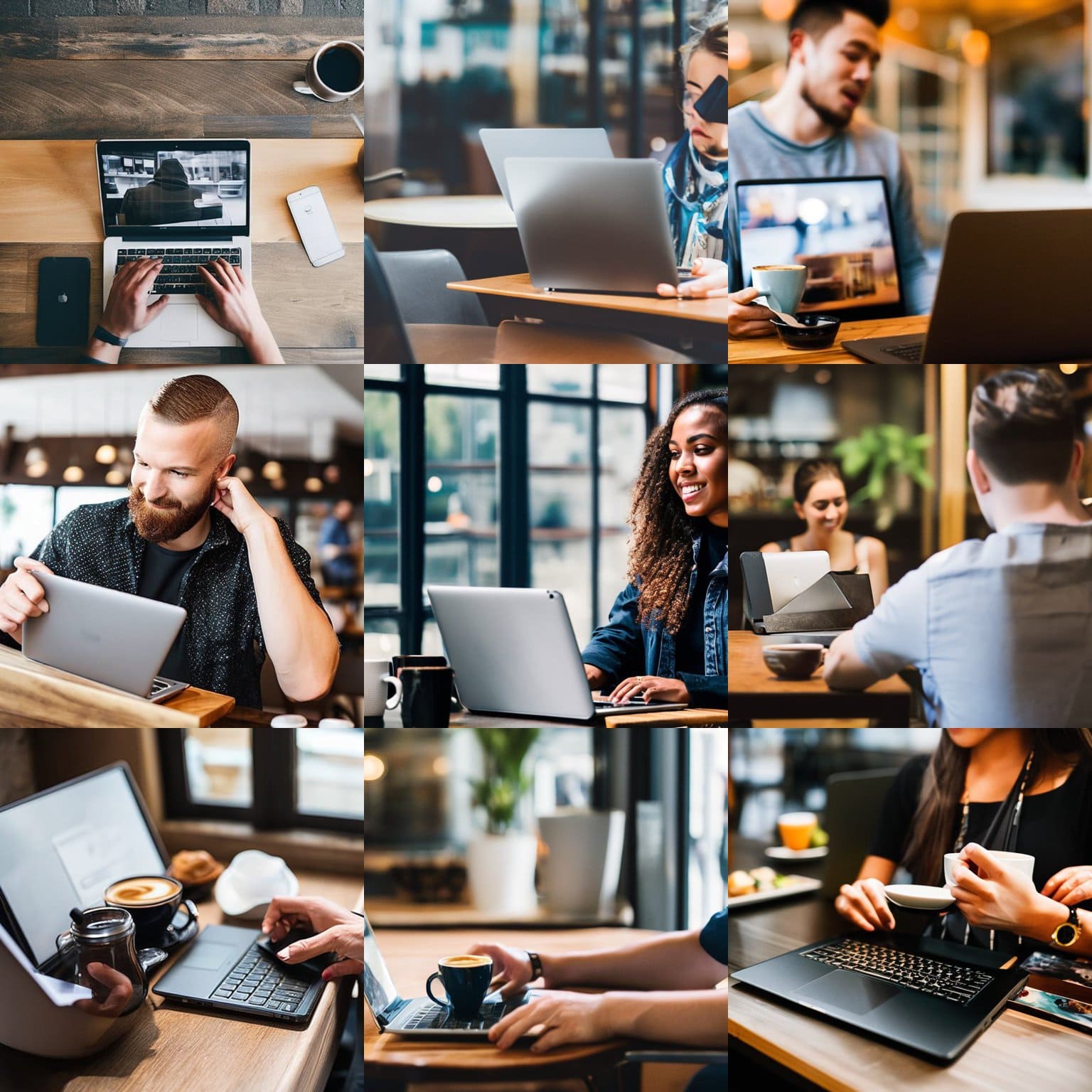}
        \captionsetup{labelformat=empty, justification=centering}
        \caption{SD-2}
    \end{subfigure}
    \hfill
    \begin{subfigure}{0.33\linewidth}
        \centering
        \vspace*{-0.12cm}
        \includegraphics[width=\linewidth]{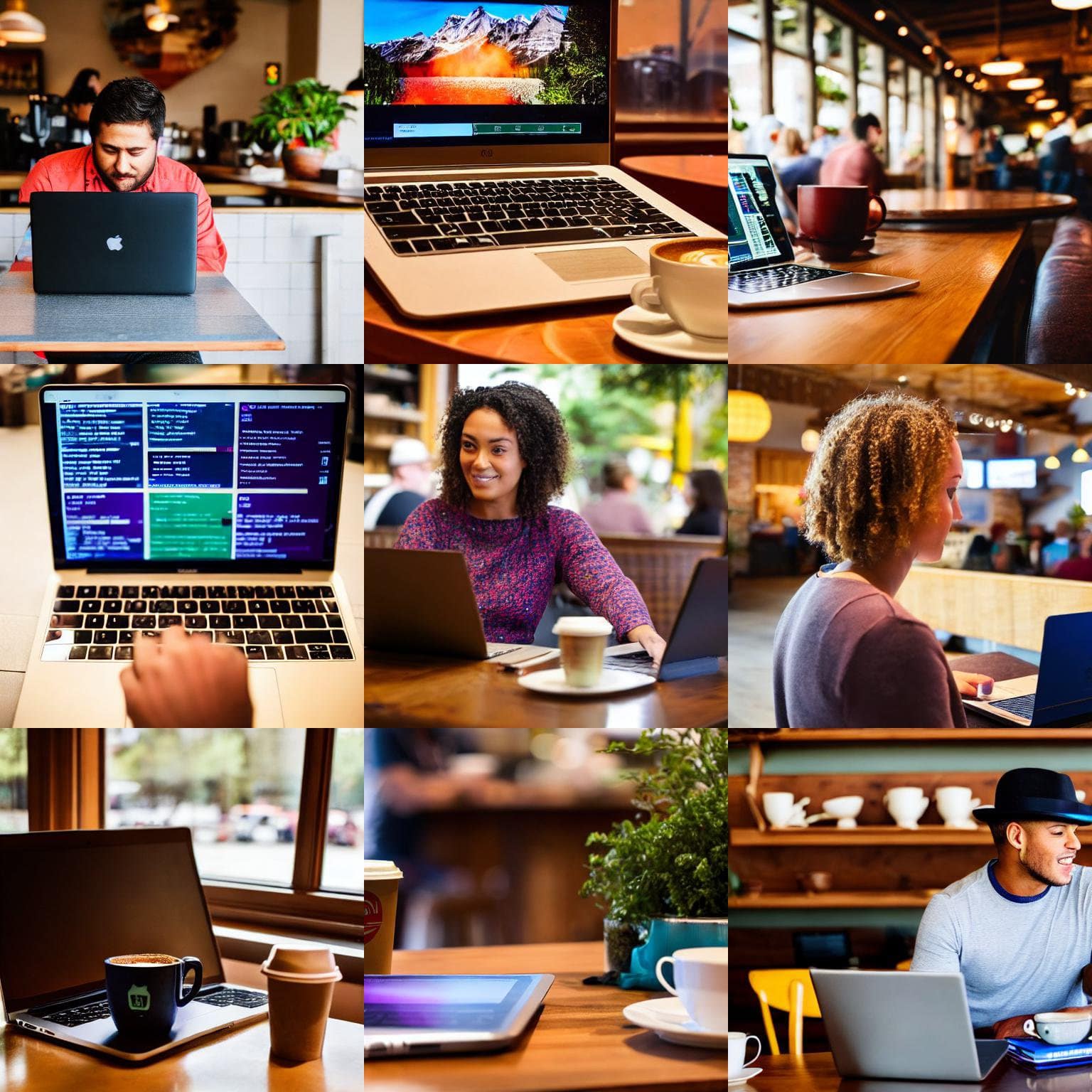}
        \captionsetup{labelformat=empty, justification=centering}
        \caption{SD-1.5}
    \end{subfigure}
    \caption{``A photo of a person on a laptop in a coffee shop".}
    \label{fig:qualitative_comp_laptop}
\end{figure*}

\begin{figure*}[htbp]
    \centering
    \begin{subfigure}{0.33\linewidth}
        \centering
        \vspace*{-0.12cm}
        \includegraphics[width=\linewidth]{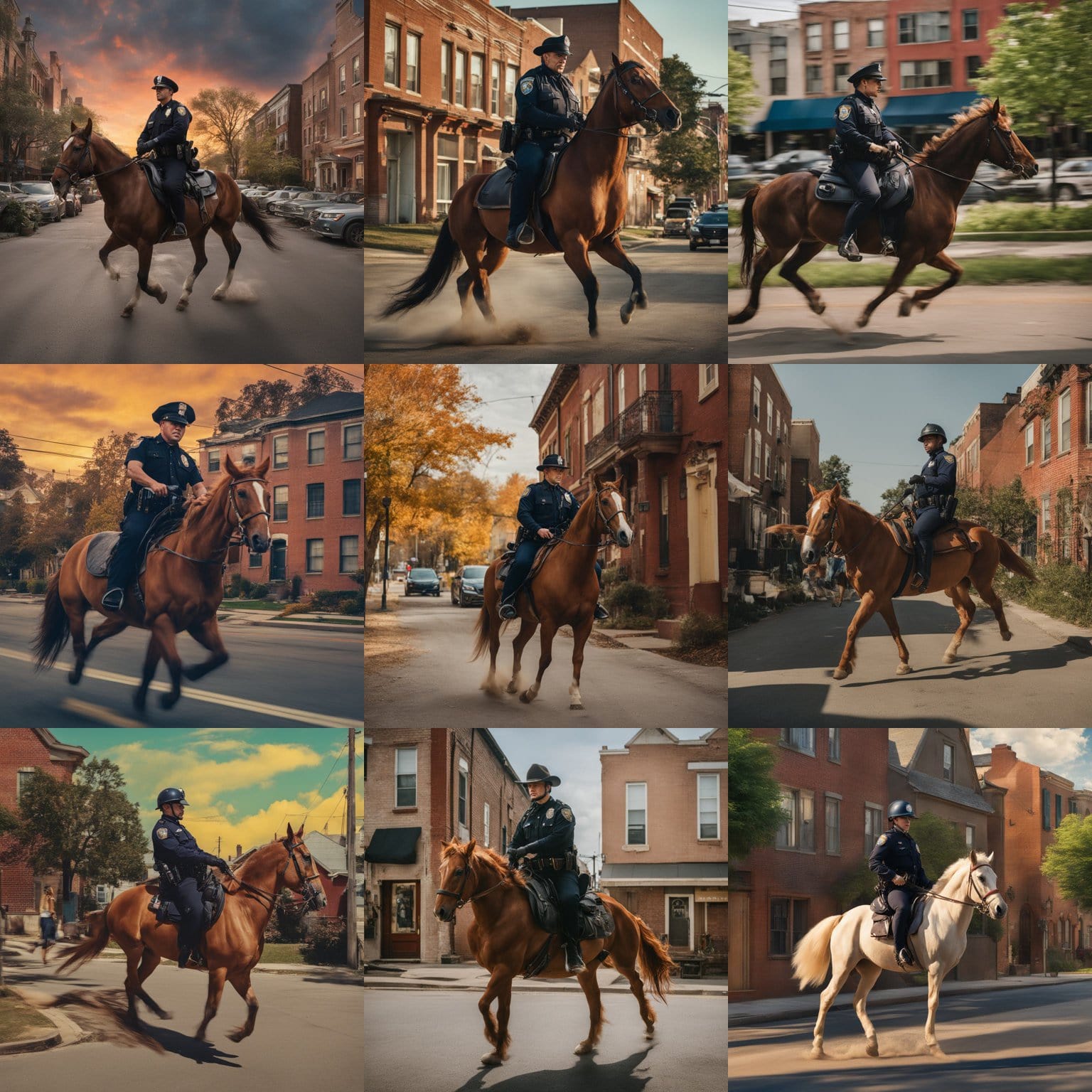}
        \captionsetup{labelformat=empty}
        \caption{SD-XL}
    \end{subfigure}
    \hfill
    \begin{subfigure}{0.33\linewidth}
        \centering
        \textbf{Horse breed}\par\medskip
        \vspace*{-0.12cm}
        \includegraphics[width=\linewidth]{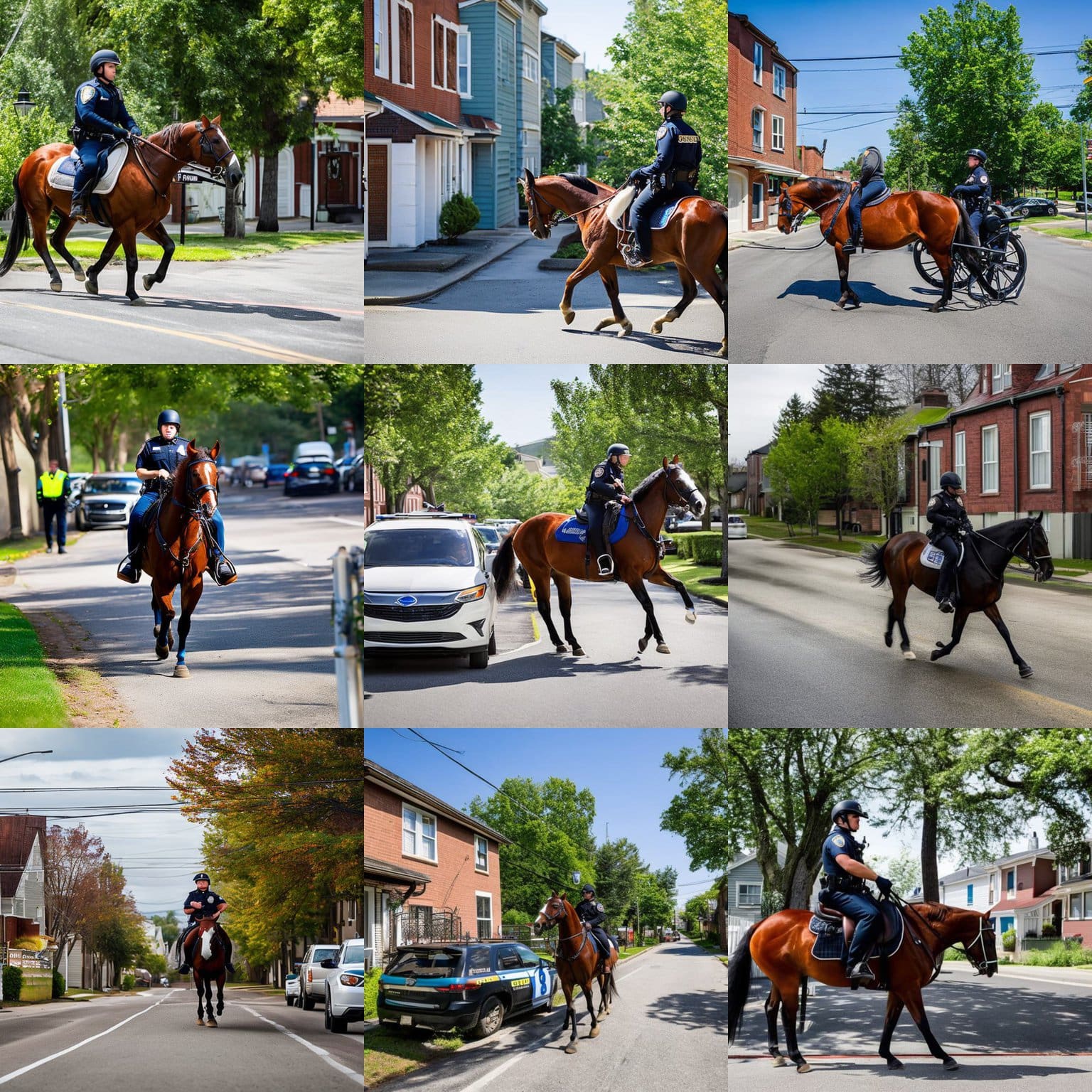}
        \captionsetup{labelformat=empty, justification=centering}
        \caption{SD-2}
    \end{subfigure}
    \hfill
    \begin{subfigure}{0.33\linewidth}
        \centering
        \vspace*{-0.12cm}
        \includegraphics[width=\linewidth]{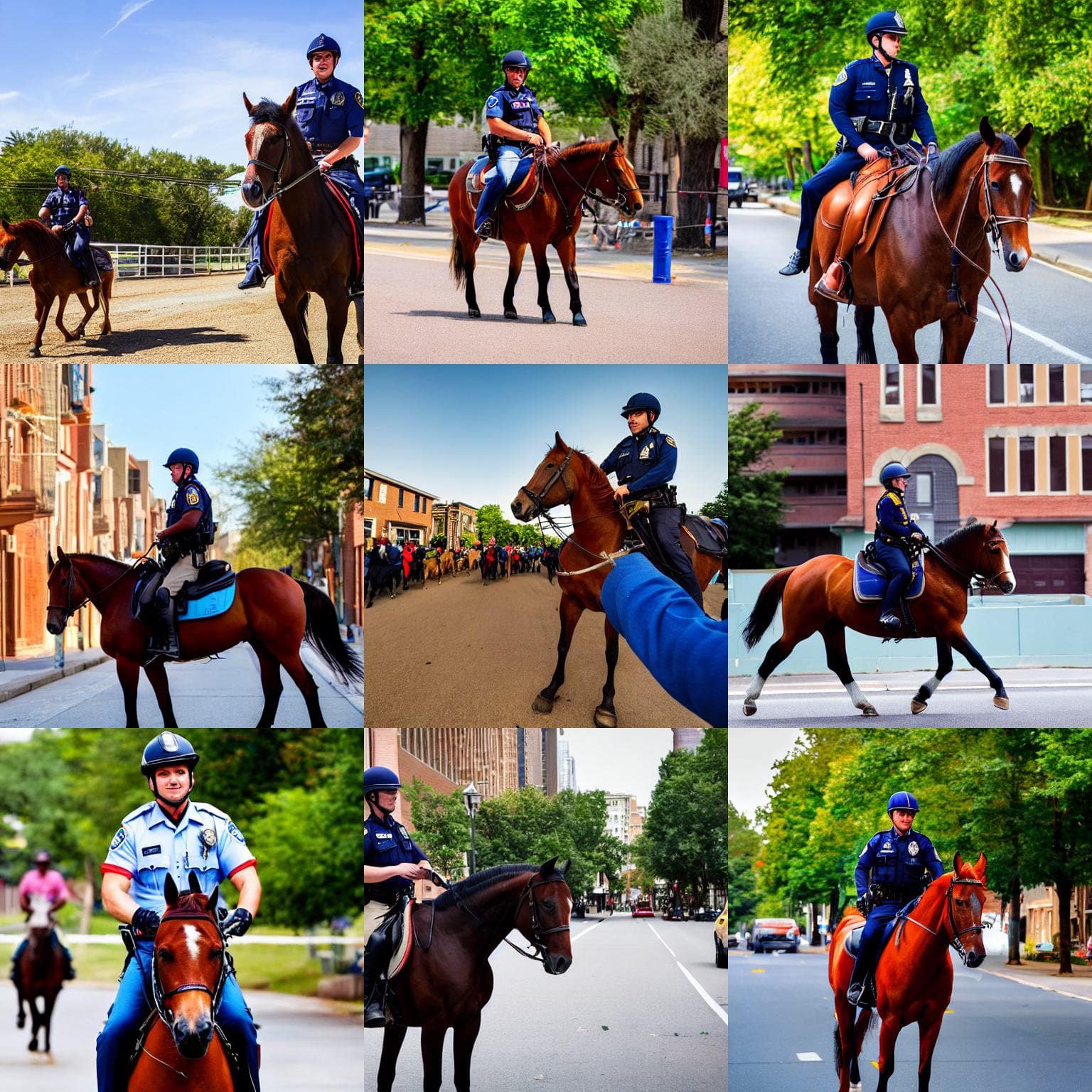}
        \captionsetup{labelformat=empty, justification=centering}
        \caption{SD-1.5}
    \end{subfigure}
    \caption{``A woman riding a horse in front of a car next to a fence".}
    \label{fig:qualitative_comp_horse}
\end{figure*}

\begin{figure*}[htbp]
    \centering
    \begin{subfigure}{0.33\linewidth}
        \centering
        \vspace*{-0.12cm}
        \includegraphics[width=\linewidth]{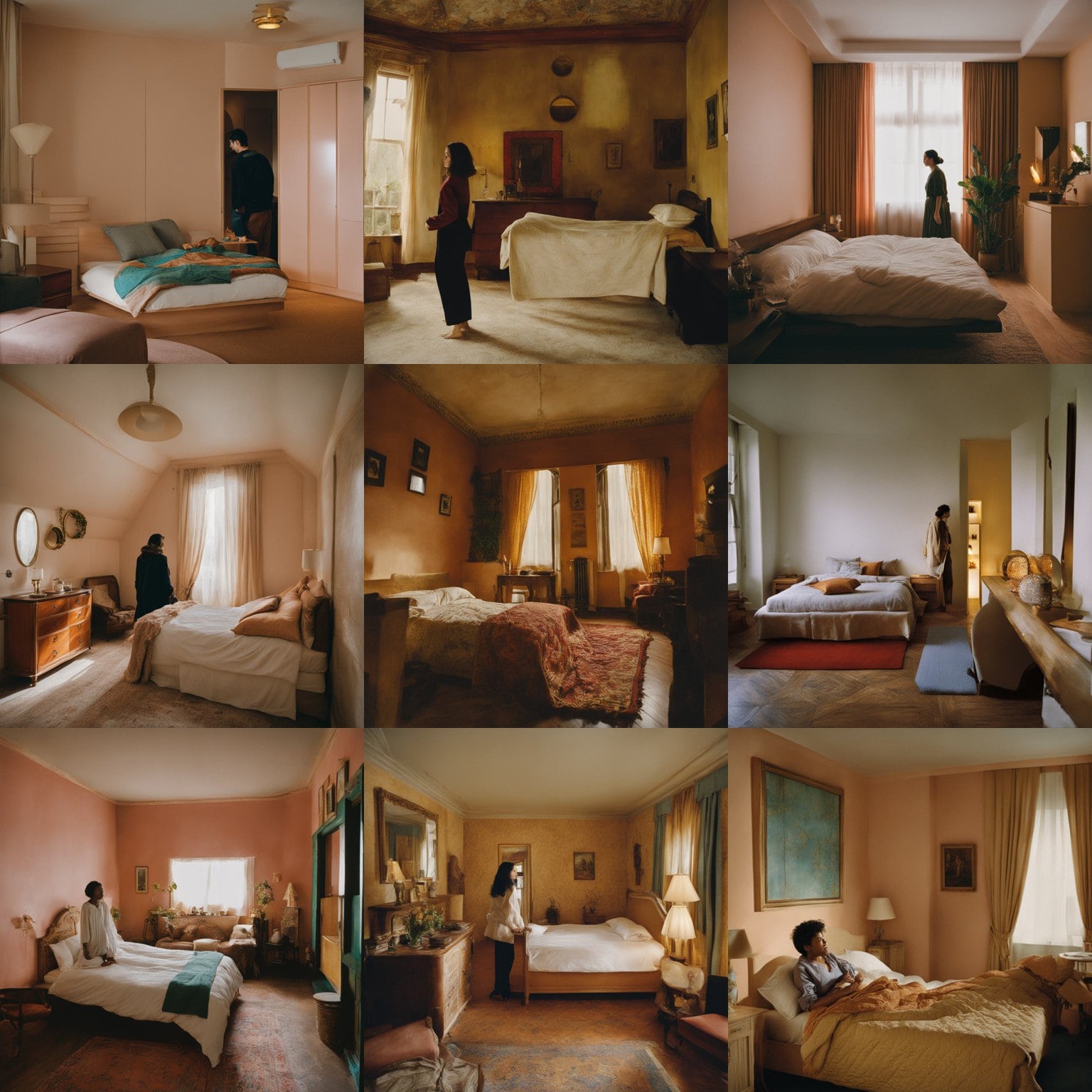}
        \captionsetup{labelformat=empty}
        \caption{SD-XL}
    \end{subfigure}
    \hfill
    \begin{subfigure}{0.33\linewidth}
        \centering
        \textbf{Bed type}\par\medskip
        \vspace*{-0.12cm}
        \includegraphics[width=\linewidth]{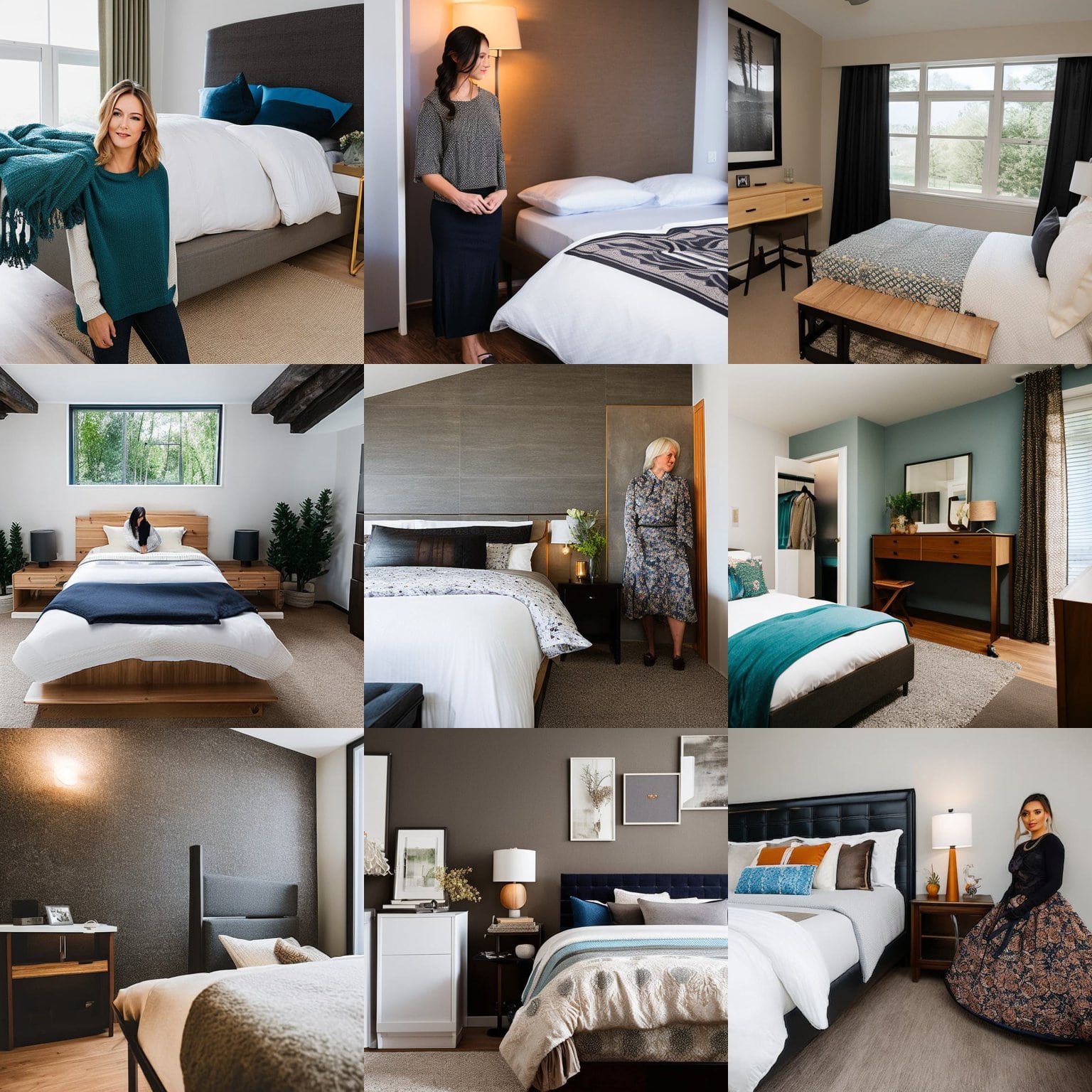}
        \captionsetup{labelformat=empty, justification=centering}
        \caption{SD-2}
    \end{subfigure}
    \hfill
    \begin{subfigure}{0.33\linewidth}
        \centering
        \vspace*{-0.12cm}
        \includegraphics[width=\linewidth]{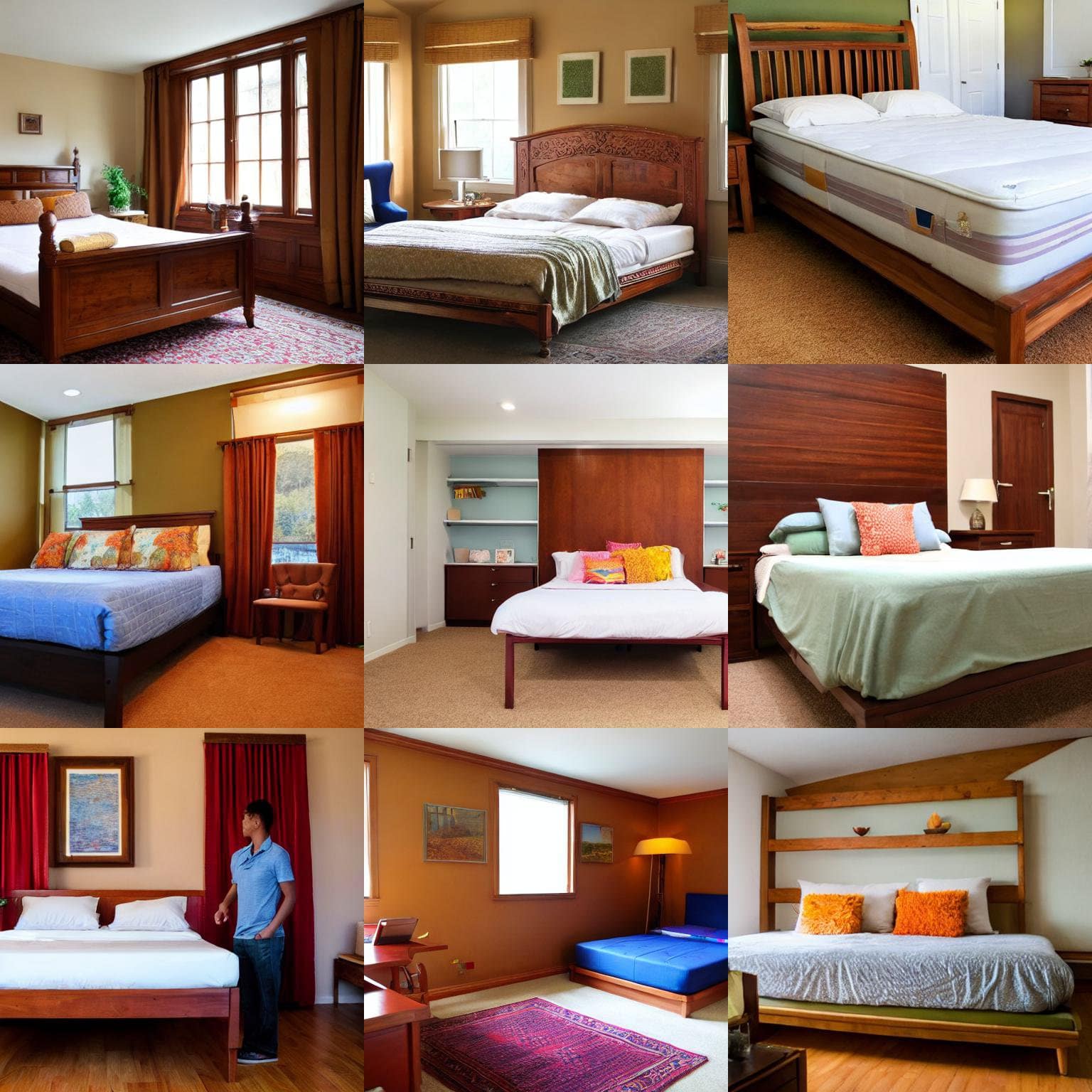}
        \captionsetup{labelformat=empty, justification=centering}
        \caption{SD-1.5}
    \end{subfigure}
    \caption{``A person standing in a bedroom with a bed and a table".}
    \label{fig:qualitative_comp_bed}
\end{figure*}

\begin{figure*}[htbp]
    \centering
    \begin{subfigure}{0.33\linewidth}
        \centering
        \vspace*{-0.12cm}
        \includegraphics[width=\linewidth]{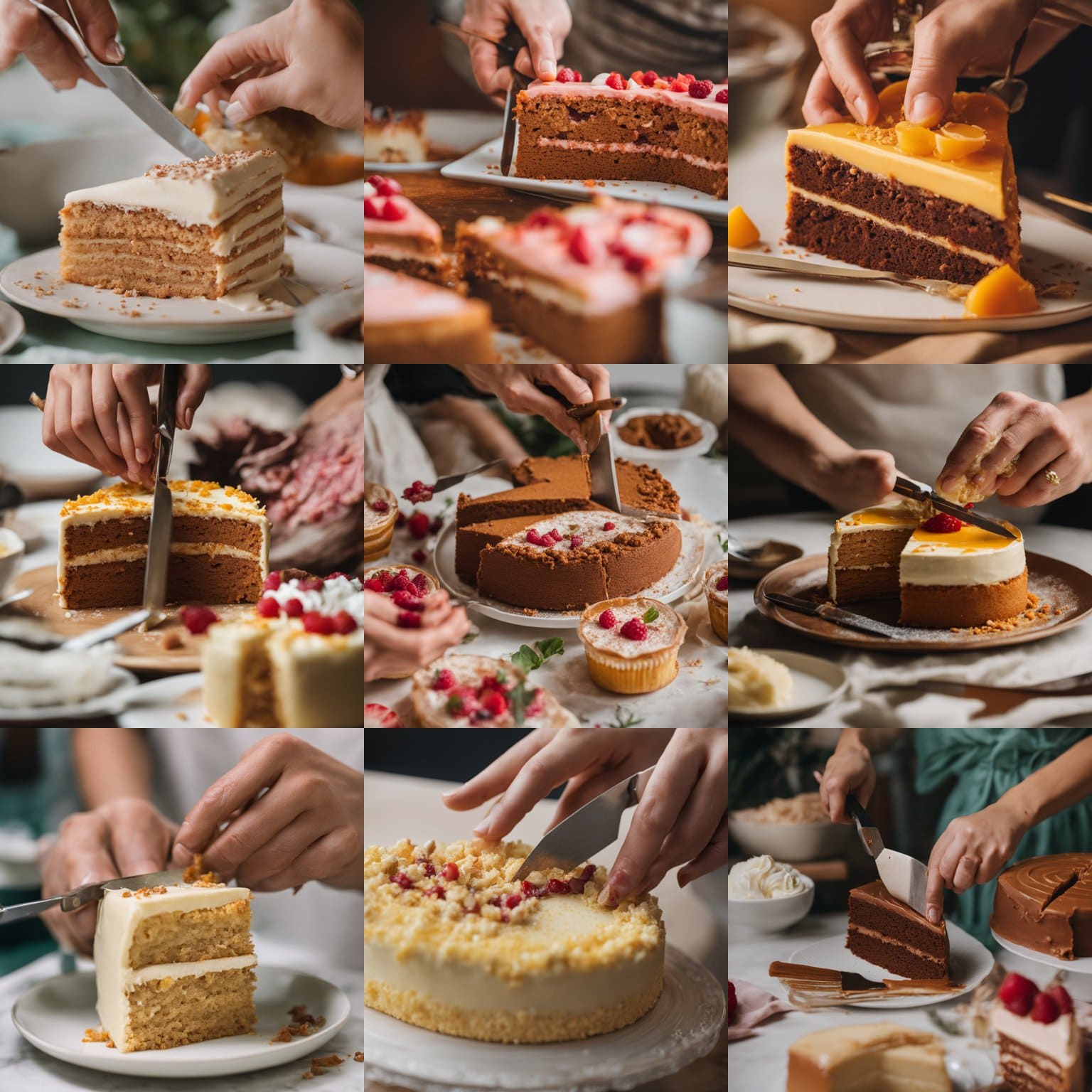}
        \captionsetup{labelformat=empty}
        \caption{SD-XL}
    \end{subfigure}
    \hfill
    \begin{subfigure}{0.33\linewidth}
        \centering
        \textbf{Cake type}\par\medskip
        \vspace*{-0.12cm}
        \includegraphics[width=\linewidth]{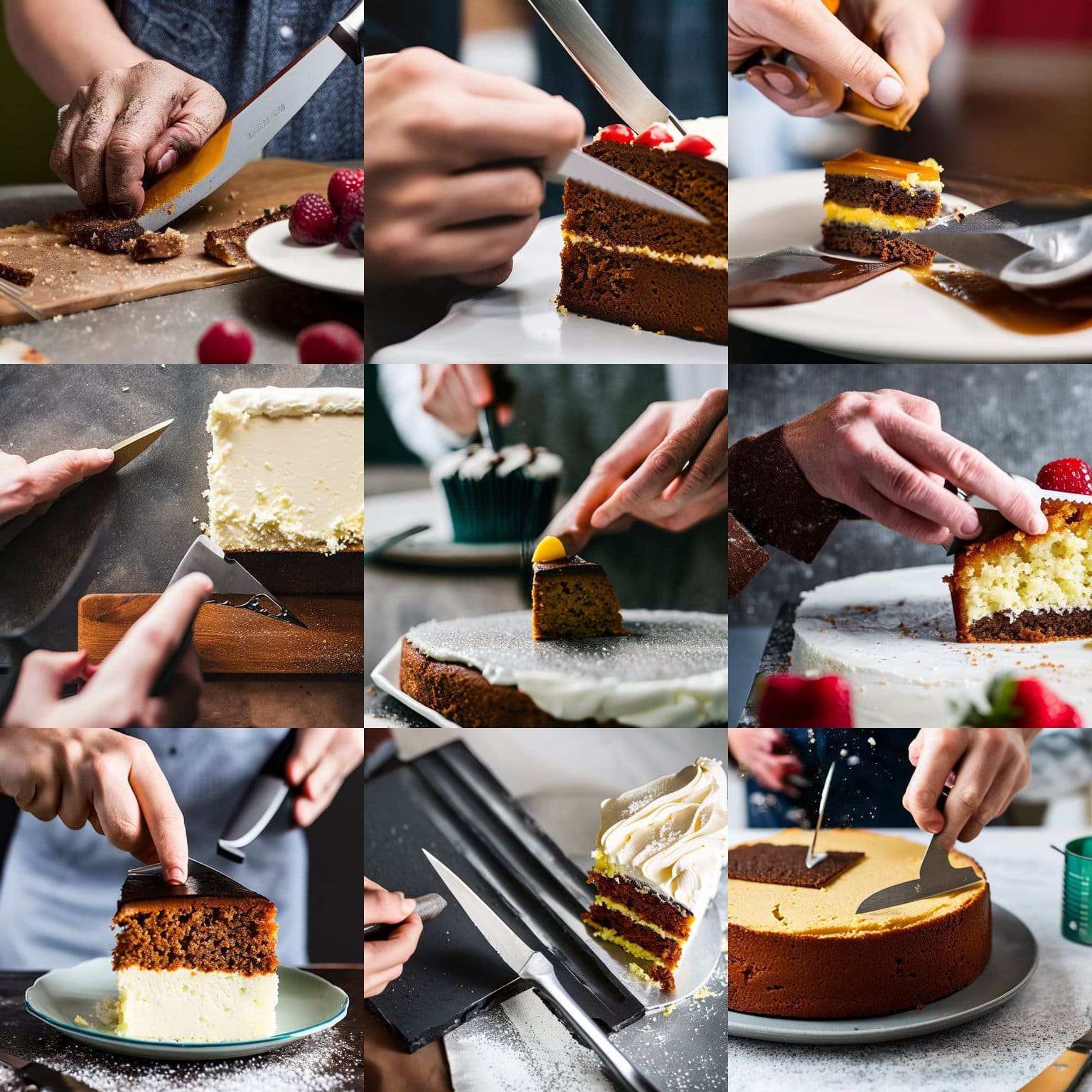}
        \captionsetup{labelformat=empty, justification=centering}
        \caption{SD-2}
    \end{subfigure}
    \hfill
    \begin{subfigure}{0.33\linewidth}
        \centering
        \vspace*{-0.12cm}
        \includegraphics[width=\linewidth]{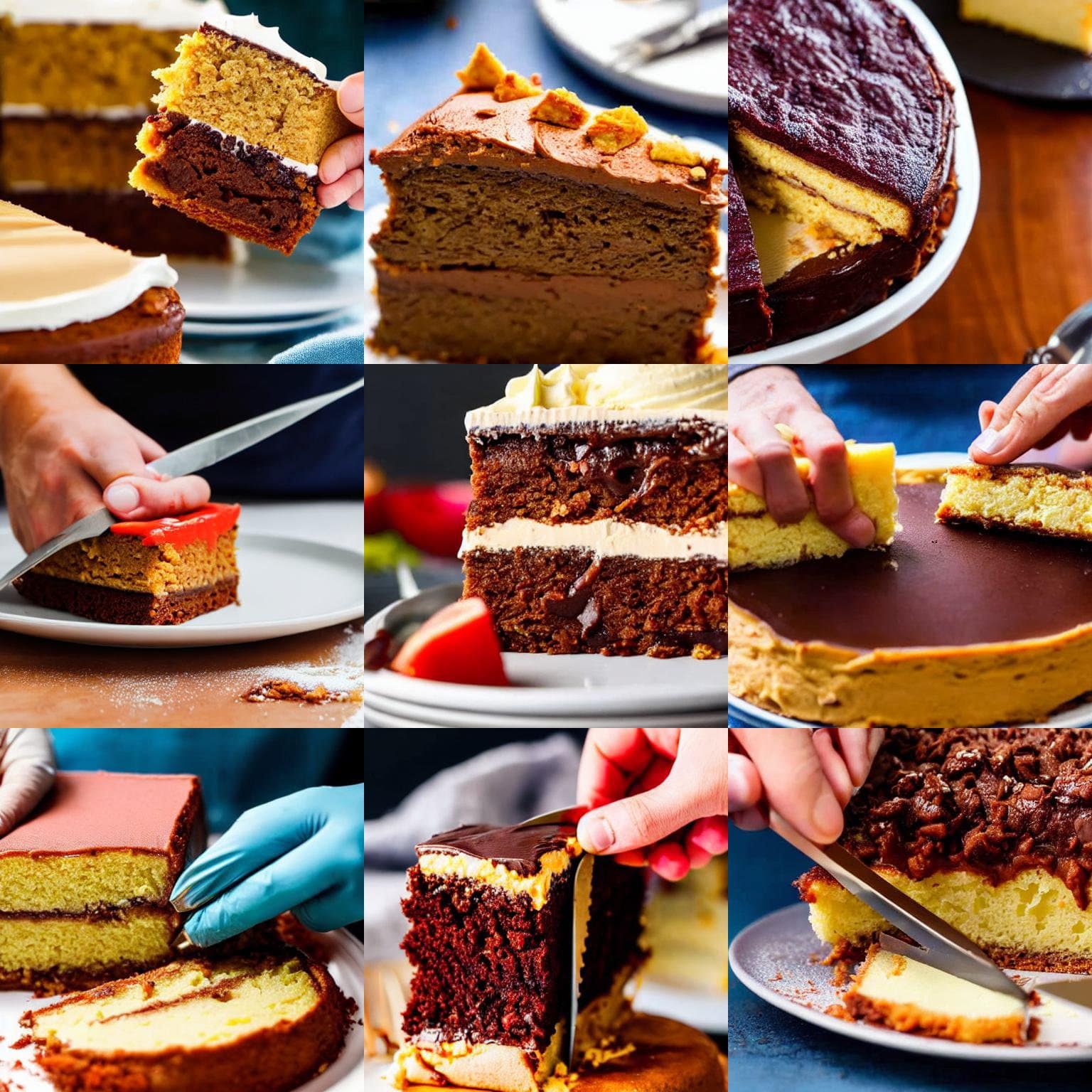}
        \captionsetup{labelformat=empty, justification=centering}
        \caption{SD-1.5}
    \end{subfigure}
    \caption{``A close-up of a person cutting a piece of cake".}
    \label{fig:qualitative_comp_cake}
\end{figure*}

\begin{figure*}[htbp]
    \centering
    \begin{subfigure}{0.33\linewidth}
        \centering
        \vspace*{-0.12cm}
        \includegraphics[width=\linewidth]{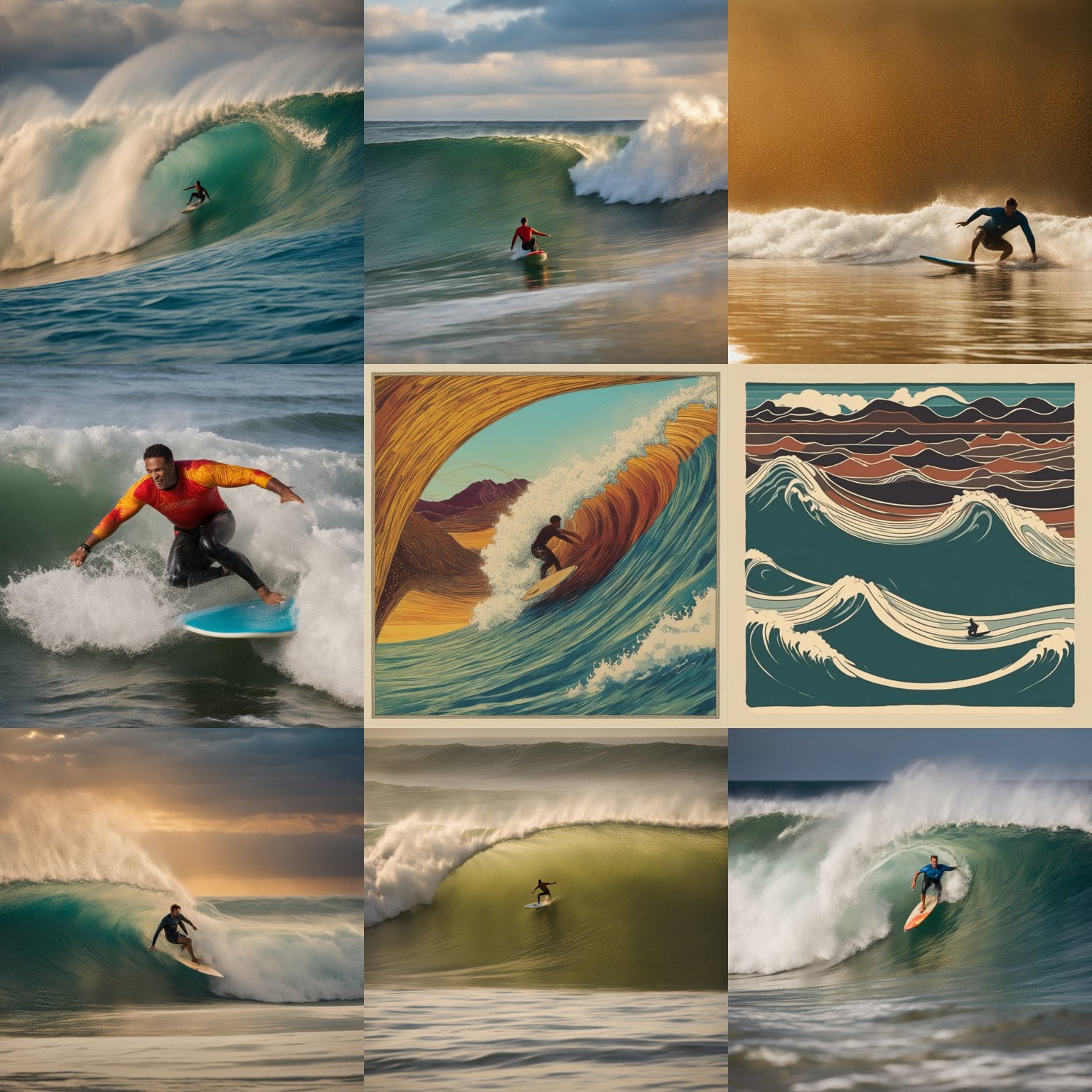}
        \captionsetup{labelformat=empty}
        \caption{SD-XL}
    \end{subfigure}
    \hfill
    \begin{subfigure}{0.33\linewidth}
        \centering
        \textbf{Wave size}\par\medskip
        \vspace*{-0.12cm}
        \includegraphics[width=\linewidth]{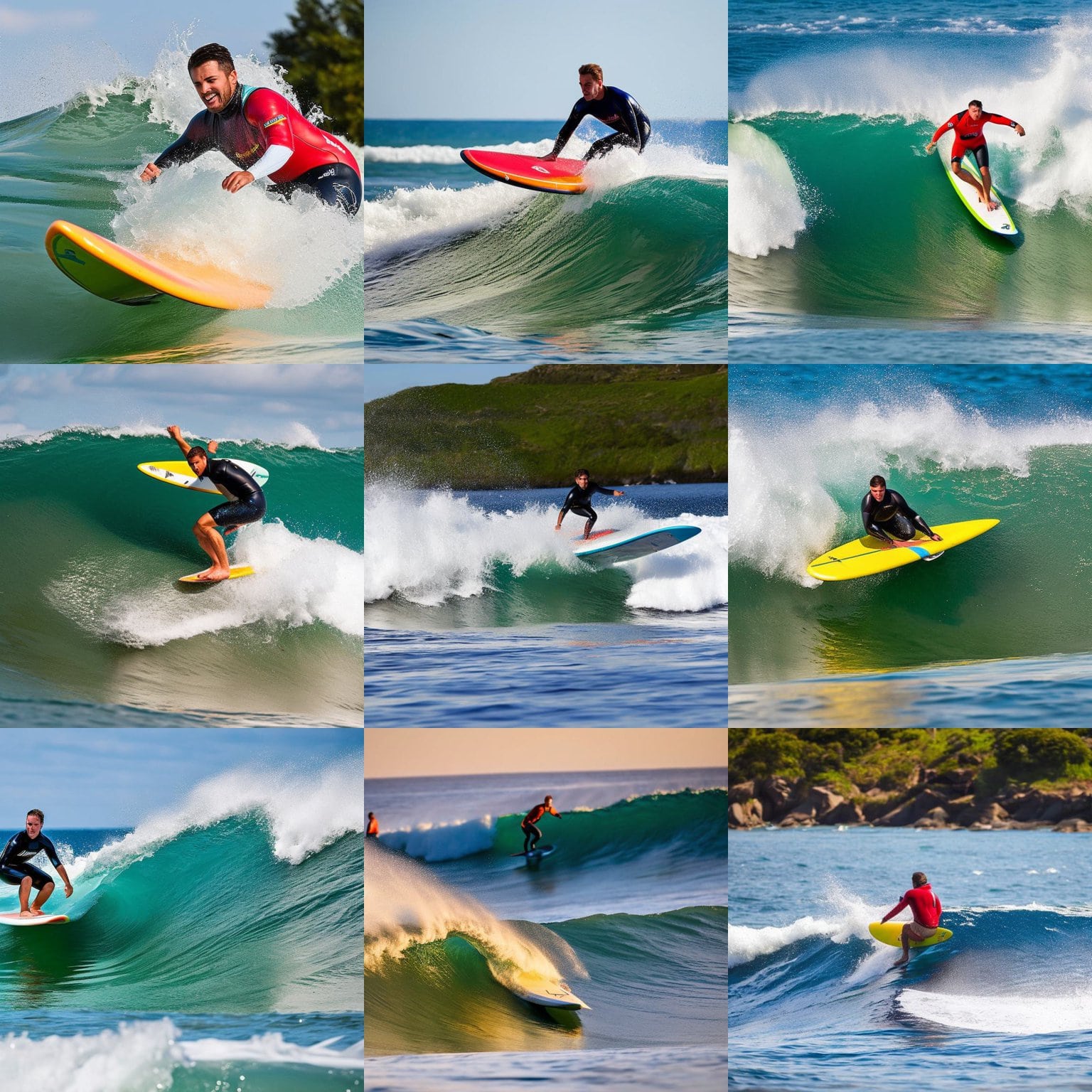}
        \captionsetup{labelformat=empty, justification=centering}
        \caption{SD-2}
    \end{subfigure}
    \hfill
    \begin{subfigure}{0.33\linewidth}
        \centering
        \vspace*{-0.12cm}
        \includegraphics[width=\linewidth]{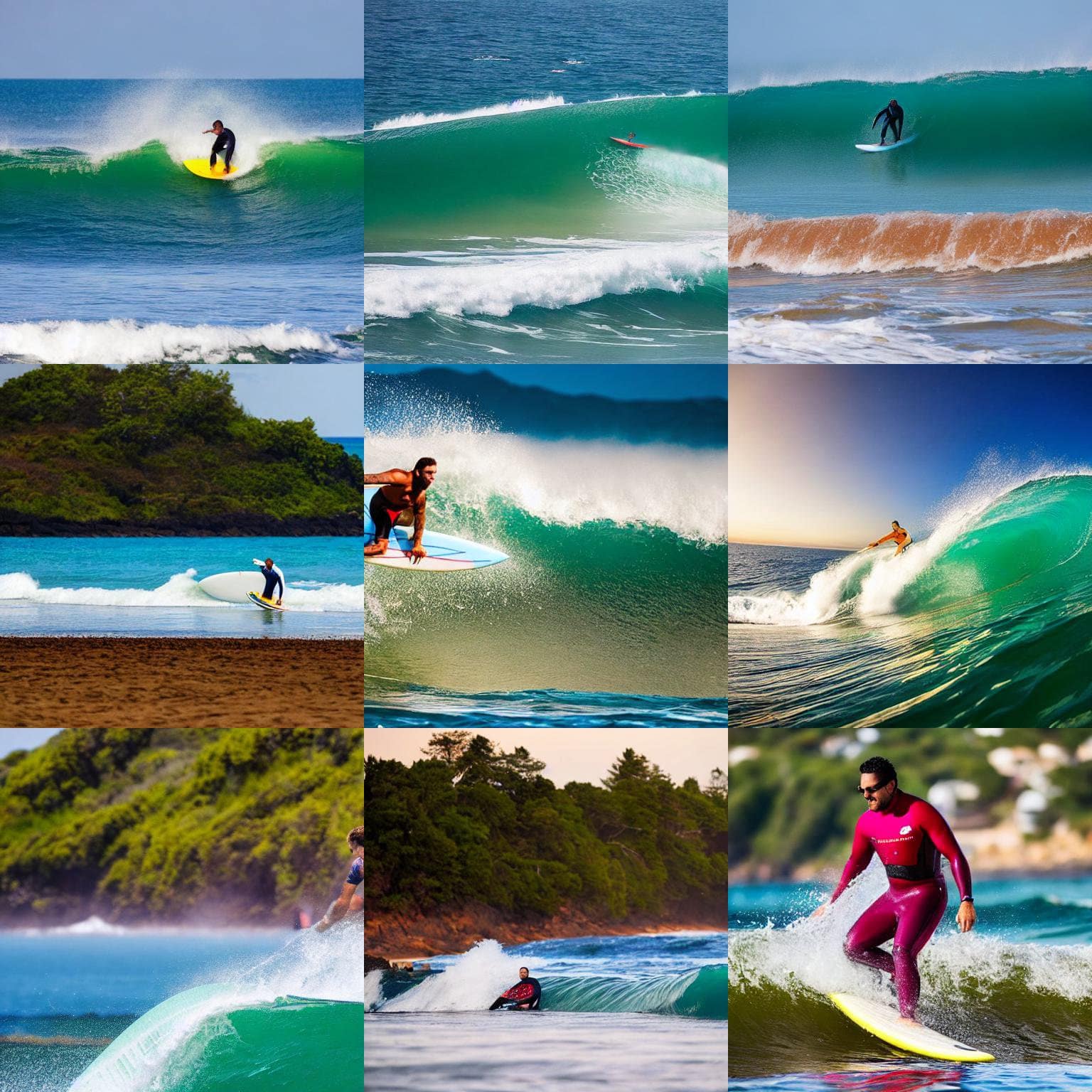}
        \captionsetup{labelformat=empty, justification=centering}
        \caption{SD-1.5}
    \end{subfigure}
    \caption{``A man rides a wave on a surfboard".}
    \label{fig:qualitative_comp_wave}
\end{figure*}

We present more qualitative examples for OpenBias from Figures~\ref{fig:qualitative_comp_gender} through~\ref{fig:qualitative_comp_wave}, which highlight several biases present in the three Stable Diffusion models we examined~\cite{podell2023sdxl, LDM_2022_CVPR}. To facilitate comparison, each bias is shown with images generated using the same randomly selected caption. The biases illustrated include those discussed in
~\Cref{sec:findings} 
of the main paper, such as \textit{“person race”}, \textit{“child race”}, and \textit{“train color”}, as well as some new ones like \textit{“bed type”}, \textit{“cake type”}, and \textit{“wave size”}.

The extent of these biases varies among the models, indicating that they respond differently to the same prompts. This is particularly evident in the \textit{“child race”} bias shown in Figure \ref{fig:qualitative_comp_child_race}, where Stable Diffusion versions 2 and 1.5 tend to produce images of children with lighter skin tones. Similarly, Figure \ref{fig:qualitative_comp_attire} (\textit{“person attire”}) shows these models generating images of individuals dressed more casually compared to the XL version. Overall, the bias levels in these two models are generally lower than in the XL version, which is consistent with the ranking results of
~\Cref{sec:findings}.
Despite this, all three models display these biases, underscoring the robustness of OpenBias. This robustness is further demonstrated in Figure \ref{fig:qualitative_comp_child_gender} (\textit{“child gender”}), where there is a tendency to generate more male children, and in Figure \ref{fig:qualitative_comp_bed} (\textit{“bed type”}), where double beds are predominantly generated.

\section*{D: User Study}
\begin{figure*}[htbp]
    \centering
    \includegraphics[width=0.8\linewidth]{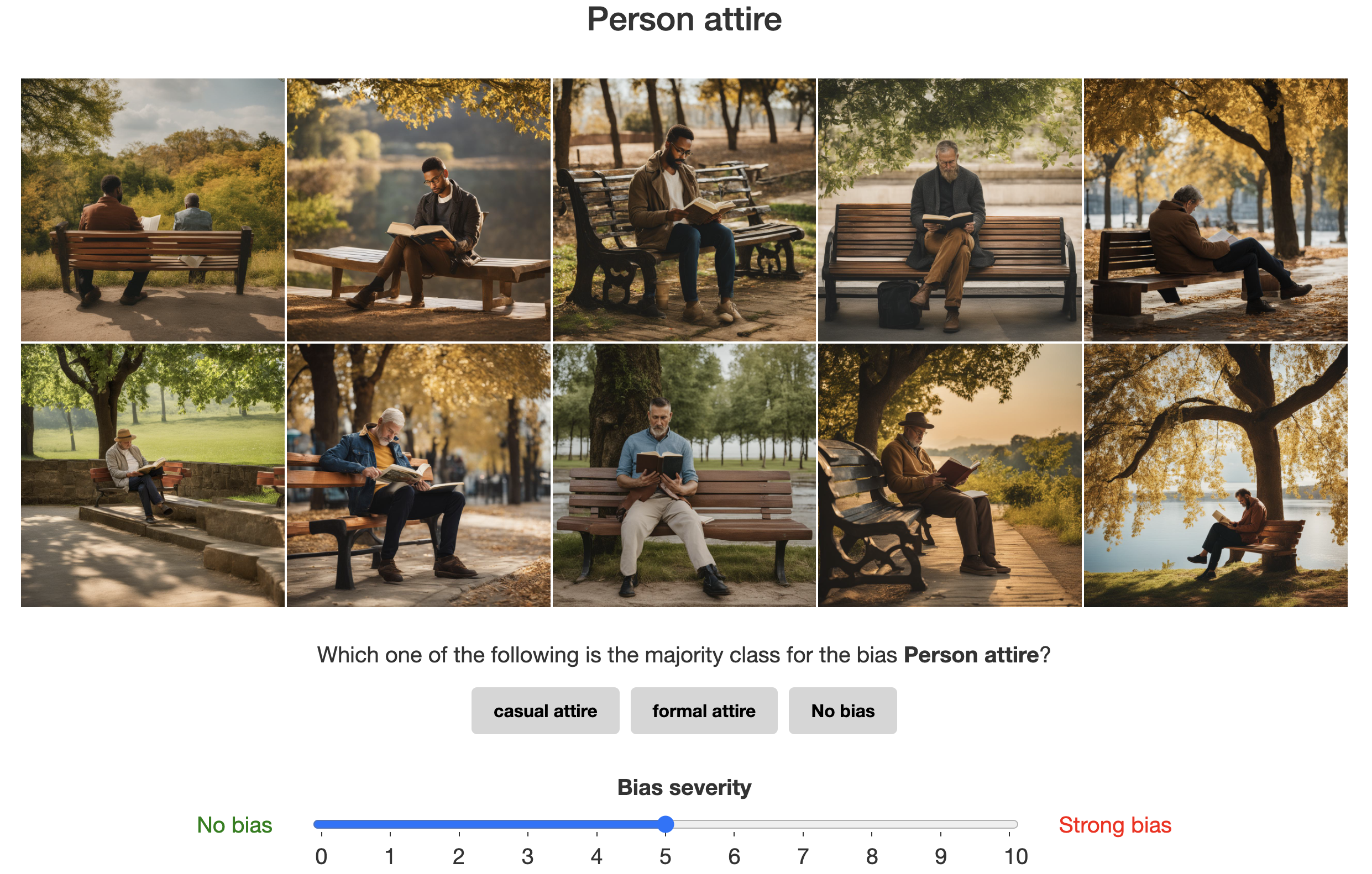}
    \caption{User study screenshot conducted to evaluate OpenBias.}
    \label{fig:user_study_screenshot}
    \vspace{-0.3cm}
\end{figure*}

\Cref{fig:user_study_screenshot} is a screenshot of the user study used to evaluate OpenBias as described
~\Cref{sec:OpenBias_quantitatives}. 
The study shows, for each bias, the generated images at the context level (\ie generated with the same caption). The user has to choose the majority class and the magnitude of each bias.

\end{document}